\documentclass[10pt,twocolumn,letterpaper]{article}

\usepackage[pagenumbers]{cvpr} %
\makeatletter
\@namedef{ver@everyshi.sty}{}
\makeatother

\usepackage{graphicx}
\usepackage{amsmath}
\usepackage{amssymb}
\usepackage{booktabs}
\usepackage{mathtools}
\usepackage{lipsum}
\usepackage{multirow}
\usepackage{textcomp}
\usepackage[numbers,sort]{natbib}
\usepackage{xcolor}
\usepackage{microtype}
\usepackage{overpic}
\usepackage{comment}
\usepackage{pgfplots}
\pgfplotsset{compat=1.17}
\usepackage{pgfplotstable}
\graphicspath{{figures/}}

\usepackage[export]{adjustbox}
\usepackage{tikz}
\usetikzlibrary{calc}

\usepackage[ruled]{algorithm2e} %
\DontPrintSemicolon

\SetAlFnt{\small}
\SetAlCapFnt{\small}
\SetAlCapNameFnt{\small}
\SetAlCapHSkip{0pt}

\makeatletter
\renewcommand\paragraph{\@startsection{paragraph}{4}{\z@}%
	{0.75ex \@plus.5ex \@minus.2ex}%
	{-1em}%
	{\normalfont\normalsize\bfseries\maybe@addperiod}}
\newcommand{\maybe@addperiod}[1]{#1\@addpunct{.}}
\makeatother

\usepackage[breaklinks,colorlinks,bookmarks=false,linkcolor=blue,citecolor=blue]{hyperref}
\usepackage[capitalise,noabbrev,nameinlink]{cleveref}
\creflabelformat{equation}{#2#1#3}
\crefmultiformat{equation}{Equations~#2#1#3}{ and~#2#1#3}{, #2#1#3}{ and~#2#1#3}
\AtBeginDocument{%
\setlength\abovedisplayskip{5pt plus 1pt minus 2.5pt} %
\setlength\belowdisplayskip{5pt plus 1pt minus 2.5pt} %
\setlength\abovedisplayshortskip{0pt plus 3pt} %
\setlength\belowdisplayshortskip{3pt plus 1.5pt minus 1.5pt} %
}

\newcommand{\tb}[1]{\textbf{#1}}

\def\cvprPaperID{0000} %
\def\confName{CVPR}
\def\confYear{2023}

\begin{document}

\newcommand{\mo}[1]{\textcolor{red}{[\texttt{Matt}: #1]}}
\newcommand{\MZ}[1]{\textcolor{red}{[\texttt{MZ}: #1]}}

\newcommand{\isdraft}{false}

\newcommand{\flux}{\Phi}
\newcommand{\normal}{\mathbf{n}}
\newcommand{\position}{\mathbf{x}}
\newcommand{\xpos}{\mathbf{x}}
\newcommand{\ypos}{\mathbf{y}}
\newcommand{\direction}{\boldsymbol{\omega}}
\newcommand{\dirin}{\direction_\text{i}}
\newcommand{\dirout}{\vec{\direction}}
\newcommand{\hemisphere}{\altmathcal{H}^2}
\newcommand{\withnorm}{[\normal]}
\newcommand{\withgeom}{[\normal_g]}
\newcommand{\withshading}{[\normal_s]}
\newcommand{\withboth}{[\normal_g, \normal_s]}
\newcommand{\inout}{\dirin \rightarrow \dirout}
\newcommand{\outin}{\dirout \rightarrow \dirin}
\newcommand{\emitted}{L_\text{e}}
\newcommand{\importance}{W_\text{e}}
\newcommand{\freeflight}{p_\text{ff}}
\newcommand{\im}{\mathrm{i}}
\newcommand{\tnear}{{t_\text{n}}}
\newcommand{\tfar}{{t_\text{f}}}
\newcommand{\network}{F_{\boldsymbol\theta}}
\newcommand{\rgbL}{L_{\text{RGB}}}
\newcommand{\irL}{L_{\text{IR}}}
\newcommand{\tofL}{L_{\text{ToF}}}
\newcommand{\rgbhatL}{\hat{L}_{\text{RGB}}}
\newcommand{\irhatL}{\hat{L}_{\text{IR}}}
\newcommand{\tofhatL}{\hat{L}_{\text{ToF}}}
\newcommand{\Iref}{I_{\text{s}}}
\newcommand{\timet}{\tau}
\newcommand{\latent}{\mathbf{z}}
\newcommand{\static}{\text{stat}}
\newcommand{\dynamic}{\text{dyn}}

\newcommand{\firstplane}{\pi^{xy}}
\newcommand{\secondplane}{\pi^{uv}}
\newcommand{\posenc}{\boldsymbol\gamma}
\newcommand{\ray}{\mathbf{r}}
\newcommand{\rayparam}{x, y, u, v}
\newcommand{\voxset}{\altmathcal{V}}
\newcommand{\vox}{\upsilon}
\newcommand{\col}{\mathbf{c}}

\newcommand{\mpage}[2]
{
\begin{minipage}{#1\linewidth}\centering
#2
\end{minipage}
}

\title{
HyperReel: High-Fidelity 6-DoF Video with Ray-Conditioned Sampling
}

\author{%
\parbox{.2\linewidth}{\centering%
Benjamin Attal$^{1,4}$
}
\hspace{-3mm}\and\hspace{-3mm}
\parbox{.2\linewidth}{\centering%
Jia-Bin Huang$^{2,4}$%
}
\hspace{-3mm}\and\hspace{-3mm}
\parbox{.2\linewidth}{\centering%
Christian Richardt$^{3}$%
}
\hspace{-3mm}\and\hspace{-3mm}
\parbox{.2\linewidth}{\centering%
Michael Zollh\"ofer$^{3}$%
}
\hspace{-3mm}\and\hspace{-3mm}
\parbox{.2\linewidth}{\centering%
Johannes Kopf$^{4}$%
}
\hspace{-3mm}\and\hspace{-3mm}
\parbox{.2\linewidth}{\centering%
Matthew O'Toole$^{1}$%
}
\hspace{-3mm}\and\hspace{-3mm}
\parbox{.2\linewidth}{\centering%
Changil Kim$^{4}$%
}
}

\twocolumn[{
\renewcommand\twocolumn[1][]{#1}
\maketitle
\vspace{-10mm}
\begin{center}
$^1$Carnegie Mellon University\quad
$^2$University of Maryland\quad
$^3$Reality Labs Research\quad
$^4$Meta\quad
\end{center}
\begin{center}
\textbf{\url{https://hyperreel.github.io}}
\end{center}
\vspace{7mm}
\vspace{-5mm}
\centering
\footnotesize
\captionsetup{type=figure}
  \mpage{0.02}{%
    \rotatebox{90}{Dynamic 6-DoF rendering}%
  }%
  \hspace{-1.5mm}\hfill%
  \mpage{0.975}{%
    \adjincludegraphics[width=0.14\linewidth,trim={{0.25\width} {0.25\height} {0.35\width} {0.25\height}},clip]{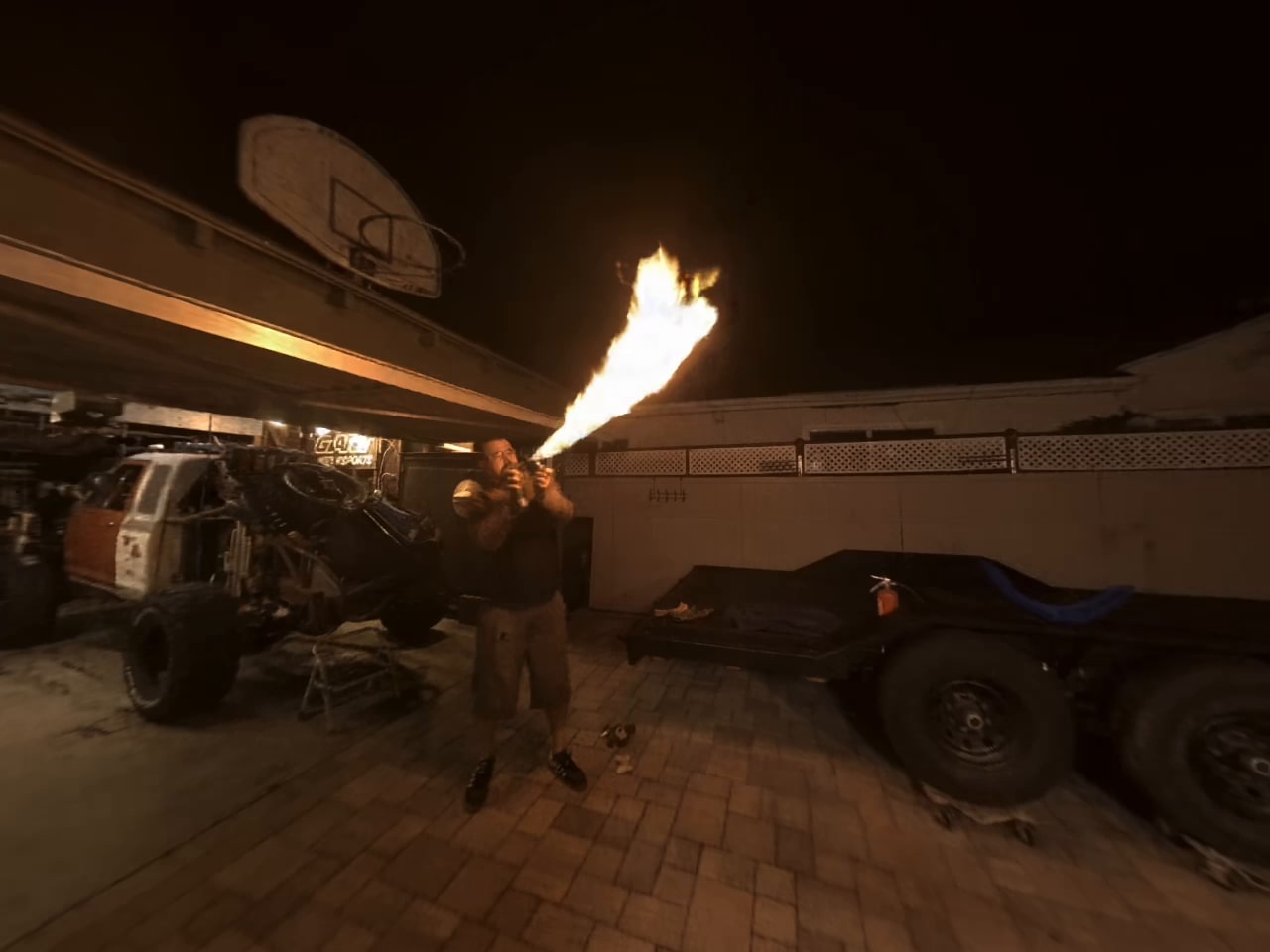}\hfill%
    \adjincludegraphics[width=0.14\linewidth,trim={{0.25\width} {0.25\height} {0.35\width} {0.25\height}},clip]{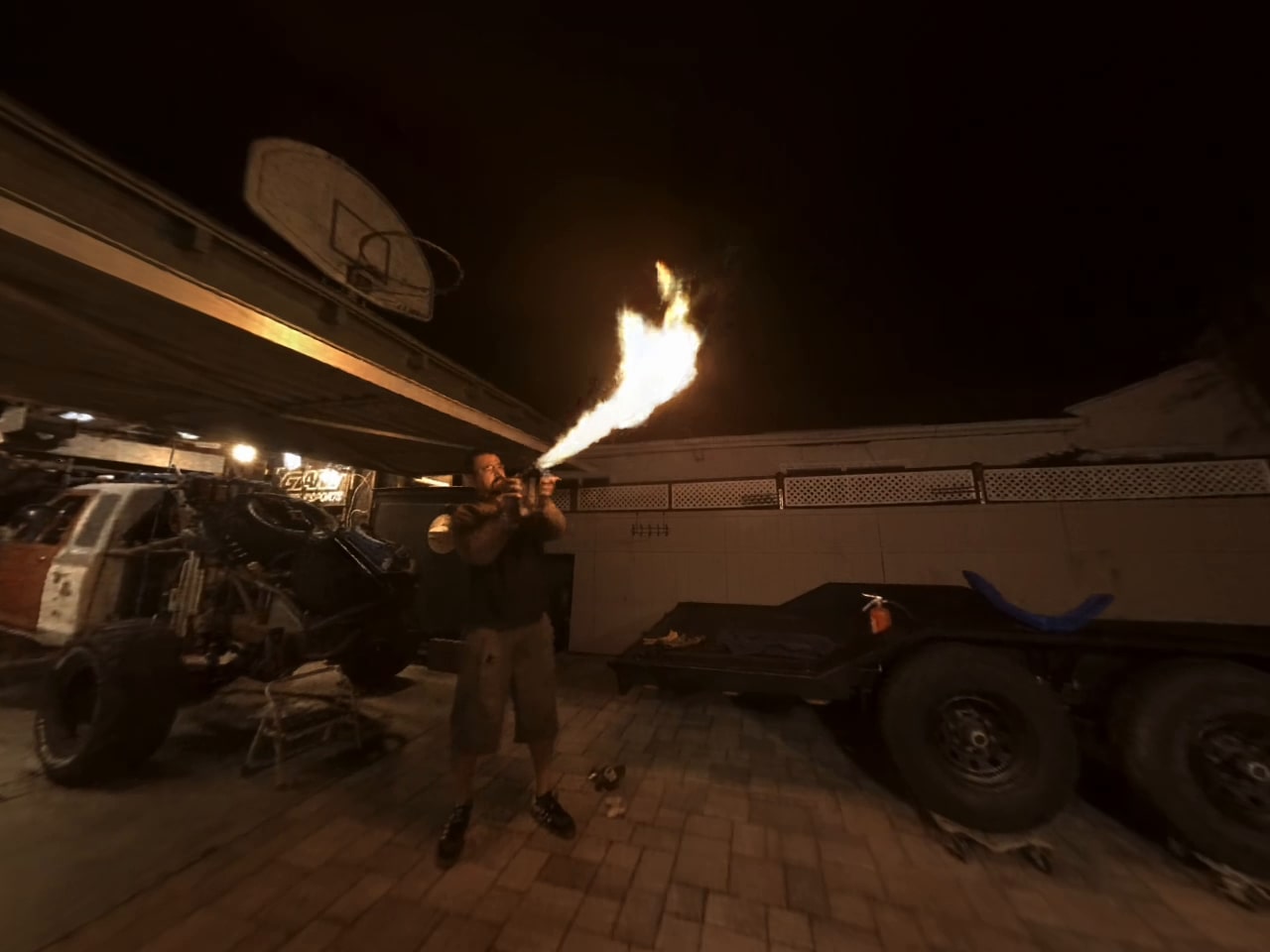}\hfill%
    \adjincludegraphics[width=0.14\linewidth,trim={{0.25\width} {0.25\height} {0.35\width} {0.25\height}},clip]{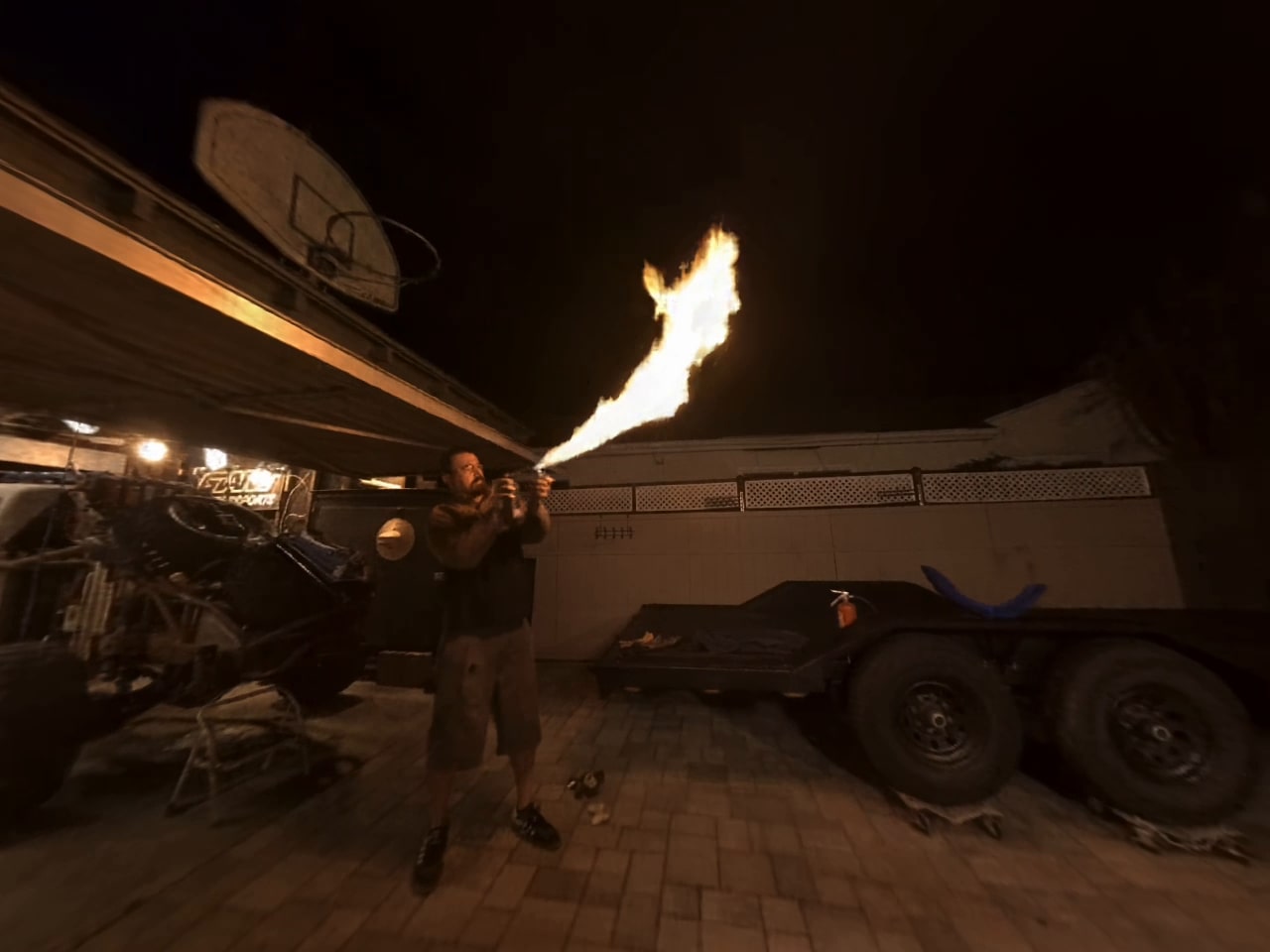}\hfill%
    \adjincludegraphics[width=0.14\linewidth,trim={{0.25\width} {0.25\height} {0.35\width} {0.25\height}},clip]{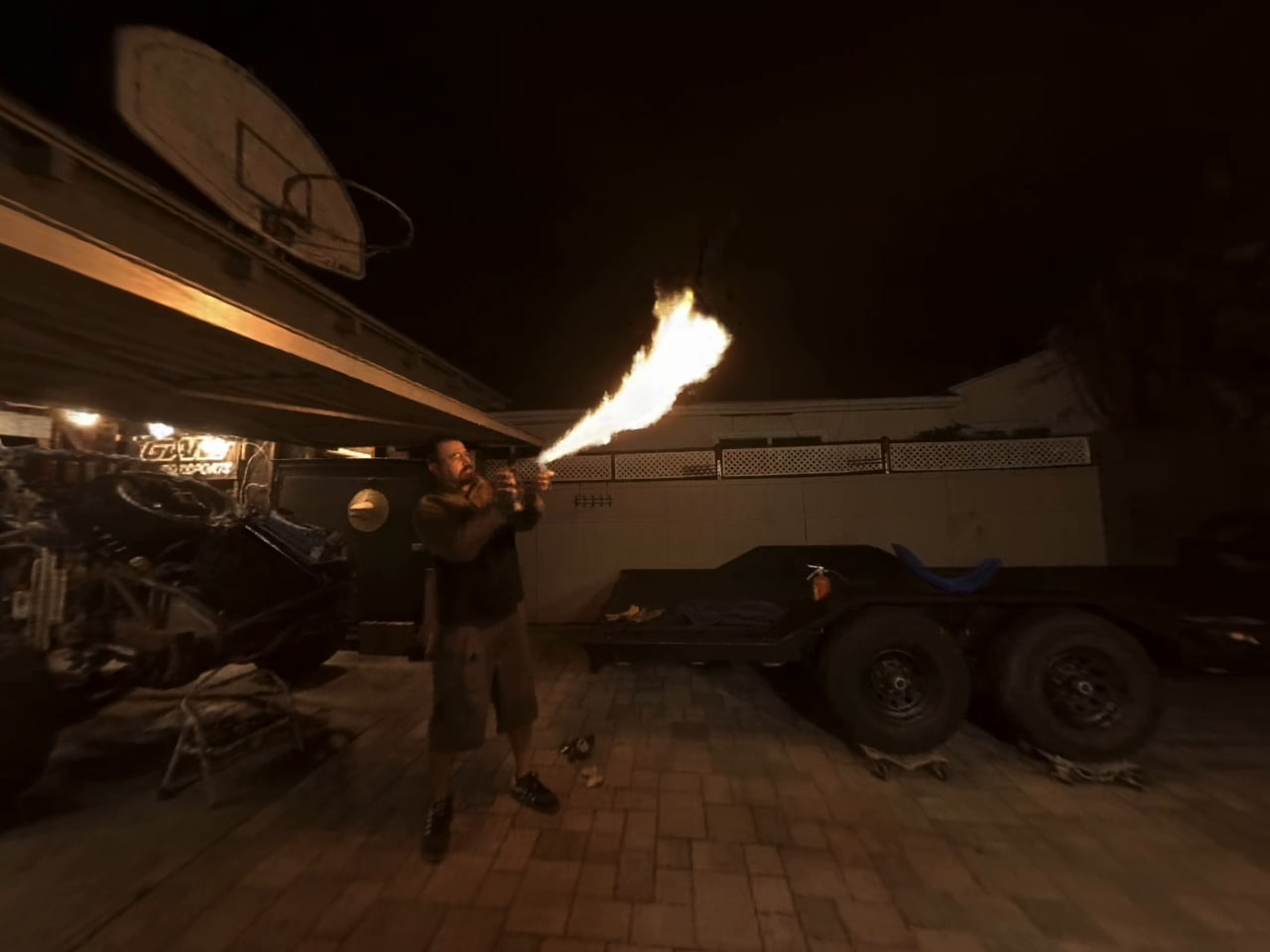}\hfill%
    \adjincludegraphics[width=0.14\linewidth,trim={{0.25\width} {0.25\height} {0.35\width} {0.25\height}},clip]{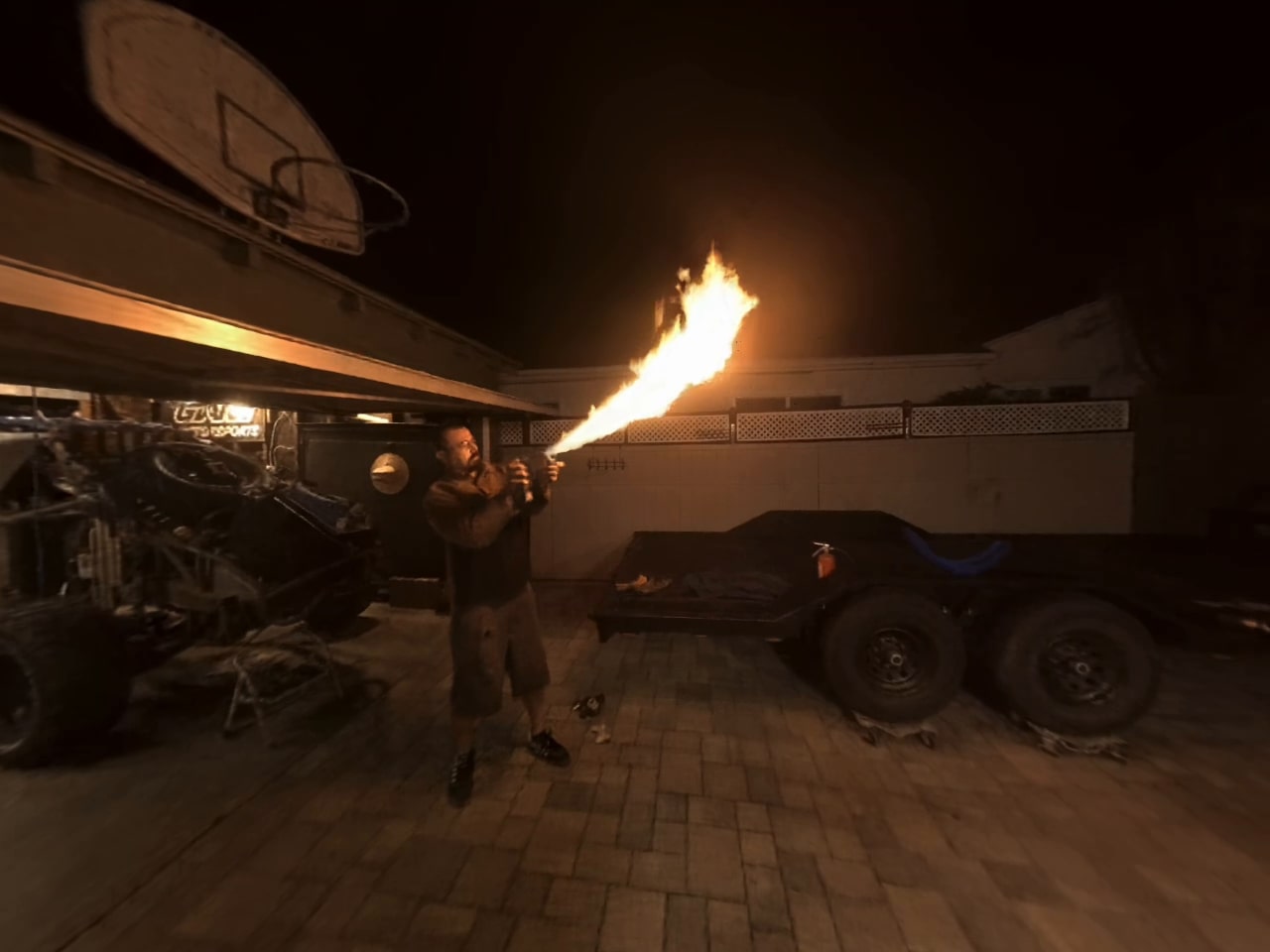}\hfill%
    \adjincludegraphics[width=0.14\linewidth,trim={{0.25\width} {0.25\height} {0.35\width} {0.25\height}},clip]{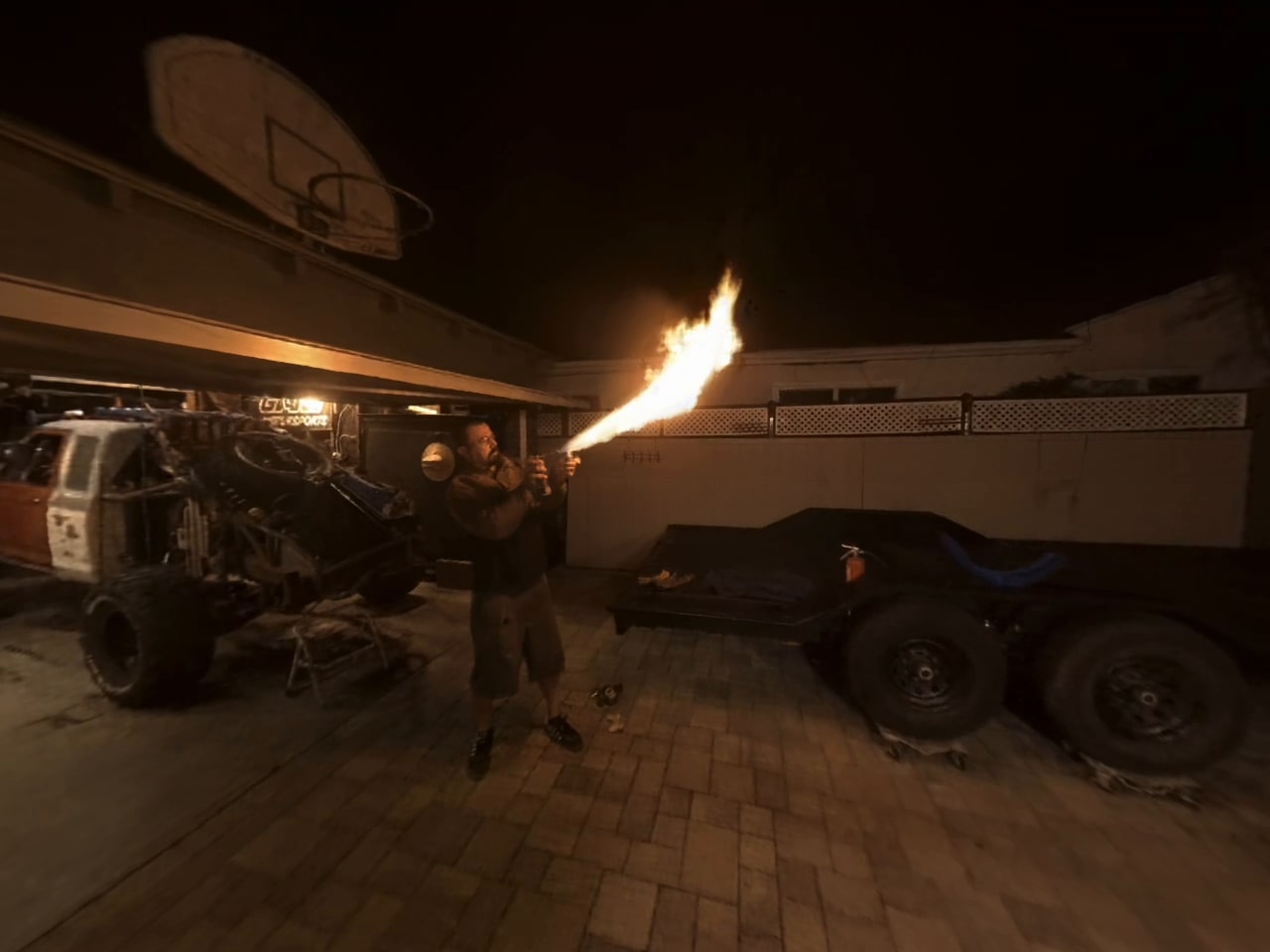}\hfill%
    \adjincludegraphics[width=0.14\linewidth,trim={{0.25\width} {0.25\height} {0.35\width} {0.25\height}},clip]{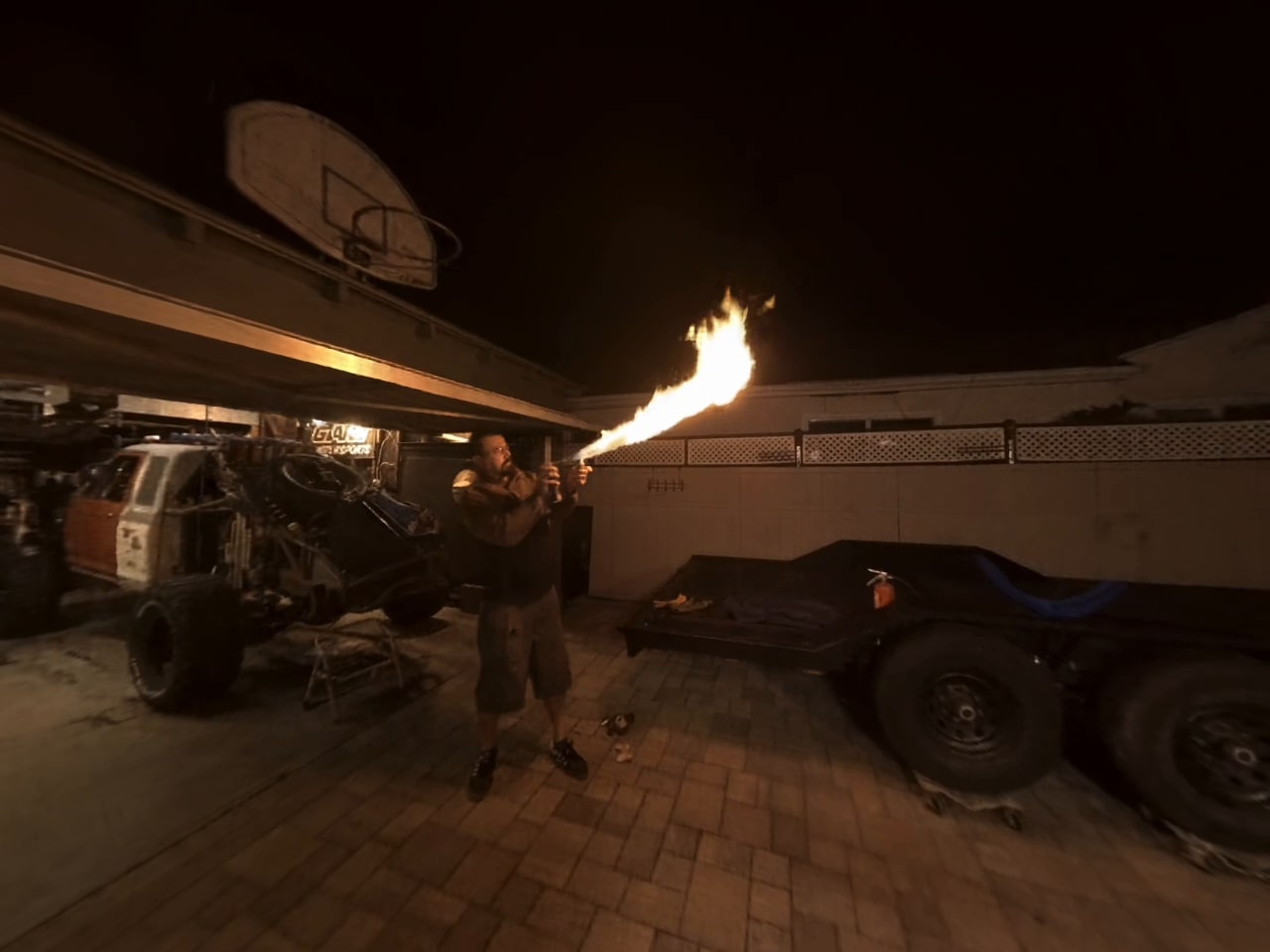}\\[0.5mm]%
    \adjincludegraphics[height=0.14\linewidth,trim={{0.14\width} {0.0\height} {0.3\width} {0.0\height}},clip,angle=90]{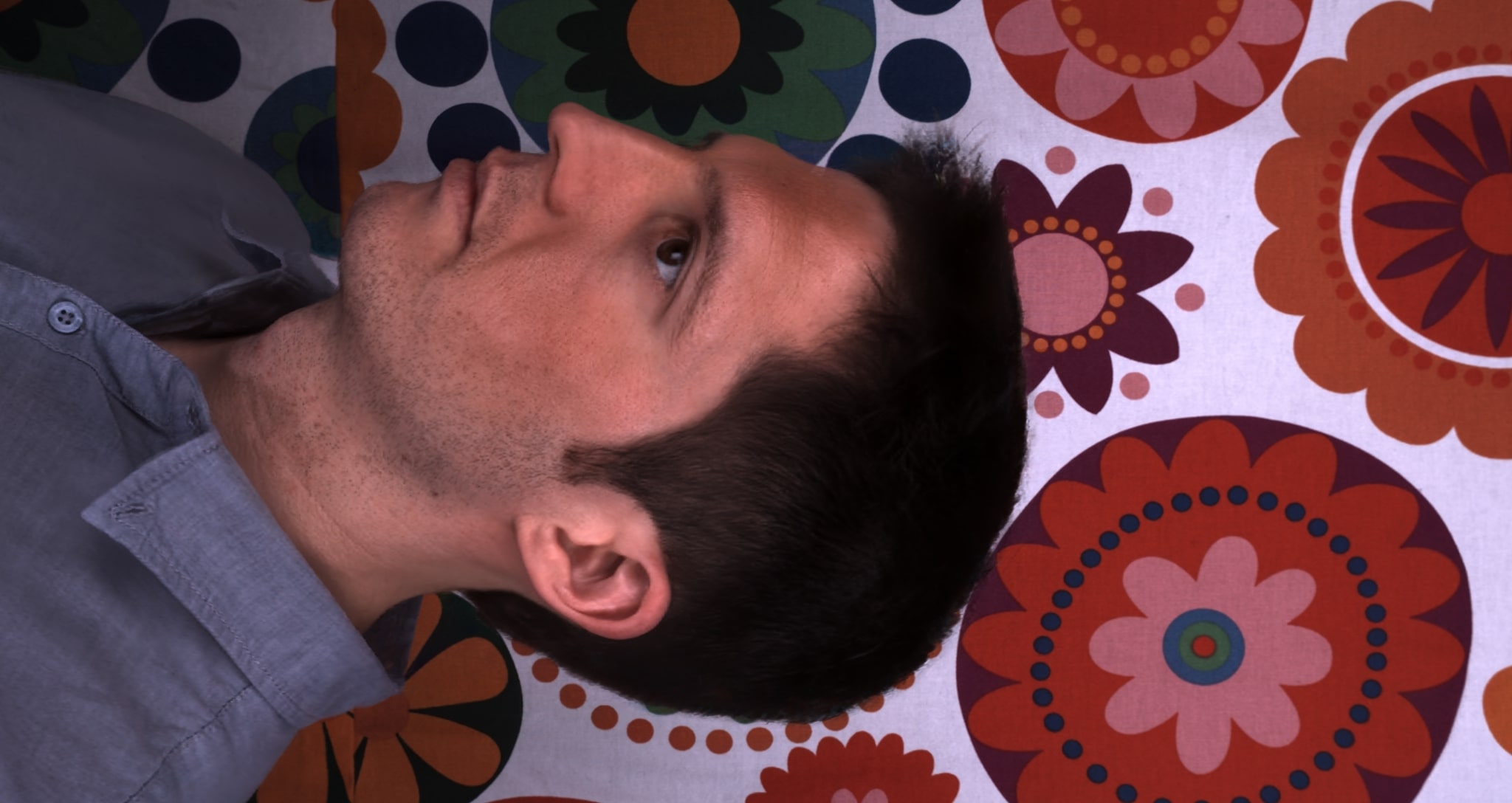}\hfill%
    \adjincludegraphics[height=0.14\linewidth,trim={{0.14\width} {0.0\height} {0.3\width} {0.0\height}},clip,angle=90]{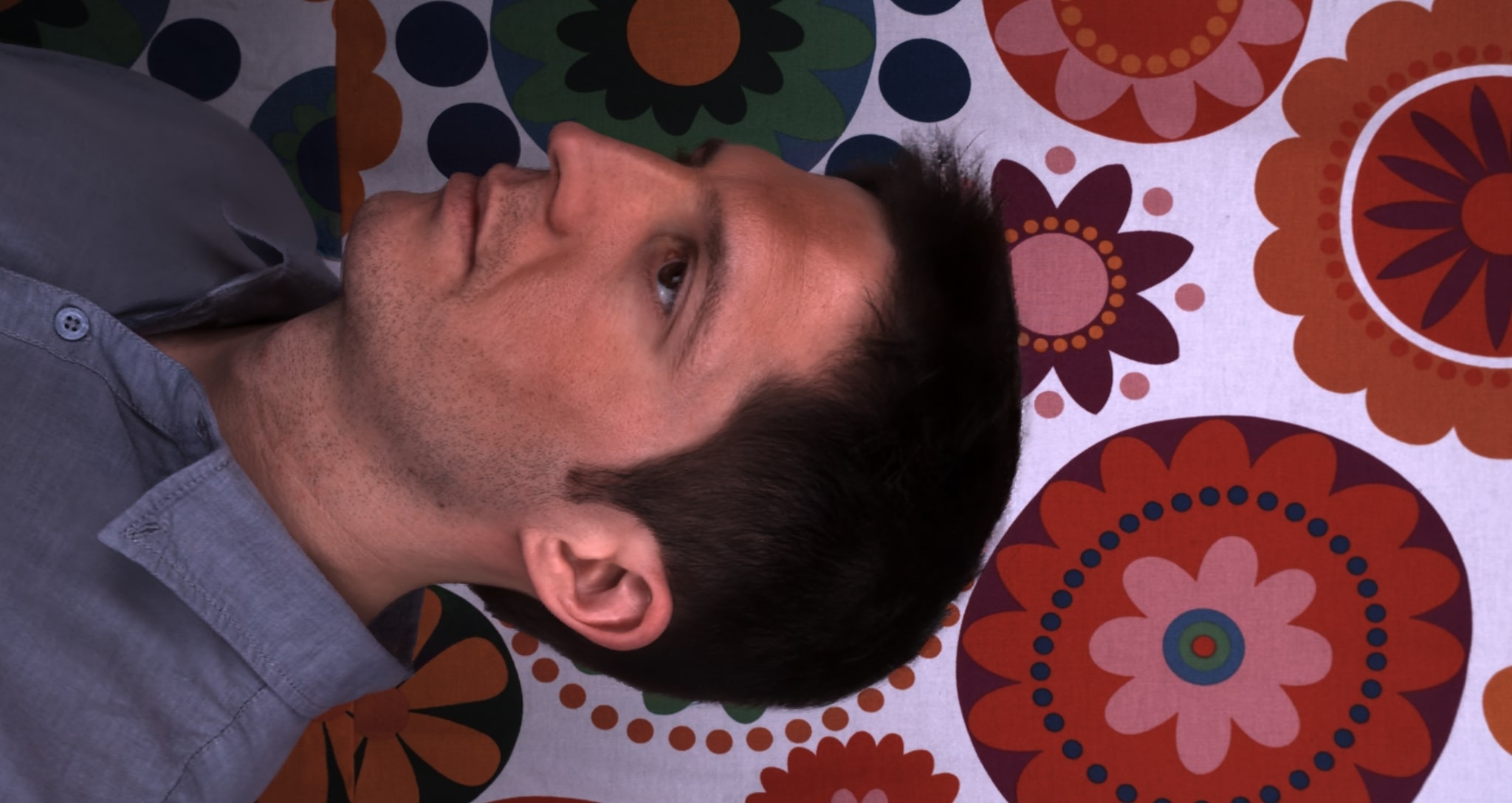}\hfill%
    \adjincludegraphics[height=0.14\linewidth,trim={{0.14\width} {0.0\height} {0.3\width} {0.0\height}},clip,angle=90]{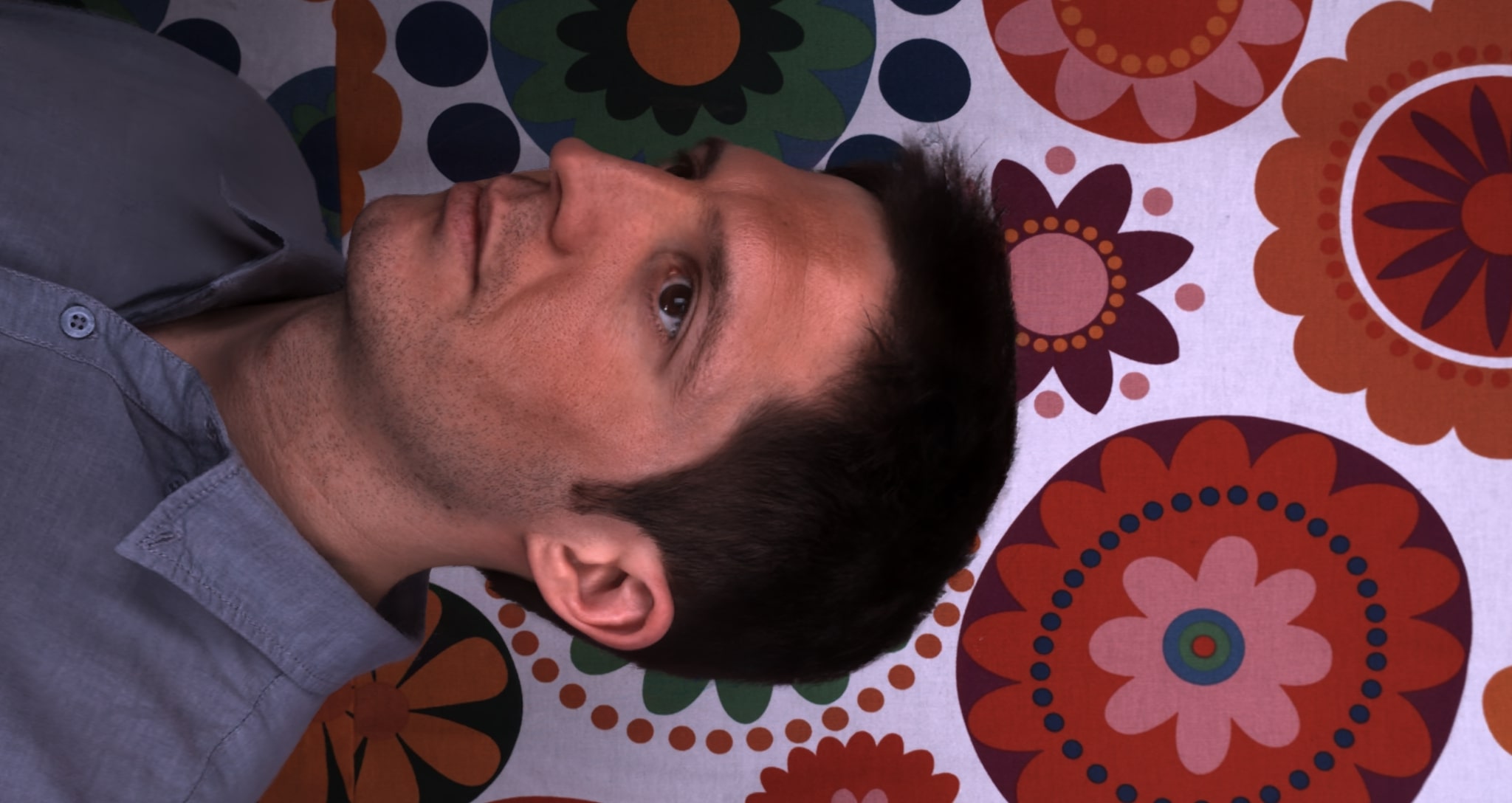}\hfill%
    \adjincludegraphics[height=0.14\linewidth,trim={{0.14\width} {0.0\height} {0.3\width} {0.0\height}},clip,angle=90]{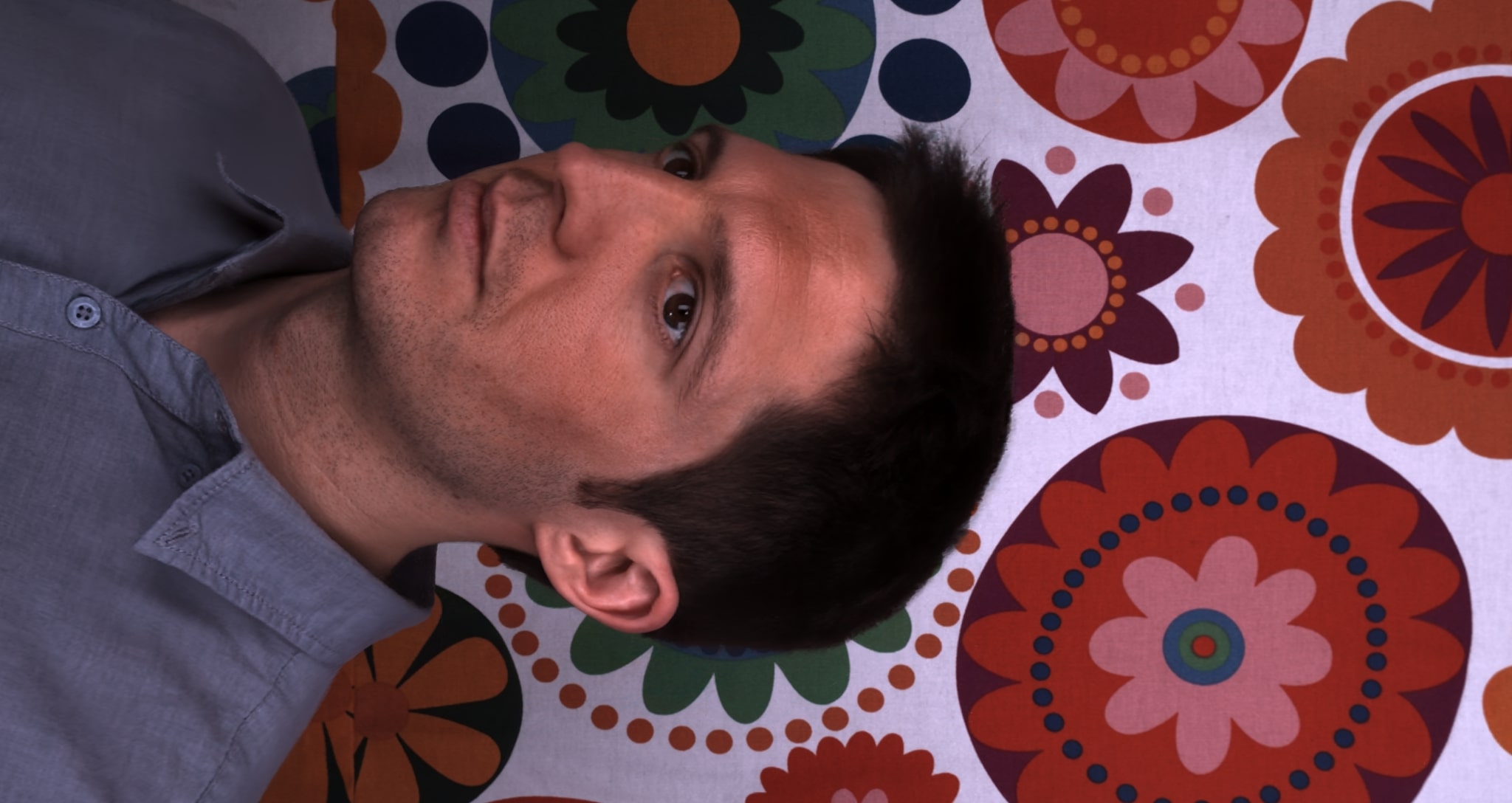}\hfill%
    \adjincludegraphics[height=0.14\linewidth,trim={{0.14\width} {0.0\height} {0.3\width} {0.0\height}},clip,angle=90]{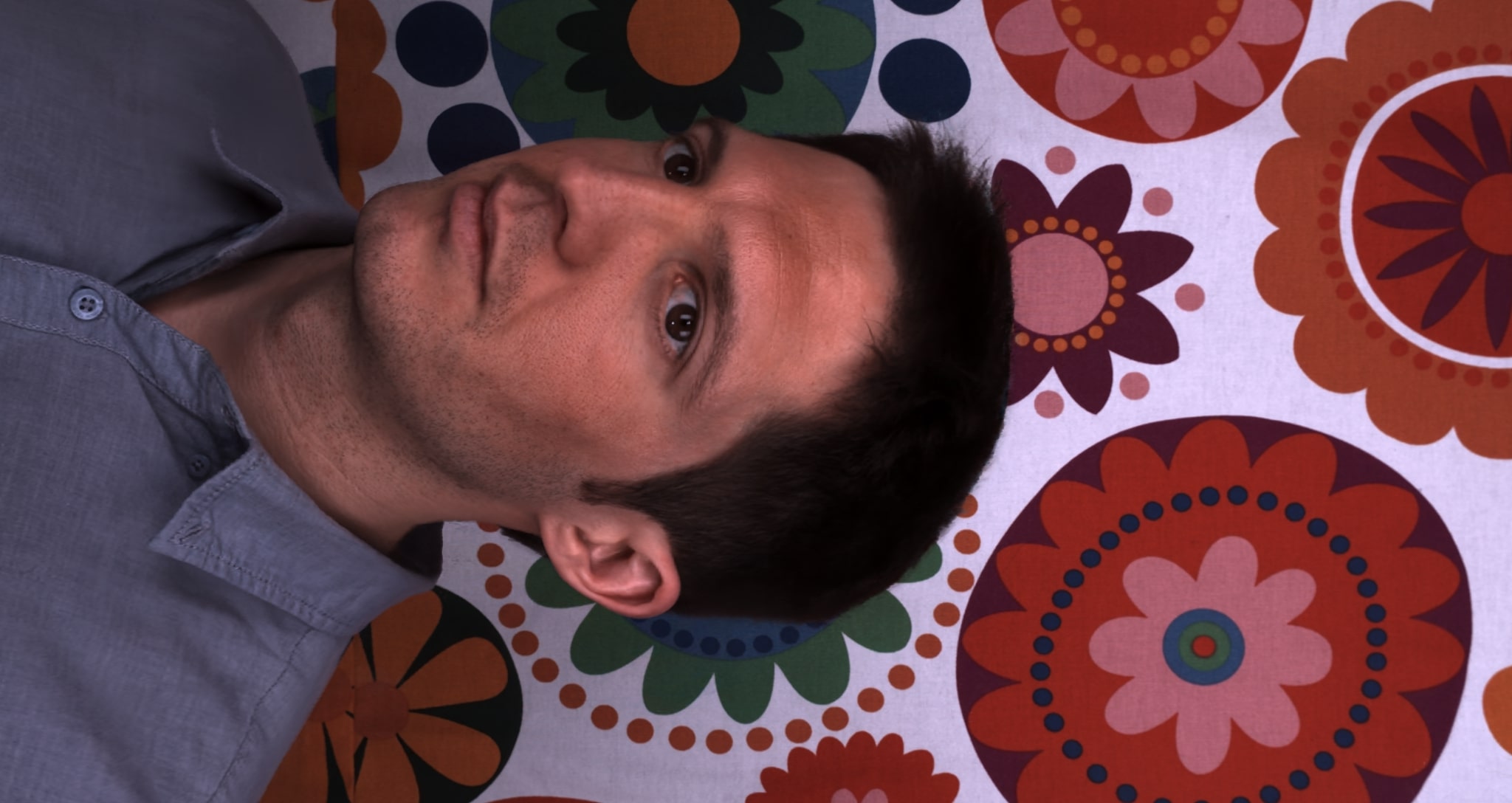}\hfill%
    \adjincludegraphics[height=0.14\linewidth,trim={{0.14\width} {0.0\height} {0.3\width} {0.0\height}},clip,angle=90]{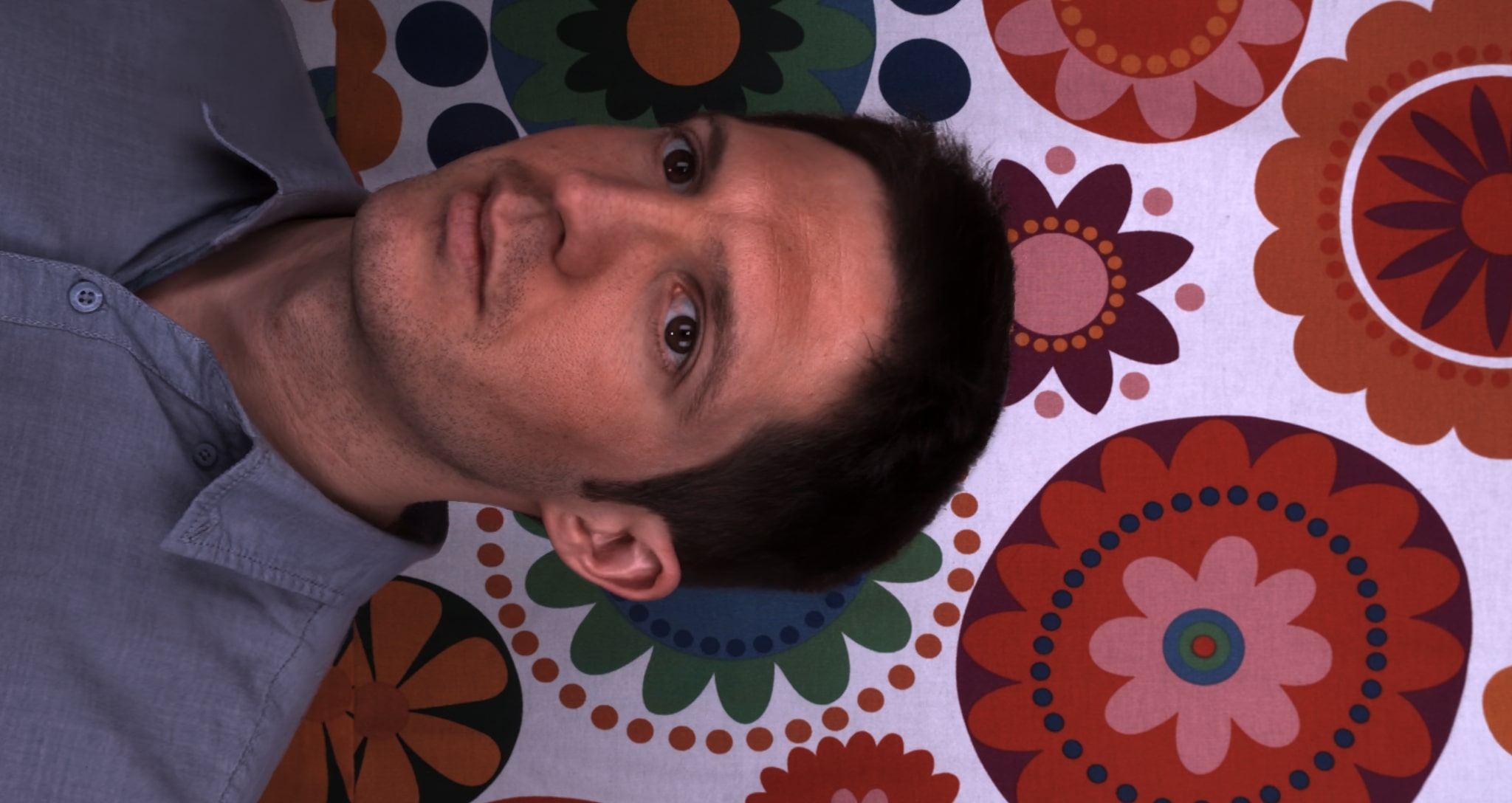}\hfill%
    \adjincludegraphics[height=0.14\linewidth,trim={{0.14\width} {0.0\height} {0.3\width} {0.0\height}},clip,angle=90]{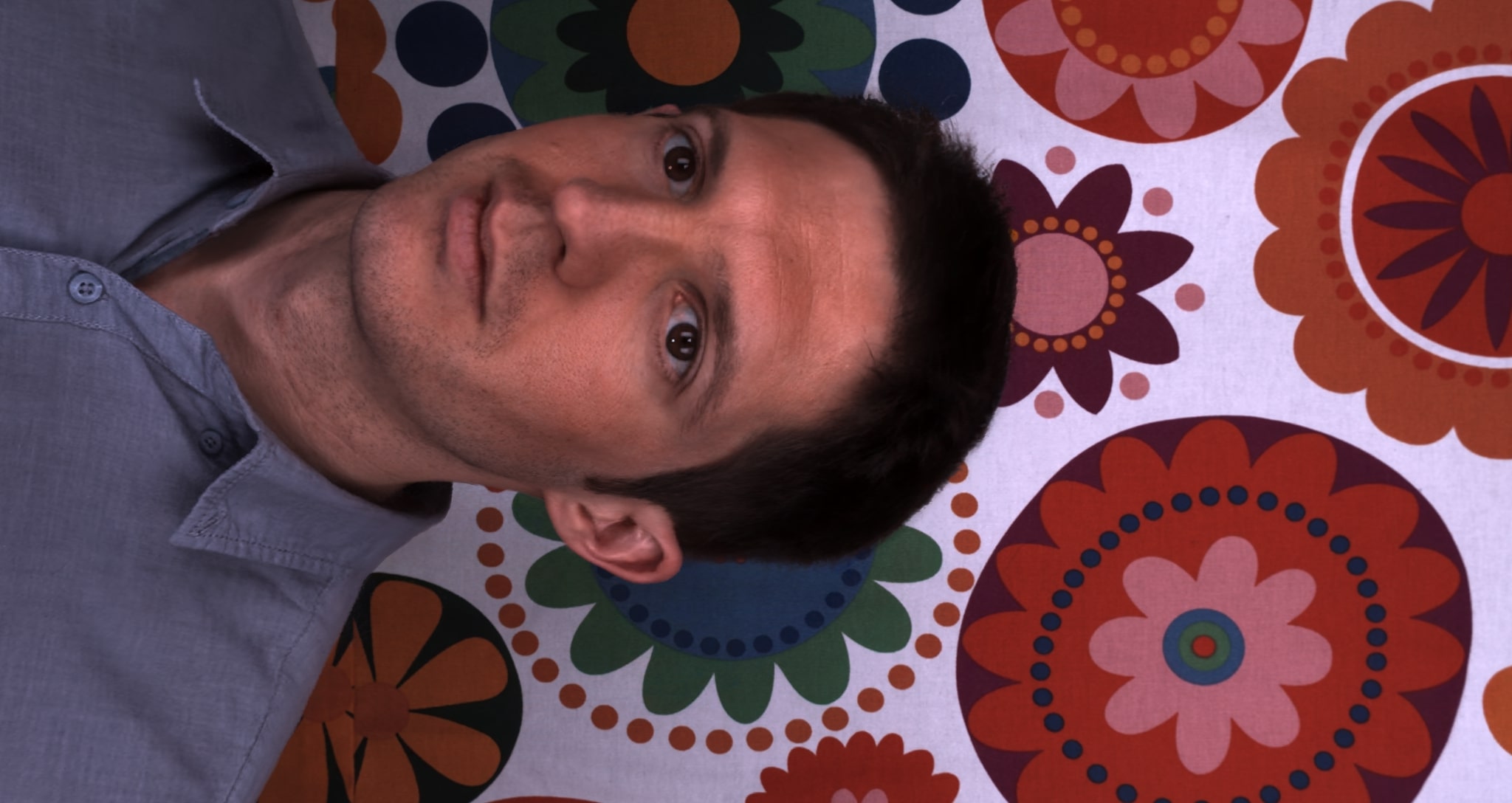}\\[-1mm]%
    \parbox{0.14\linewidth}{\centering $t=0$\,s}\hfill%
    \parbox{0.14\linewidth}{\centering $t=0.25$\,s}\hfill%
    \parbox{0.14\linewidth}{\centering $t=0.5$\,s}\hfill%
    \parbox{0.14\linewidth}{\centering $t=0.75$\,s}\hfill%
    \parbox{0.14\linewidth}{\centering $t=1$\,s}\hfill%
    \parbox{0.14\linewidth}{\centering $t=1.25$\,s}\hfill%
    \parbox{0.14\linewidth}{\centering $t=1.5$\,s}\\[2mm]%
  }%
  \\[0.5mm]%
  \rule{\linewidth}{1pt}%
  \\[0.5mm]%
  \mpage{0.02}{%
    \rotatebox{90}{Static 6-DoF rendering}%
  }%
  \hspace{-1.5mm}\hfill%
  \mpage{0.975}{%
    \adjincludegraphics[width=0.14\linewidth,trim={{0.22\width} {0.05\height} {0.22\width} {0.05\height}},clip]{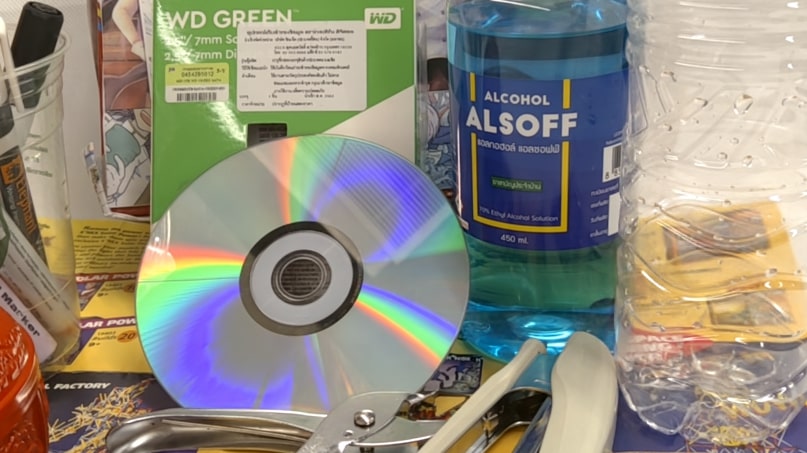}\hfill%
    \adjincludegraphics[width=0.14\linewidth,trim={{0.22\width} {0.05\height} {0.22\width} {0.05\height}},clip]{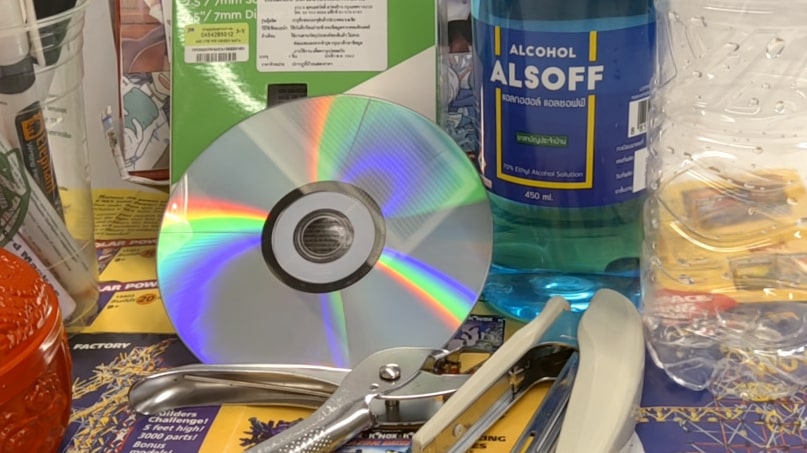}\hfill%
    \adjincludegraphics[width=0.14\linewidth,trim={{0.22\width} {0.05\height} {0.22\width} {0.05\height}},clip]{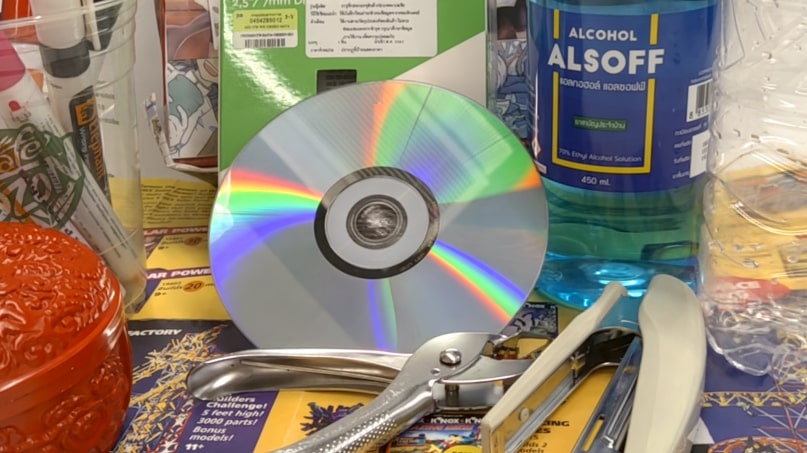}\hfill%
    \adjincludegraphics[width=0.14\linewidth,trim={{0.22\width} {0.05\height} {0.22\width} {0.05\height}},clip]{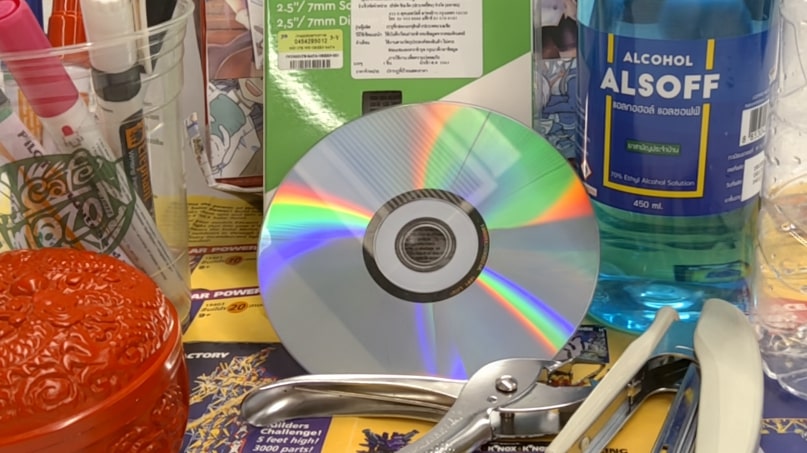}\hfill%
    \adjincludegraphics[width=0.14\linewidth,trim={{0.22\width} {0.05\height} {0.22\width} {0.05\height}},clip]{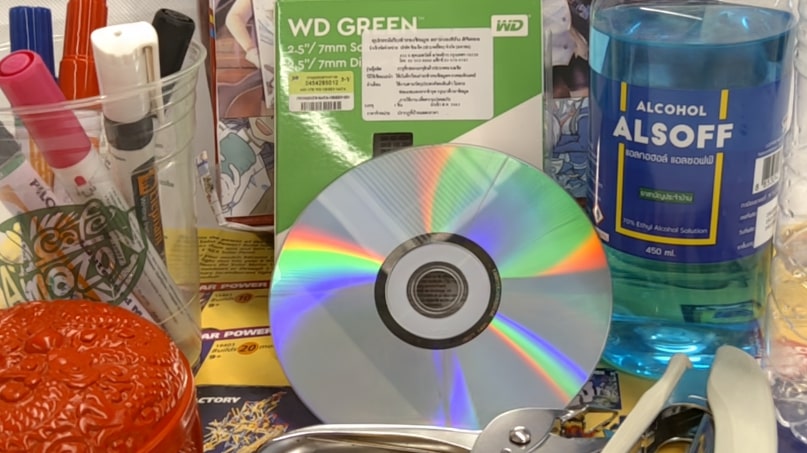}\hfill%
    \adjincludegraphics[width=0.14\linewidth,trim={{0.22\width} {0.05\height} {0.22\width} {0.05\height}},clip]{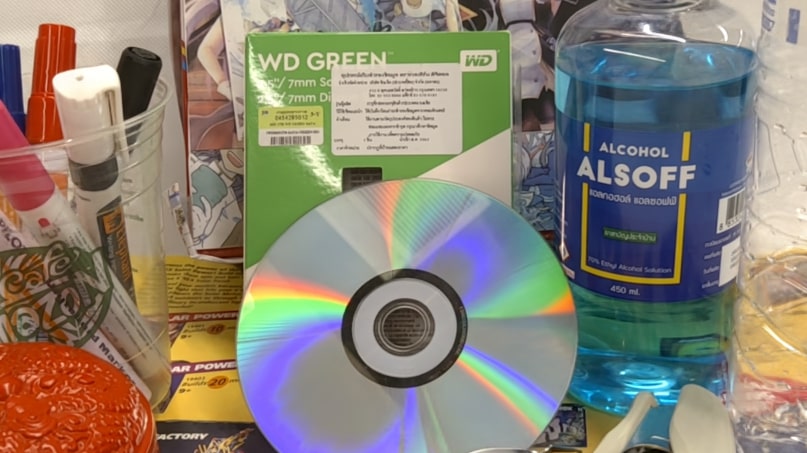}\hfill%
    \adjincludegraphics[width=0.14\linewidth,trim={{0.22\width} {0.05\height} {0.22\width} {0.05\height}},clip]{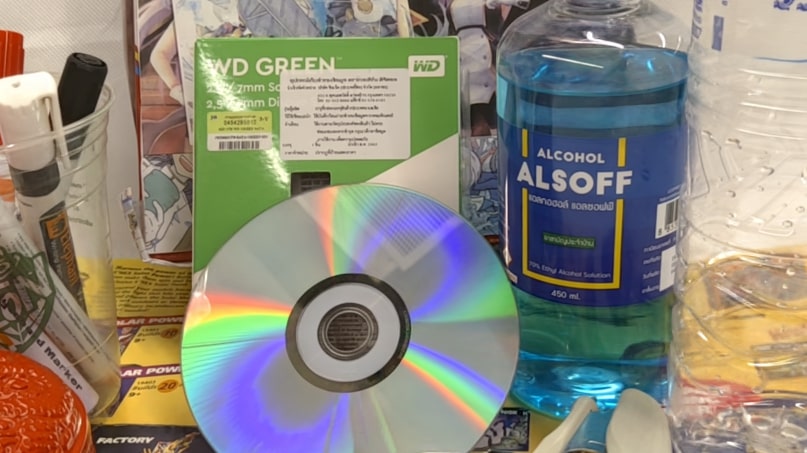}\\[0.5mm]%
    \adjincludegraphics[width=0.14\linewidth,trim={{0.2\width} {0.15\height} {0.2\width} {0.25\height}},clip]{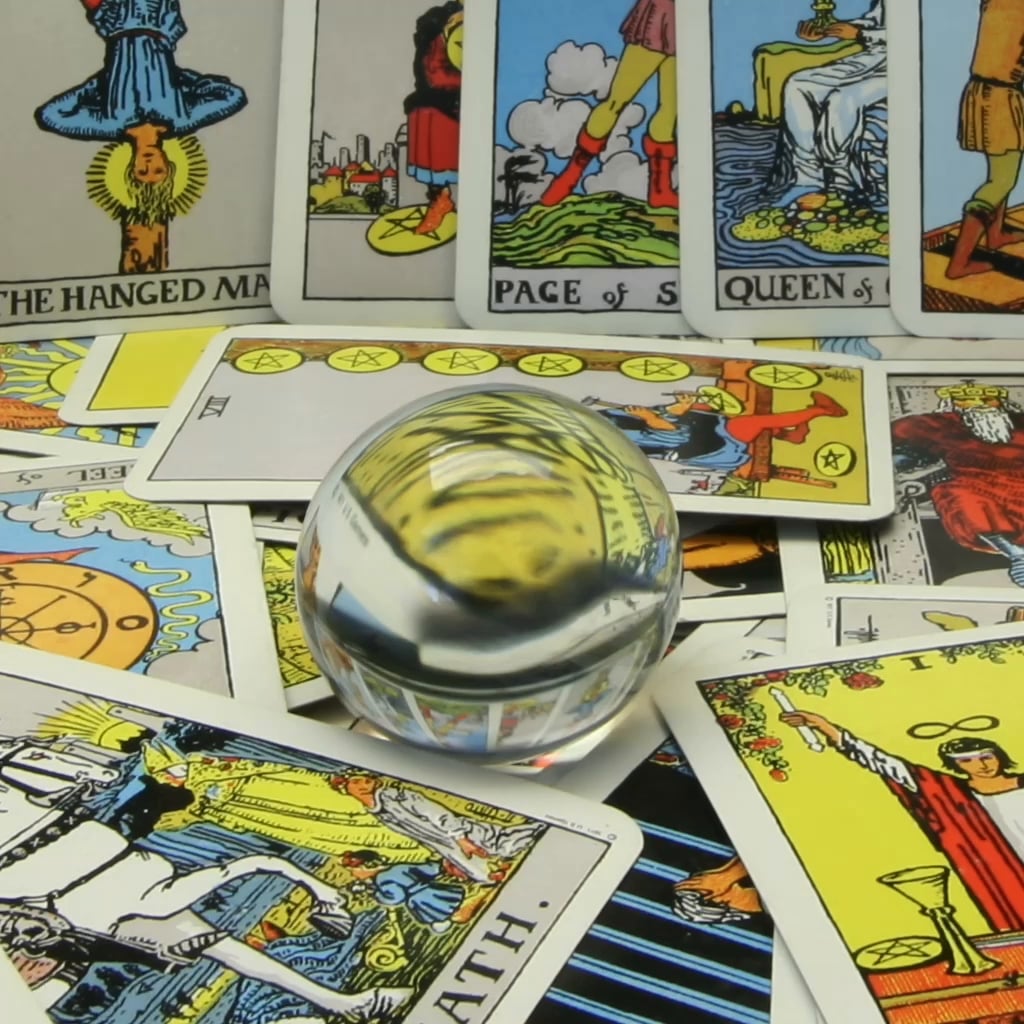}\hfill%
    \adjincludegraphics[width=0.14\linewidth,trim={{0.2\width} {0.15\height} {0.2\width} {0.25\height}},clip]{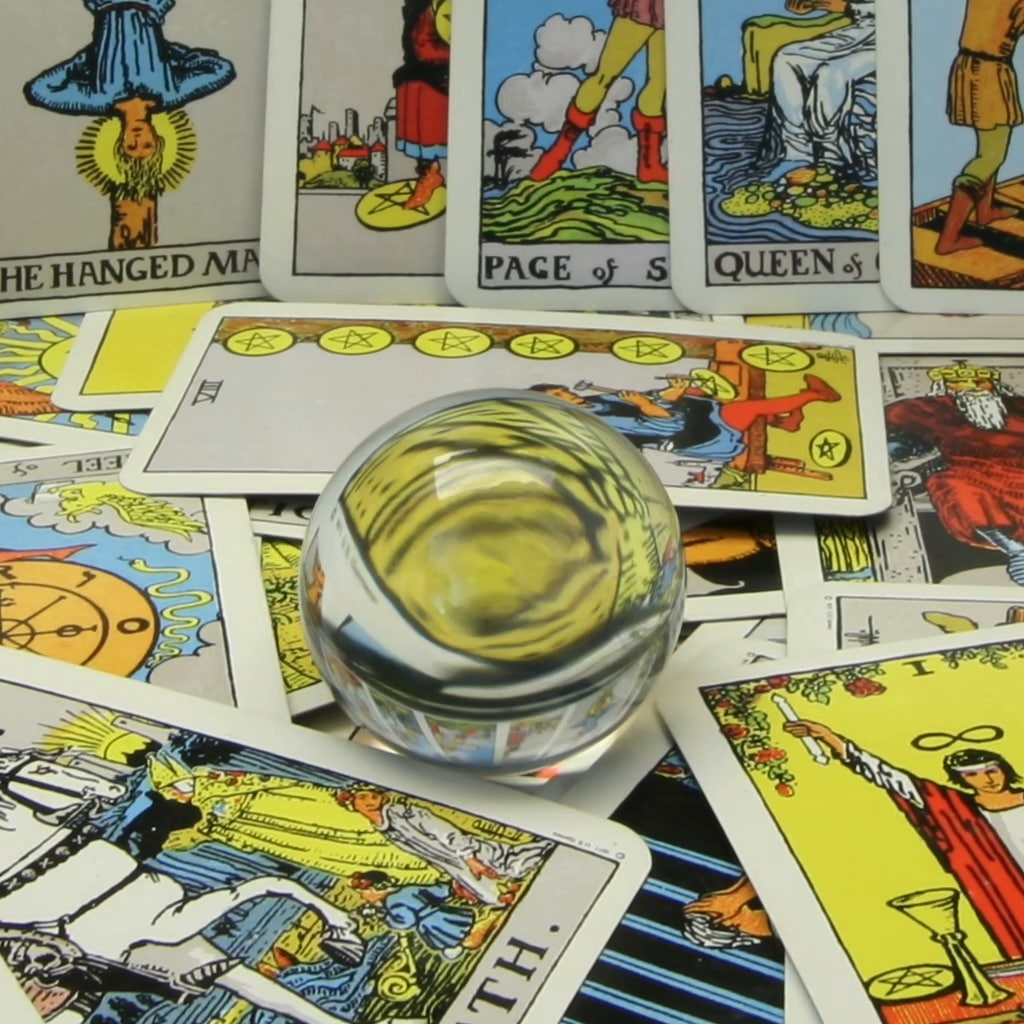}\hfill%
    \adjincludegraphics[width=0.14\linewidth,trim={{0.2\width} {0.15\height} {0.2\width} {0.25\height}},clip]{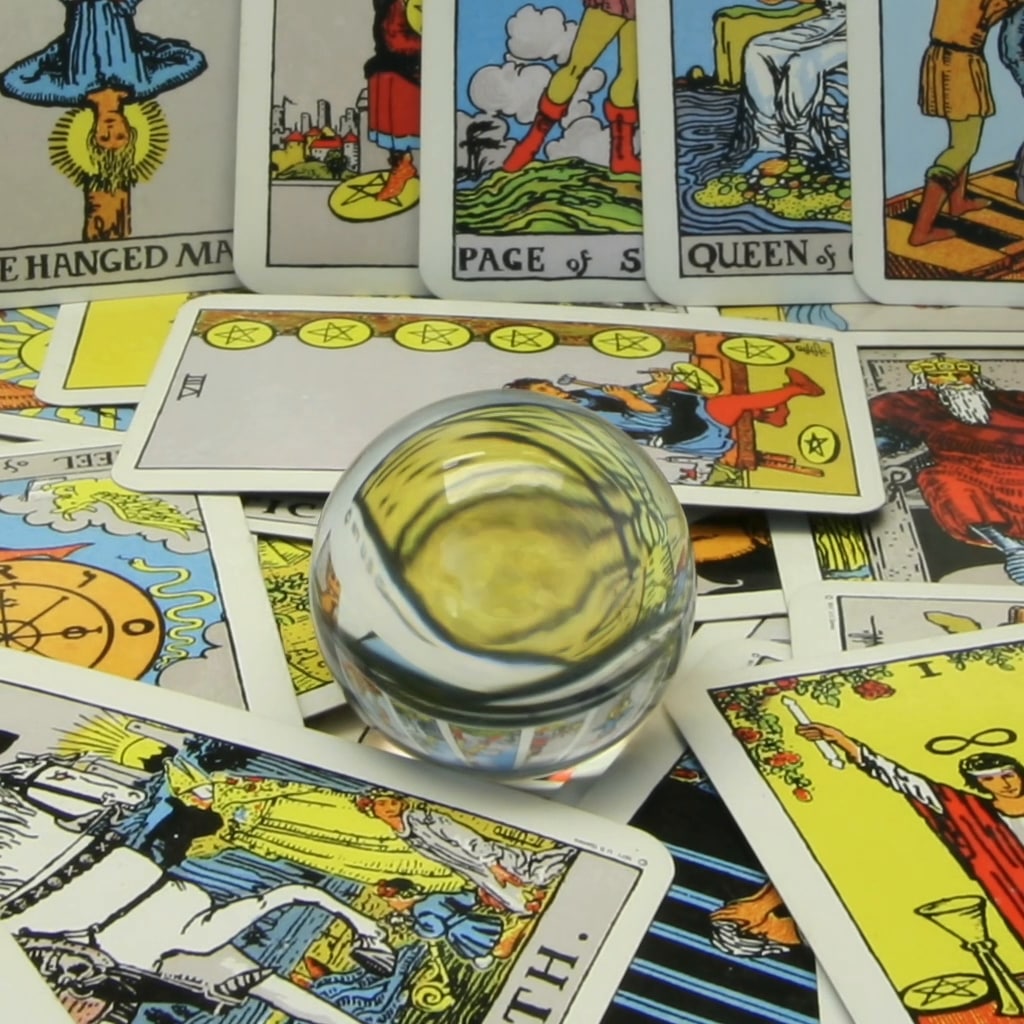}\hfill%
    \adjincludegraphics[width=0.14\linewidth,trim={{0.2\width} {0.15\height} {0.2\width} {0.25\height}},clip]{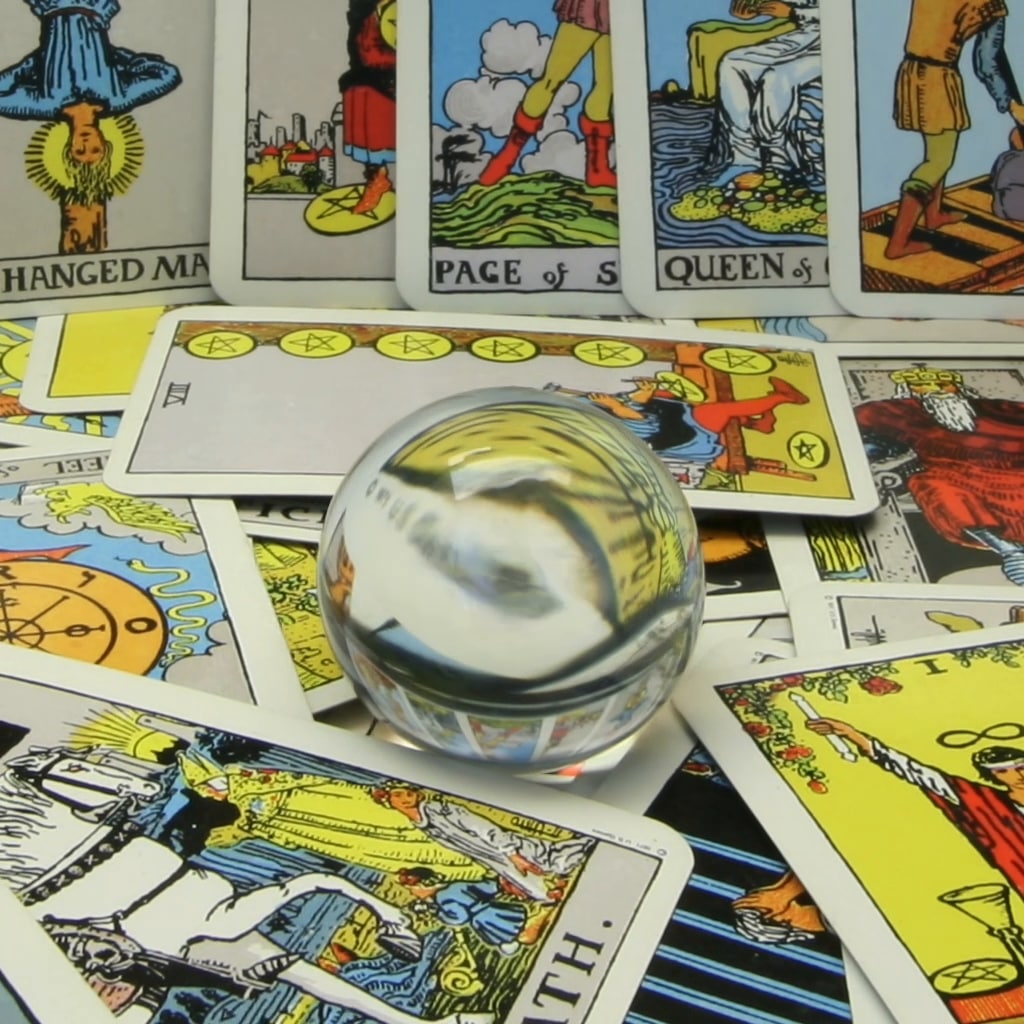}\hfill%
    \adjincludegraphics[width=0.14\linewidth,trim={{0.2\width} {0.15\height} {0.2\width} {0.25\height}},clip]{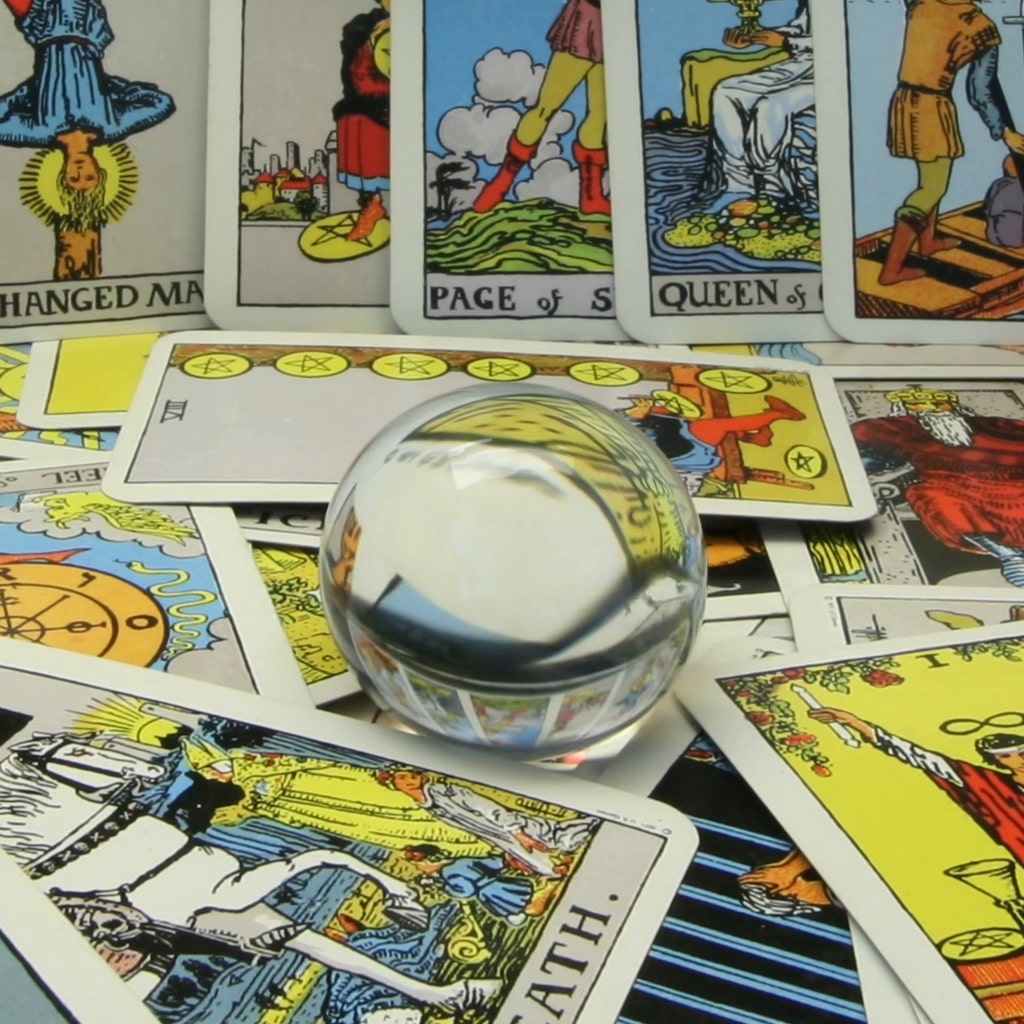}\hfill%
    \adjincludegraphics[width=0.14\linewidth,trim={{0.2\width} {0.15\height} {0.2\width} {0.25\height}},clip]{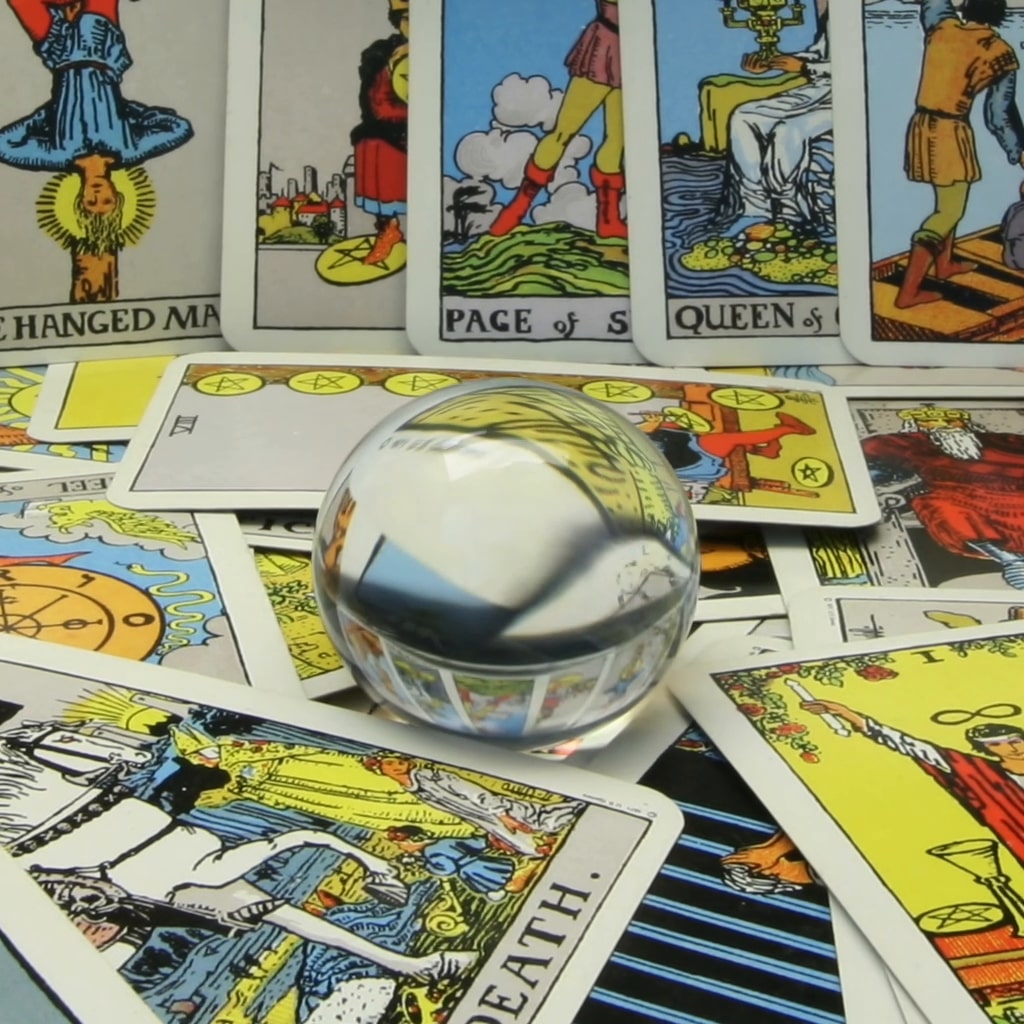}\hfill%
    \adjincludegraphics[width=0.14\linewidth,trim={{0.2\width} {0.15\height} {0.2\width} {0.25\height}},clip]{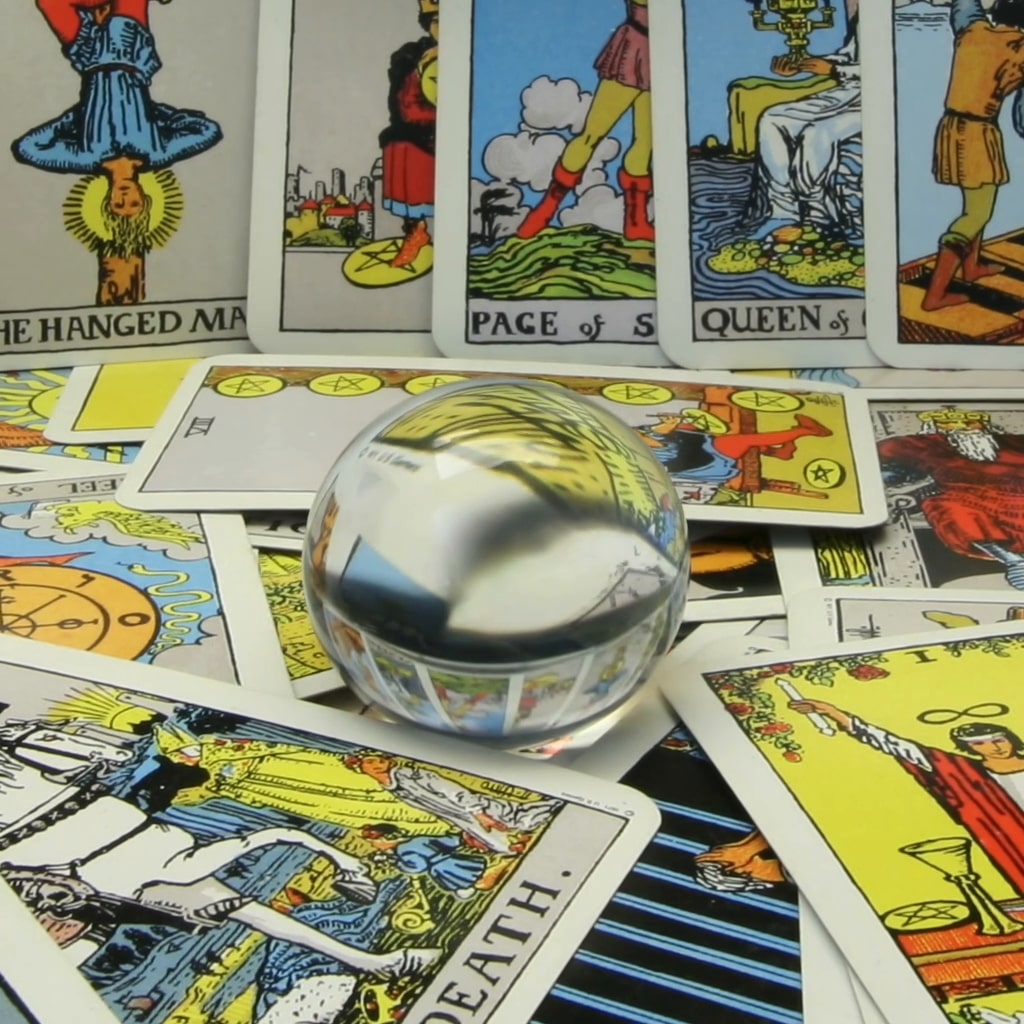}\\[0.5mm]%
  }%
  \\[-1mm]%
\captionof{figure}{\label{fig:teaser}%
\tb{HyperReel: A novel 6-DoF video representation.}
HyperReel converts synchronized multi-view video streams into a high-fidelity, memory efficient scene representation that can be rendered from novel views and time steps at interactive rates.
HyperReel's combination of high rendering quality, speed, and compactness sets it apart from existing 6-DoF video representations.
The upper two rows show 6-DoF (i.e., varying viewpoint and viewing orientation) rendering of dynamic scenes \cite{BroxtFOEHDDBWD2020,sabater2017dataset}; the lower two of static scenes \cite{wizadwongsa2021nex,wilburn2005high}.
}%
\vspace{1em}
}]

\maketitle

\begin{abstract}%
Volumetric scene representations enable photorealistic view synthesis for static scenes and form the basis of several existing 6-DoF video techniques.
However, the volume rendering procedures that drive these representations necessitate careful trade-offs in terms of quality, rendering speed, and memory efficiency. 
In particular, existing methods fail to simultaneously achieve real-time performance, small memory footprint, and high-quality rendering for challenging real-world scenes. 
To address these issues, we present HyperReel --- a novel 6-DoF video representation.
The two core components of HyperReel are: 
(1) a ray-conditioned sample prediction network that enables high-fidelity, high frame rate rendering at high resolutions and 
(2) a compact and memory-efficient dynamic volume representation. 
Our 6-DoF video pipeline achieves the best performance compared to prior and contemporary approaches in terms of visual quality with small memory requirements, while also rendering at up to 18 frames-per-second at megapixel resolution without any custom CUDA code. 

\end{abstract}

\section{Introduction}
\label{sec:introduction}

Six--Degrees-of-Freedom (6-DoF) videos allow for free exploration of an environment by giving the users the ability to change their head position (3 degrees of freedom) and orientation (3 degrees of freedom). 
As such, 6-DoF videos offer immersive experiences with many exciting applications in AR/VR.
The underlying methodology that drives 6-DoF video is \emph{view synthesis}: the process of rendering new, unobserved views of an environment---static or dynamic---from a set of posed images or videos.
Volumetric scene representations such as neural radiance fields \cite{MildeSTBRN2020} and instant neural graphics primitives \cite{MuelleESK2022} have recently made great strides toward photorealistic view synthesis for static scenes.

While several recent works build dynamic view synthesis pipelines on top of these volumetric representations \cite{XianHKK2021, LiNSW2021, GaoSKH2021, ParkSBBGSB2021, LiSZGLKSLGL2022}, it remains a challenging task to create a 6-DoF video format that can achieve high quality, fast rendering, and a small memory footprint (even given many synchronized video streams from multi-view camera rigs \cite{sabater2017dataset, ParraTFHMSSC2019, BroxtFOEHDDBWD2020}).
Existing approaches that attempt to create \textit{memory-efficient} 6-DoF video can take nearly a minute to render a single megapixel image \cite{LiSZGLKSLGL2022}.
Works that target \textit{rendering speed} and represent dynamic volumes directly with 3D textures require gigabytes of storage even for short video clips \cite{WangZLZZZWXY2022}.
While other volumetric methods achieve memory efficiency \textit{and} speed by leveraging sparse or compressed volume storage for static scenes \cite{MuelleESK2022, ChenXGYS2022}, only contemporary work \cite{LiSWST2022, SongCLCCYXG2023} addresses the extension of these approaches to dynamic scenes.
Moreover, all of the above representations struggle to capture highly view-dependent appearance, such as reflections and refractions caused by non-planar surfaces.

In this paper, we present \emph{HyperReel}, a novel 6-DoF video representation that achieves state-of-the-art quality while being memory efficient and real-time renderable at high resolution.
The first ingredient of our approach is a novel \emph{ray-conditioned sample prediction network} that predicts sparse point samples for volume rendering.
In contrast to existing static view synthesis methods that use sample networks \cite{NeffSPKCKS2021, KurzNLZS2022},
our design is unique in that it both 
(1) \emph{accelerates} volume rendering and at the same time 
(2) \emph{improves} rendering quality for challenging view-dependent scenes.

Second, we introduce a memory-efficient dynamic volume representation that achieves a high compression rate by exploiting the spatio-temporal redundancy of a dynamic scene.
Specifically, we extend Tensorial Radiance Fields~\cite{ChenXGYS2022} to compactly represent a set of volumetric keyframes, and capture intermediate frames with trainable scene flow. 

The combination of these two techniques comprises our high-fidelity 6-DoF video representation, \emph{HyperReel}.
We validate the individual components of our approach and our representation as a whole with comparisons to state-of-the-art sampling network-based approaches for static scenes as well as 6-DoF video representations for dynamic scenes.
Not only does HyperReel outperform these existing works, but it also provides high-quality renderings for scenes with challenging non-Lambertian appearances.
Our system renders at up to 18 frames-per-second at megapixel resolution \emph{without} using any custom CUDA code.

The contributions of our work include the following:
\begin{enumerate}\itemsep0em
    \item
    A novel sample prediction network for volumetric view synthesis that accelerates volume rendering and accurately represents complex view-dependent effects.

    \item
    A memory-efficient dynamic volume representation that compactly represents a dynamic scene.
    
    \item
    HyperReel, a 6-DoF video representation that achieves a desirable trade-off between speed, quality, and memory, while rendering in real time at high resolutions.
\end{enumerate}

\section{Related Work}
\label{sec:relatedwork}

\paragraph{Novel View Synthesis} 
Novel-view synthesis is the process of rendering new views of a scene given a set of input posed images.
Classical image-based rendering techniques use approximate scene geometry to reproject and blend source image content onto novel views \cite{BuehlBMGC2001,ShumCK2007,PenneZ2017}.
Recent works leverage the power of deep learning and neural fields \cite{XieTSLYKTTSS2022} to improve image-based rendering from both structured (e.g., light fields \cite{GortlGSC1996, LevoyH1996}) and unstructured data \cite{BemanMSR2020, SuhaiESM2022}.
Rather than performing image-based rendering, which requires storing the input images, another approach is to optimize some 3D scene representation augmented with appearance information \cite{RichaTW2020}.
Examples of such representations include point clouds \cite{AlievSKUL2020, RakhiALB2022}, voxel grids \cite{LombaSSSLS2019, NguyeLTRY2019, SitzmTHNWZ2019}, meshes \cite{RieglK2020, RieglK2021}, or layered representations like multi-plane \cite{ZhouTFFS2018, MildeSOKRNK2019, FlynnBDDFOST2019} or multi-sphere images \cite{AttalLGRT2020, BroxtFOEHDDBWD2020}.

\paragraph{Neural Radiance Fields}
NeRFs are one such 3D scene representation for view synthesis \cite{MildeSTBRN2020} that parameterize the appearance and density of every point in 3D space with a multilayer perceptron (MLP).
While NeRFs enable high-quality view synthesis at a small memory cost, they do not lend themselves to real-time rendering. 
To render the color of a ray from a NeRF, one must evaluate and integrate the color and opacity of many points along a ray---necessitating, in the case of NeRF, hundreds of MLP evaluations per pixel.
Still, due to its impressive performance for static view synthesis, recent methods build on NeRFs in the quest for higher visual quality, more efficient training, and faster rendering speed \cite{TewarTMSTWLSMLSTNBWZG2022, GaoGHLXL2022}.
Several works improve the quality of NeRFs by
accounting for finite pixels and apertures \cite{BarroMTHMS2021, WuLPLCZ2022},
by enabling application to unbounded scenes \cite{ZhangRSK2020, BarroMVSH2022, YuFTCRK2022}, large scenes~\cite{TanciCYPMSBK2022,meuleman2023localrf} or by modifying the representation to 
allow for better reproduction of challenging view-dependent appearances like  reflections and refractions \cite{GuoKBHZ2022, VerbiHMZBS2022, KopanLRJD2022, BemanMFSR2022}.
One can achieve significant training and inference speed improvements by replacing the deep multilayer perceptron with a feature voxel grid in combination with a small neural network \cite{SunSC2022, MuelleESK2022, ChenXGYS2022} or no network at all \cite{YuFTCRK2022, KarneRWM2022}.
Several other works achieve both fast rendering and memory-efficient storage with tensor factorizations \cite{ChenXGYS2022}, learned appearance codebooks, or quantized volumetric features \cite{TakikETMMJF2022}.

\paragraph{Adaptive Sampling for Neural Volume Rendering}
Other works aim to improve the speed of volumetric representations by reducing the number of volume queries required to render a single ray.
Approaches like DoNeRF \cite{NeffSPKCKS2021}, TermiNeRF \cite{PialaC2021}, and AdaNeRF \cite{KurzNLZS2022} learn weights for each segment along a ray 
as a function of the ray itself, and use these weights for adaptive evaluation of the underlying NeRF. 
In doing so, they can achieve near-real-time rendering. 
NeuSample \cite{FangXWZLT2021} replaces the NeRF coarse network with a module that directly predicts the distance to each sample point along a ray.
Methods like AutoInt \cite{LindeMW2021}, DIVeR \cite{WuLBWF2022}, and neural light fields \cite{AttalHZKK2022, SitzmRFTD2021, LiSLYX2022} learn integrated opacity and color along a small set of ray segments (or just one segment),
requiring only a single network evaluation per segment.
A key component of our framework is a flexible sampling network, which is among one of the few schemes that both \textit{accelerates} volume rendering, and also \textit{improves} volume rendering quality for challenging scenes.

\begin{figure*}
\centering
\includegraphics[width=1.0\linewidth]{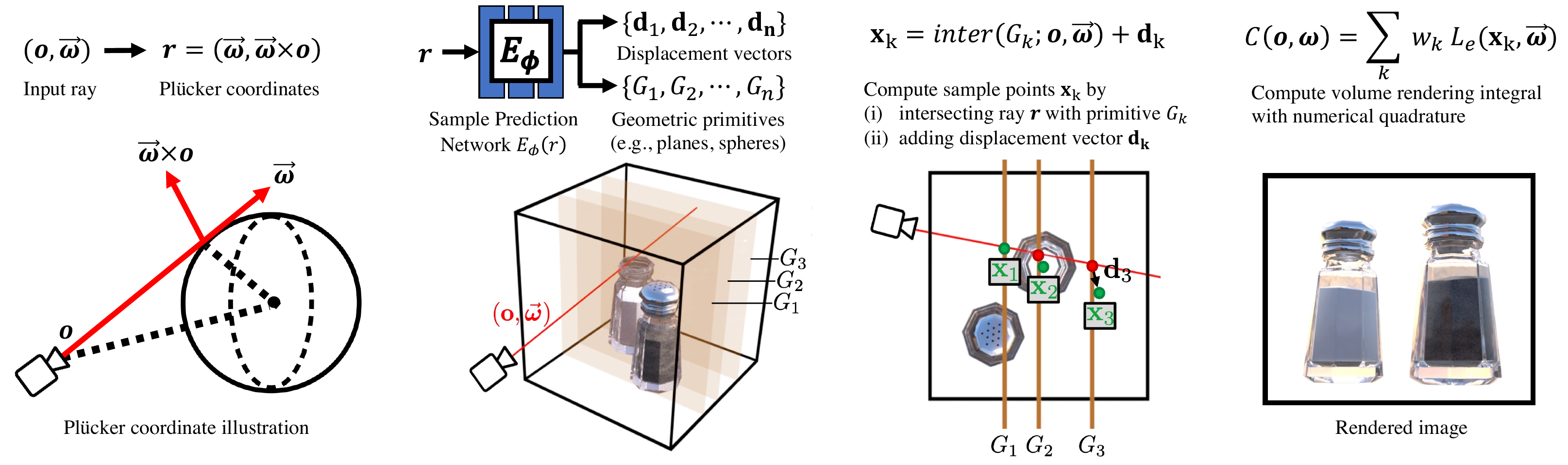}
\mpage{0.25}{(a) Ray parameterization} \hfill
\mpage{0.24}{(b) Sample prediction network} \hfill
\mpage{0.23}{(c) Sample generation} \hfill
\mpage{0.235}{(d) Volume rendering}
\vspace{-5mm}
\caption{\label{fig:overview}%
\tb{Overview of HyperReel for static scenes.}
Given a set of images and camera poses, the training objective is to reconstruct the measured color associated with every ray.
(a) For a ray originating at the camera origin $\mathbf{o}$ and traveling in direction $\vec{\boldsymbol{\omega}}$, we first reparameterize the ray using Plücker coordinates.
(b) A network $E_{\phi}$ takes this ray as input and outputs the parameters for a set of geometric primitives $\{G_k\}$ (such as axis-aligned planes and spheres) and displacement vectors $\{\mathbf{d}_k\}$.
(c) To generate sample points $\{\mathbf{x}_k\}$ for volume rendering, we compute the intersections between the ray and the geometric primitives, and add the displacement vectors to the results. 
(d) Finally, we perform volume rendering via \cref{eqn:quadrature} to produce a pixel color and supervise training based on the corresponding observation.
}
\vspace{-5mm}
\end{figure*}

\paragraph{6--Degrees-of-Freedom Video}
6-DoF video is an emergent technology that allows users to explore new views within videos \cite{RichaTW2020}. 
Systems for 6-DoF video \cite{ParraTFHMSSC2019} use multi-view camera rigs that capture a full 360-degree field of view and use variants of depth-based reprojection \cite{SerraKCDGHM2019} for view synthesis at each frame of the video. 
Other methods optimize time-varying multi-sphere images (MSIs) \cite{BroxtFOEHDDBWD2020, AttalLGRT2020}, which can provide better visual quality but at a higher training cost.

\paragraph{6-DoF from Monocular Captures}
Due to the success of neural radiance fields for static view synthesis, many recent approaches attempt to extend volumetric scene representations to dynamic scenes.
Several such works reconstruct 6-DoF video from single-view (i.e. monocular) RGB sequences \cite{LiNSW2021,GaoSKH2021,ParkSBBGSB2021,liu2023robust}. 
This is a highly under-constrained setting, which requires decoupling camera and object motion. 
The natural signal priors provided by neural radiance fields help during reconstruction. However, most methods typically rely on additional priors, such as off-the-shelf networks for predicting scene flow and geometry or depth from ToF cameras \cite{XianHKK2021, AttalLGKRTO2021}. 
Still, other approaches model the scene at different time steps as smoothly ``warped'' copies of some canonical frame \cite{ParkSBBGSB2021, PumarCPM2021}, which works best for small temporal windows and smooth object motion. 

\paragraph{6-DoF from Multi-View Captures}
Other methods, like ours, aim to produce 6-DoF video from multi-view camera rigs \cite{LombaSSSLS2019, BroxtFOEHDDBWD2020, LiSZGLKSLGL2022}. 
Despite the additional constraints provided by multiple cameras, this remains a challenging task; an ideal 6-DoF video format must simultaneously achieve high visual quality, rendering speed, and memory efficiency.
Directly extending recent volumetric methods to dynamic scenes can achieve high quality and rendering speed \cite{WangZLZZZWXY2022}, but at the cost of substantial memory requirements, potentially gigabytes of memory \cite{YuFTCRK2022} for each video frame.
Contemporary works such as StreamRF \cite{LiSWST2022} and NeRFPlayer \cite{SongCLCCYXG2023} design volumetric 6-DoF video representations that mitigate storage requirements but sacrifice either rendering speed or visual quality.
On the other hand, our approach achieves both fast and high-quality 6-DoF video rendering while maintaining a small memory footprint.

\section{Method}
\label{sec:technical}

We start by considering the problem of optimizing a volumetric representation for static view synthesis.
Volume representations like NeRF \cite{MildeSTBRN2020} model the density and appearance of a static scene at every point in the 3D space.
More specifically, a function $\network: (\xpos, \dirout) \rightarrow (L_\text{e}(\xpos,\dirout), \sigma(\xpos))$
maps position $\xpos$ and direction $\dirout$ along a ray
to a color $L_\text{e}(\xpos, \dirout)$ and density $\sigma(\xpos)$.
Here, the trainable parameters $\boldsymbol\theta$ may be neural network weights, $N$-dimensional array entries, or a combination of both.

We can then render new views of a static scene with
\begin{align}
    C(\mathbf{o}, \dirout) = \int_\tnear^\tfar \!\!\!
    \underbrace{T\!\left(\mathbf{o}, \xpos_t\right)}_\text{\scriptsize Transmittance} \,
    \underbrace{\sigma\!\left(\xpos_t\right)}_\text{\scriptsize Density} \,
    \underbrace{L_\text{e}\!\left(\xpos_t, \dirout\right)}_\text{\scriptsize Radiance} \, \mathrm{d}t \text{,}
\label{eqn:volume_rendering}
\end{align}
where $T\left(\mathbf{o}, \xpos_t\right)$ denotes the transmittance from $\mathbf{o}$ to $\xpos_t$.

In practice, we can evaluate \cref{eqn:volume_rendering} using numerical quadrature by taking many sample points along a given ray:
\begin{align}
    C(\mathbf{o}, \dirout) \approx \sum_{k = 1}^{N} w_k \,
    L_\text{e}\!\left(\xpos_k, \dirout \right) \text{,}
\label{eqn:quadrature}
\end{align}
where the weights $w_k = \hat{T}\left(\mathbf{o}, \xpos_k\right) (1 - e^{-\sigma(\xpos_k) \Delta\xpos_k})$ specify the contribution of each sample point's color to the output.

\subsection{Sample Networks for Volume Rendering}
\label{sec:sampling}

Most scenes consist of solid objects whose surfaces lie on a 2D manifold within the 3D scene volume.
In this case, only a small set of sample points contributes to the rendered color for each ray.
To accelerate volume rendering, we would like to query color and opacity only for points with non-zero $w_k$.
While most volume representations use importance sampling and pruning schemes that help reduce sample counts, they often require hundreds or even thousands of queries per ray to produce accurate renderings \cite{ChenXGYS2022, MuelleESK2022}.

As shown in \cref{fig:overview}, we use a feed-forward network to predict a set of sample locations $\xpos_k$.
Specifically, we use a \emph{sample prediction network} $E_{\boldsymbol\phi} : \left(\mathbf{o}, \dirout\right) \rightarrow \left(\xpos_1, \dots, \xpos_n \right)$ that maps a ray $(\mathbf{o}, \dirout)$ to the sample points $\xpos_k$ for volume rendering in \cref{eqn:quadrature}. 
We use either the two-plane parameterization~\cite{LevoyH1996} (for forward facing scenes) or the Plücker parameterization (for all other scenes) to represent the ray:
\begin{align}
    \mathbf{r} = \textit{Pl\"{u}cker}(\mathbf{o}, \dirout) = \left( \dirout, \dirout \times \mathbf{o} \right) \text{.}
\end{align}
While many designs for the sample prediction network $E_{\boldsymbol\phi}$ are possible, giving the network too much flexibility may negatively affect view synthesis quality.
For example, if $\left(\xpos_1, \dots, \xpos_n \right)$ are completely arbitrary points, then renderings may not appear to be multi-view-consistent.

To address this problem, we choose to predict the parameters of a set of geometric primitives $G_1, \dots, G_n$ defined in the world coordinate frame, where the primitive parameters themselves are a function of the input ray.
To get our sample points, we then intersect the ray with each primitive:
\begin{align}
    \!\!\!\!E_{\boldsymbol\phi}(\mathbf{o}, \dirout) &= \left( G_1, \dots, G_n\right) \, , \\
    \!\!\!\!\left( \xpos_1, \dots, \xpos_n\right) &= \left(\textit{inter}(G_1; \mathbf{o}, \dirout), \dots,  \textit{inter}(G_n; \mathbf{o}, \dirout) \right) \!\text{.}
\end{align}
Above, $\textit{inter}(G_k; \mathbf{o}, \dirout)$ is a differentiable operation that intersects the ray with the primitive $G_k$. 
In all of our experiments, we use axis-aligned $z$-planes (for forward-facing scenes) or concentric spherical shells centered at the origin (for all other scenes) as our geometric primitives.

This approach is constrained in that it produces sample points that initially lie along the ray. Further, predicting primitives defined in world space makes the sample signal easier to interpolate. 
For example, if two distinct rays observe the same point in the scene, then the sample network needs only predict one primitive for both rays (i.e., defining a primitive that passes through the point). 
In contrast, existing works such as NeuSample \cite{FangXWZLT2021}, AdaNeRF \cite{KurzNLZS2022}, and TermiNeRF \cite{PialaC2021} predict distances or per-segment weights that do not have this property.

\paragraph{Flexible Sampling for Challenging Appearance.} To grant our samples additional flexibility to better represent challenging view-dependent appearance, we also predict a set of Tanh-activated per-sample-point offsets $\left(\mathbf{e}_1, \dots, \mathbf{e}_n \right)$, as well as a set of scalar values $\left(\delta_1, \dots, \delta_n \right)$. 
We convert these scalar values to weights with a sigmoid activation, \ie, $\left(\gamma(\delta_1), \dots, \gamma(\delta_n) \right)$ where $\gamma$ is the sigmoid operator. 
Specifically, we have:
\begin{align}
    \left( \mathbf{d}_1, \dots \mathbf{d}_n\right) &= \left(\gamma(\delta_1) \mathbf{e}_1, \dots, \gamma(\delta_n) \mathbf{e}_n \right) \\
    \left( \xpos_1, \dots \xpos_n\right) &\leftarrow \left( \xpos_1 + \mathbf{d}_1, \dots, \xpos_n + \mathbf{d}_n \right) \text{,}
    \label{eqn:point-offsets}
\end{align}
where we use $\left( \mathbf{d}_1, \dots, \mathbf{d}_n\right)$ to denote the final displacement, or ``point-offset'' added to each point.

\begin{figure}
\centering
\includegraphics[trim={0 8cm 12cm 5cm},width=\linewidth]{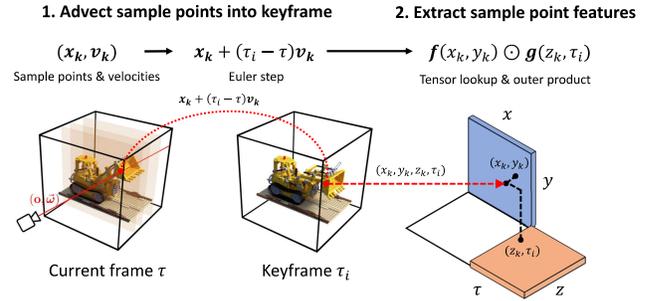}
\caption{\textbf{Extracting sample point appearance and opacity in the dynamic setting from our keyframe-based representation.} (1) We first advect the sample points $\{\mathbf{x}_k\}$ at time $\tau$ into the nearest keyframe $\tau_i$, using velocities $\{\mathbf{v}_k\}$ from the sample prediction network. 
(2) We then query the outer products of space-time textures in order to produce per-sample-point appearance and opacity features, which are converted to colors/densities via \cref{eqn:tensorappearance}.}\label{fig:dynamic}
\vspace{-5mm}
\end{figure}

While the sample network outputs may appear to be over-parameterized and under-constrained, this is essential to achieve good-quality view synthesis.
In particular, initializing the scalars $(\delta_1, \dots, \delta_n)$ to negative values, where the sigmoid is close to 0, and its gradient is small, implicitly discourages the network from unmasking the point offsets, while still allowing the network to use them as necessary.

In addition to enabling real-time rendering with low sample counts, one added benefit of our sample network architecture is the improved modeling of complex view-dependent appearance.
For example, distorted refractions break epipolar geometry and appear to change the depth of the refracted content depending on the viewpoint.
As illustrated in \cref{fig:overview}, our sample network, on the other hand, has the flexibility to model sample points that warp depending on viewpoint, similar to flow-based models of scene appearance in IBR~\cite{nieto2017linearizing}

Existing works like Eikonal fields \cite{BemanMFSR2022} can be considered a special case of this sample warping approach; they use physically derived Eikonal constraints to learn ray-conditional warp fields for refractive objects.
Although our sample network is not guaranteed to be physically interpretable, it can handle \textit{both} reflections and refractions. Further, it is far more efficient at inference time and does not require evaluating costly multi-step ODE solvers during rendering.
See \cref{fig:teaser} and our supplemental materials for additional results and comparisons on challenging view-dependent scenes.

\subsection{Keyframe-Based Dynamic Volumes}

So far, we have covered how to efficiently \textit{sample} a 3D scene volume, but have not yet discussed how we \textit{represent} the volume itself.
In the static case, we use memory-efficient Tensorial Radiance Fields (TensoRF) approach (\cref{sec:tensorialradiancefields}), and in the dynamic case we extend TensoRF to a keyframe-based dynamic volume representation (\cref{sec:keyframe}).

\subsubsection{Representing 3D Volumes with TensoRF~\cite{ChenXGYS2022}}
\label{sec:tensorialradiancefields}

Recall that TensoRF factorizes a 3D volume as a set of outer products between functions of one or more spatial dimensions.
Specifically, we can write the set of spherical harmonic coefficients $A\left(\xpos_k\right)$ capturing the appearance of a point $\xpos_k = (x_k, y_k, z_k)$ as:
\begin{align}
    A\left(\xpos_k\right)
    &= \mathcal{B}_1 \!\left( \mathbf{f}_1(x_k, y_k) \odot \mathbf{g}_1(z_k) \right) \nonumber \\
    &+ \mathcal{B}_2 \!\left( \mathbf{f}_2(x_k, z_k) \odot \mathbf{g}_2(y_k) \right) \label{eqn:tensorappearancestatic} \\
    &+ \mathcal{B}_3 \!\left( \mathbf{f}_3(y_k, z_k) \odot \mathbf{g}_3(x_k) \right) \text{.} \nonumber
\end{align}
Above, $\mathbf{f}_j$ and $\mathbf{g}_j$ are vector-valued functions with output dimension $M$, and `$\odot$' is an element-wise product.
In the original TensoRF work \cite{ChenXGYS2022}, the functions $\mathbf{f}_j$ and $\mathbf{g}_j$ are discretized into $M$ different 2D and and 1D arrays, respectively.

Further, $\mathcal{B}_j$ denote linear transforms that map the products of $\mathbf{f}_j$ and $\mathbf{g}_j$ to spherical harmonic coefficients.
The color $L_e(\xpos_k, \dirout)$ for point $\xpos_k$ and direction $\dirout$ is then given by the dot product of the coefficients $A\left(\xpos_k \right)$ and the spherical harmonic basis functions evaluated at ray direction $\dirout$.

Similar to appearance, for density, we have:
\begin{align}
    \sigma(\xpos_k)
    &= \mathbf{1}^\top \left( \mathbf{h}_1(x_k, y_k) \odot \mathbf{k}_1(z_k) \right) \nonumber \\
    &+ \mathbf{1}^\top \left( \mathbf{h}_2(x_k, z_k) \odot \mathbf{k}_2(y_k) \right) \label{eqn:tensoropacitystatic} \\
    &+ \mathbf{1}^\top \left( \mathbf{h}_3(y_k, z_k) \odot \mathbf{k}_3(x_k) \right) \text{,} \nonumber
\end{align}
where $\mathbf{1}$ is a vector of ones, and $\mathbf{h}_j$ and $\mathbf{k}_j$ are vector-valued functions with output dimension $M$.
Given the color $L_e(\xpos_k, \dirout)$ and density $\sigma(\xpos_k)$ for all sample points $\{\xpos_k\}$ along a ray, we can then make use of \cref{eqn:quadrature} to render the final color for that ray.

\subsubsection{Representing Keyframe-Based Volumes}
\label{sec:keyframe}

To handle dynamics, we adapt TensoRF to parameterize volumetric ``keyframes'', or snapshots of a dynamic volume at a set of discrete time steps. 
If we denote $\tau_i$ as the time step corresponding to the $i^\text{th}$ keyframe, we can write:
\begin{align}
    A\left(\xpos_k, \tau_i\right)
    &= \mathcal{B}_1 \!\left( \mathbf{f}_1(x_k, y_k) \odot \mathbf{g}_1(z_k, \tau_i) \right) \nonumber \\
    &+ \mathcal{B}_2 \!\left( \mathbf{f}_2(x_k, z_k) \odot \mathbf{g}_2(y_k, \tau_i) \right) \label{eqn:tensorappearance} \\
    &+ \mathcal{B}_3 \!\left( \mathbf{f}_3(y_k, z_k) \odot \mathbf{g}_3(x_k, \tau_i) \right) \text{,} \nonumber\\[4pt]
    \sigma(\xpos_k, \tau_i)
    &= \mathbf{1}^\top \left( \mathbf{h}_1(x_k, y_k) \odot \mathbf{k}_1(z_k, \tau_i) \right) \nonumber \\
    &+ \mathbf{1}^\top \left( \mathbf{h}_2(x_k, z_k) \odot \mathbf{k}_2(y_k, \tau_i) \right) \label{eqn:tensoropacity} \\
    &+ \mathbf{1}^\top \left( \mathbf{h}_3(y_k, z_k) \odot \mathbf{k}_3(x_k, \tau_i) \right) \text{,} \nonumber
\end{align}
where the only change from \cref{sec:tensorialradiancefields} is that $\mathbf{g}_j$ and $\mathbf{k}_j$ now depend on time, in addition to one spatial dimension. 

We note that the above factorization of the dynamic volume representing \textit{all keyframes in a video} has a similar memory footprint to a static TensoRF for a \textit{single frame}, assuming that the number of keyframes is small relative to the resolution of our spatial dimensions.
In particular, if the spatial resolution of our volume is $(N_x, N_y, N_z)$ and the number of keyframes is $N_t$, then we can store a single component of $\mathbf{f}_1$ with an $N_x \times N_y$ array, and store a single component of $\mathbf{g}_1$ with an $N_z \times N_t$ array.
Because $N_t \ll N_{x/y/z}$, the arrays $\mathbf{g}_j$ do not contribute significantly to the size of the model.

\subsubsection{Rendering from Keyframe-Based Volumes}
\label{sec:samplingandrendering}

In order to combine our sampling procedure (\cref{sec:sampling}) and  keyframe-based volume representation (\cref{sec:keyframe}) to complete our system for 6-DoF video, a few additional modifications are required.
First, since the surfaces in a dynamic scene move over time, the sample points $\{\mathbf{x_k}\}$ should be time dependent. We therefore augment our sample prediction network to take the current time $\tau$ as input.
Second, the  decomposition of the dynamic scene in \cref{sec:keyframe} creates temporal ``snapshots'' of the volume at discrete keyframes $\tau_i$, but we would like to sample the volume at arbitrary times $\tau$. 
To generate the dynamic volume at all intermediate times, we also output velocities $\mathbf{v}_k \in \mathbb{R}^3$ from the sample prediction network, which we use to advect sample points into the nearest keyframe $\tau_i$ with a single forward-Euler step:
\begin{align}
    \xpos_k \leftarrow \xpos_k + \mathbf{v}_k(\tau_i - \tau) \text{.}
    \label{eq:advect}
\end{align}
\noindent \cref{eq:advect} defines a backwards warp with scene flow field $\mathbf{v}_k$ that generates the volume at time $\tau$. The process of warping sample points and querying the keyframe-based dynamic volume is illustrated in \cref{fig:dynamic}.

After querying the keyframe-based volume with $\{\xpos_k\}$, the equation for volume rendering is then:
\begin{align}
    C(\mathbf{o}, \dirout, \tau) = \sum_{k = 1}^{N} w_k \,
    L_\text{e}\left(\xpos_k, \dirout, \tau_i\right) \text{,}
\label{eqn:quadrature2}
\end{align}
where $w_k = \hat{T}\!\left(\mathbf{o}, \xpos_k, \tau_i\right) (1 - e^{-\sigma(\xpos_k, \tau_i) \Delta\xpos_k})$, and $\tau_i$ is the time step corresponding to the closest keyframe to time $\tau$.
This is effectively the same as \cref{eqn:quadrature}, except $C$, $\xpos_k$, $w_k$ and $L_\text{e}$ now depend on the time $\tau$.
The sampling procedure (\cref{sec:sampling}), volume representation (\cref{sec:keyframe}), and rendering scheme for keyframe-based volumes (\cref{sec:samplingandrendering}) comprise our 6-DoF video representation: \emph{HyperReel}.

\subsection{Optimization}
\label{sec:optimization}

We optimize our representation using \textit{only} the training images, and apply total variation and $\ell_1$ sparsity regularization to our tensor components, similar to TensoRF \cite{ChenXGYS2022}:
\begin{align}
    \mathcal{L} &= \mathcal{L}_\text{L2} + w_\text{L1} \mathcal{L}_\text{L1} + w_\text{TV} \mathcal{L}_\text{TV} \quad \text{where} \label{eq:loss} \\
    \mathcal{L}_\text{L2} &= \sum_{\mathbf{o}, \dirout, \tau} \lVert C(\mathbf{o}, \dirout, \tau) - C_\text{GT}(\mathbf{o}, \dirout, \tau) \rVert \text{.}
\end{align}
The loss is summed over training rays and times, and $C_\text{GT}$ represents the ground-truth color for a given ray and time.

We only use a subset of all training rays to make the optimization tractable on machines with limited memory.
In all dynamic experiments, for frame numbers divisible by 4, we alternate between using all training rays and using training rays from images downsampled by a 4$\times$ factor.
For all other instances, we downsample images by an 8$\times$ factor.

\section{Experiments}
\label{sec:experiments}

\paragraph{Implementation Details}
We implement our method in PyTorch \cite{PaszkGMLBCKLGADKYDRTCSFBC2019} and run experiments on a single NVIDIA RTX 3090 GPU with 24\,GB RAM.
Our sample network is a 6-layer, 256-hidden unit MLP with Leaky ReLU activations for both static and dynamic settings. 
Unless otherwise specified, for forward-facing scenes, we predict 32 $z$-planes as our geometric primitives with our ray-conditioned sample prediction network. 
In all other settings, we predict the radii of 32 spherical shells centered at the origin. 
For our keyframe-based volume representation, we use the same space contraction scheme for unbounded scenes as in mip-NeRF 360~\cite{BarroMVSH2022}. We give the $(x, y)$ and $(z, t)$ textures eight components each and four components to all other textures.
For all dynamic datasets, we use every 4th frame as a keyframe.
Further, we split every input video into 50 frame chunks. For each of these chunks, we train a model for approximately 1.5 hours.

\subsection{Comparisons on Static Scenes}

\begin{table}
\caption{\label{tab:quant_static_donerf}%
    \textbf{Static comparisons.}
    We compare our approach to others on the DoNeRF dataset \cite{NeffSPKCKS2021}. See our supplemental material for comparisons on the LLFF dataset \cite{MildeSOKRNK2019}.
    FPS is normalized per megapixel; memory in MB.
}
\vspace{-2mm}
\renewcommand{\arraystretch}{1}
\centering
\resizebox{\linewidth}{!}{%
\begin{tabular}{@{}llcc@{\hspace{0.75\tabcolsep}}r@{}}
  \toprule
  Dataset & Method & PSNR$\uparrow$ & FPS$\uparrow$ & Memory $\downarrow$ \\
  
  \midrule
  
  \multirow{3}{*}{DoNeRF 400$\times$400}
  & \textit{Single sample} & & & \\
  & \ \ R2L~\cite{WangRHOCFT2022} & 35.5 & --- & \bf 23.7 \\
  \cmidrule{2-5}
  & Ours (per-frame) & \textbf{36.7} & \bf 4.0 & 58.8 \\

  \midrule
  
  \multirow{8}{*}{DoNeRF 800$\times$800}
  & \textit{Uniform sampling} & & & \\
  & \ \ NeRF~\cite{MildeSTBRN2020} & 30.9 & 0.3 & \bf 3.8 \\
  & \ \ Instant NGP~\cite{MuelleESK2022} & 33.1 & 3.8 & 64.0 \\
  \cmidrule{2-5}
  & \textit{Adaptive sampling} & & &\\
  & \ \ DoNeRF~\cite{NeffSPKCKS2021} & 30.8 & 2.1 & 4.1 \\
  & \ \ AdaNeRF~\cite{KurzNLZS2022} & 30.9 & \textbf{4.7} & 4.1 \\
  & \ \ TermiNeRF~\cite{PialaC2021} & 29.8 & 2.1 & 4.1 \\
  \cmidrule{2-5}
  & Ours (per-frame) & \textbf{35.1} & 4.0 & 58.8 \\
  \bottomrule
\end{tabular}%
}
\end{table}

\begin{table}
\caption{\label{tab:quant_dynamic}%
    \textbf{Dynamic comparisons.}
    We compare HyperReel to existing 3D video methods. %
    All FPS numbers are for \textit{megapixel} images, and memory is in MB per frame.
    $^1$On the Neural 3D Video dataset \cite{LiSZGLKSLGL2022}, the authors of Neural 3D Video and StreamRF \cite{LiSWST2022} only evaluate their method on the \textit{flame salmon} sequence.
    $^2$StreamRF~\cite{LiSWST2022} does not provide SSIM and LPIPS scores.
}
\vspace{-2mm}
\renewcommand{\arraystretch}{1}
\centering
\resizebox{\linewidth}{!}{%
\begin{tabular}{@{}llcccr@{\hspace{0.75\tabcolsep}}r@{}}
  \toprule
  Dataset & Method & PSNR$\uparrow$ & SSIM$\uparrow$ & LPIPS$\downarrow$ & FPS$\uparrow$ & Memory$\downarrow$ \\
  \midrule
  \multirow{2}{*}{Technicolor~\cite{sabater2017dataset}}
  & Neural 3D Video~\cite{LiSZGLKSLGL2022} & 31.8 & \textbf{0.958} & 0.140 & 0.02 & \bf 0.6 \\
  & Ours & \bf 32.7 & 0.906 & \textbf{0.109} & \textbf{4.00} & 1.2 \\
  \midrule
  \multirow{4}{*}{Neural 3D Video~\cite{LiSZGLKSLGL2022}}
  & Neural 3D Video~\cite{LiSZGLKSLGL2022}$^1$ &     29.6 & \bf 0.961     & \bf 0.083     &      0.02 & \bf  0.1 \\
  & NeRFPlayer~\cite{SongCLCCYXG2023}      &     30.7 & 0.931 & 0.111 &      0.06 &     17.1 \\
  & StreamRF~\cite{LiSWST2022}$^1$             &     28.3 & ---$^2$ & ---$^2$ & \bf 10.90 &     17.7 \\
  & Ours                                   & \bf 31.1 &     0.927     &     0.096     &      2.00 &      1.2 \\
  \midrule
  \multirow{2}{*}{Google LF videos~\cite{BroxtFOEHDDBWD2020}}
  & NeRFPlayer \cite{SongCLCCYXG2023} & 25.8 & 0.848 & 0.196 & 0.12 & 17.1 \\
  & Ours                              & \textbf{28.8} & \textbf{0.874} & \textbf{0.193} &  \textbf{4.00} & \bf 1.2 \\
  \bottomrule
\end{tabular}%
}
\end{table}

\begin{table}
\caption{\label{tab:quant_ablation}%
    \textbf{Network ablations.}
    We perform several ablations on our method, including on the number of keyframes, the use of the sampling network, and model size.
    All FPS numbers per megapixel.
}
\vspace{-2mm}
\renewcommand{\arraystretch}{1}
\centering
\resizebox{\linewidth}{!}{%
\begin{tabular}{@{}llccc@{\hspace{0.75\tabcolsep}}r@{}}
  \toprule
  Dataset & Method & PSNR$\uparrow$ & SSIM$\uparrow$ & LPIPS$\downarrow$ & FPS$\uparrow$ \\
  
  \midrule
  
  \multirow{7}{*}{Technicolor}
  & Ours (keyframe: every  frame)  &     32.34 &     0.895 &      0.117 &      4.0 \\
  & Ours (keyframe: every  4 frames)  & \bf 32.73 & \bf 0.906 & \bf  0.109 &      4.0 \\
  & Ours (keyframe: every 16 frames)  &     32.07 &     0.893 &      0.112 &      4.0 \\
  & Ours (keyframe: every 50 frames)  &     32.35 &     0.896 &      0.110 &      4.0 \\
  \cmidrule{2-6}
  & Ours (w/o sample network) &     29.08 &     0.815 &     0.209 &      1.3 \\
  & Ours (\textit{Tiny})      &     30.09 &     0.835 &     0.157 & \bf 17.5 \\
  & Ours (\textit{Small})     &     31.76 &     0.903 &     0.125 &      9.1 \\
  \bottomrule
\end{tabular}%
}
\vspace{-2mm}
\end{table}

\begin{table}
\caption{\label{tab:quant_point}%
  \textbf{Point offset ablation.}
  We evaluate the performance of our network with and without point offsets.
}
\vspace{-2mm}
\renewcommand{\arraystretch}{1}
\centering
\resizebox{\linewidth}{!}{%
\begin{tabular}{@{}llcc@{\hspace{0.75\tabcolsep}}r@{}}
  \toprule
  Scene & Point offset & PSNR$\uparrow$ & SSIM$\uparrow$ & ~LPIPS$\downarrow$ \\
  \midrule
  DoNeRF \textit{``Forest''}~\cite{NeffSPKCKS2021} & Without  & 34.86 & 0.969 & 0.0146 \\
  \ \ (diffuse)                                        & With     & \bf 36.34 & \bf 0.975 & \bf 0.0122 \\
  \midrule
  Shiny \textit{``Lab''}~\cite{wizadwongsa2021nex} & Without  & 31.28 & 0.943 & 0.0416 \\
  \ \ (highly refractive)                              & With     & \bf 32.49 & \bf 0.959 & \bf 0.0294 \\
  \bottomrule
\end{tabular}%
}
\vspace{-2mm}
\end{table}

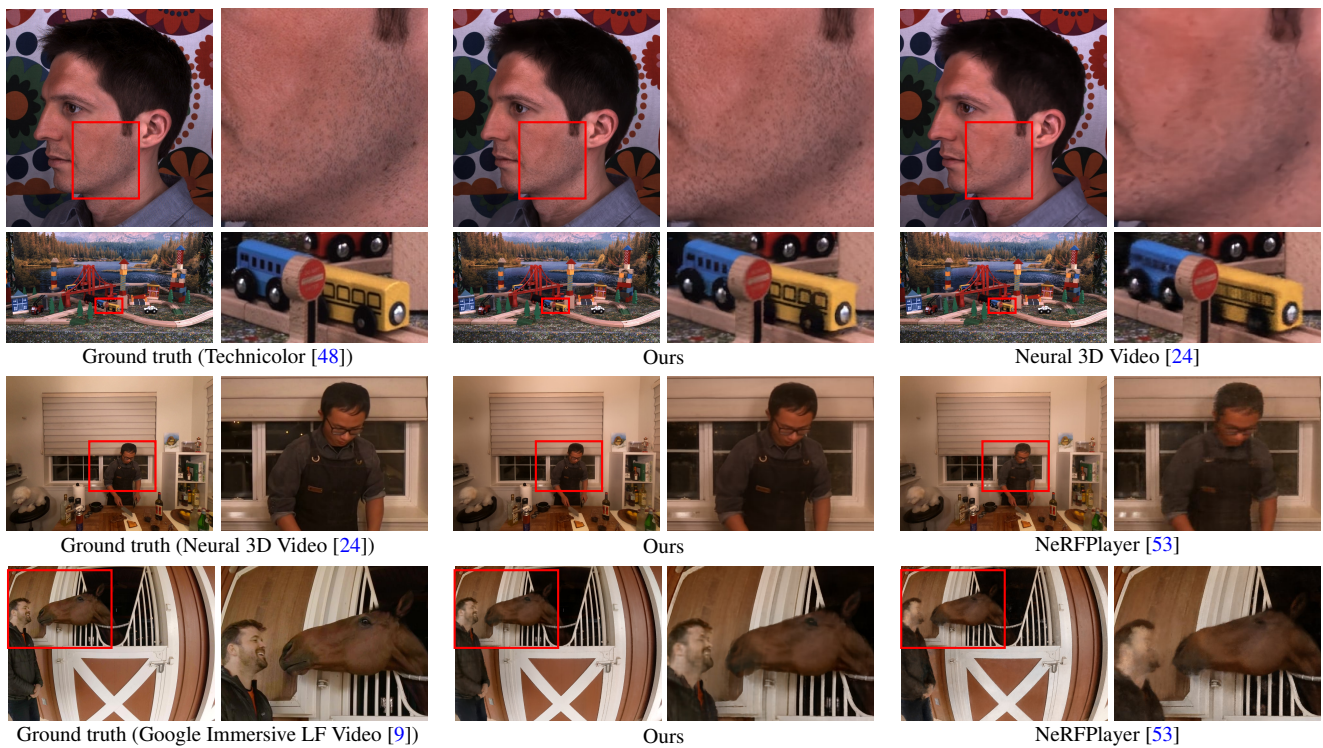
\begin{figure*}[t]
\footnotesize
\centering%
\mpage{0.32}{%
    \mpage{0.49}{%
      \begin{tikzpicture}
        \node[anchor=south west,inner sep=0] (image) at (0,0) {\adjincludegraphics[height=\linewidth,trim={{0.14\width} 0 {0.3\width} 0},clip,angle=90]{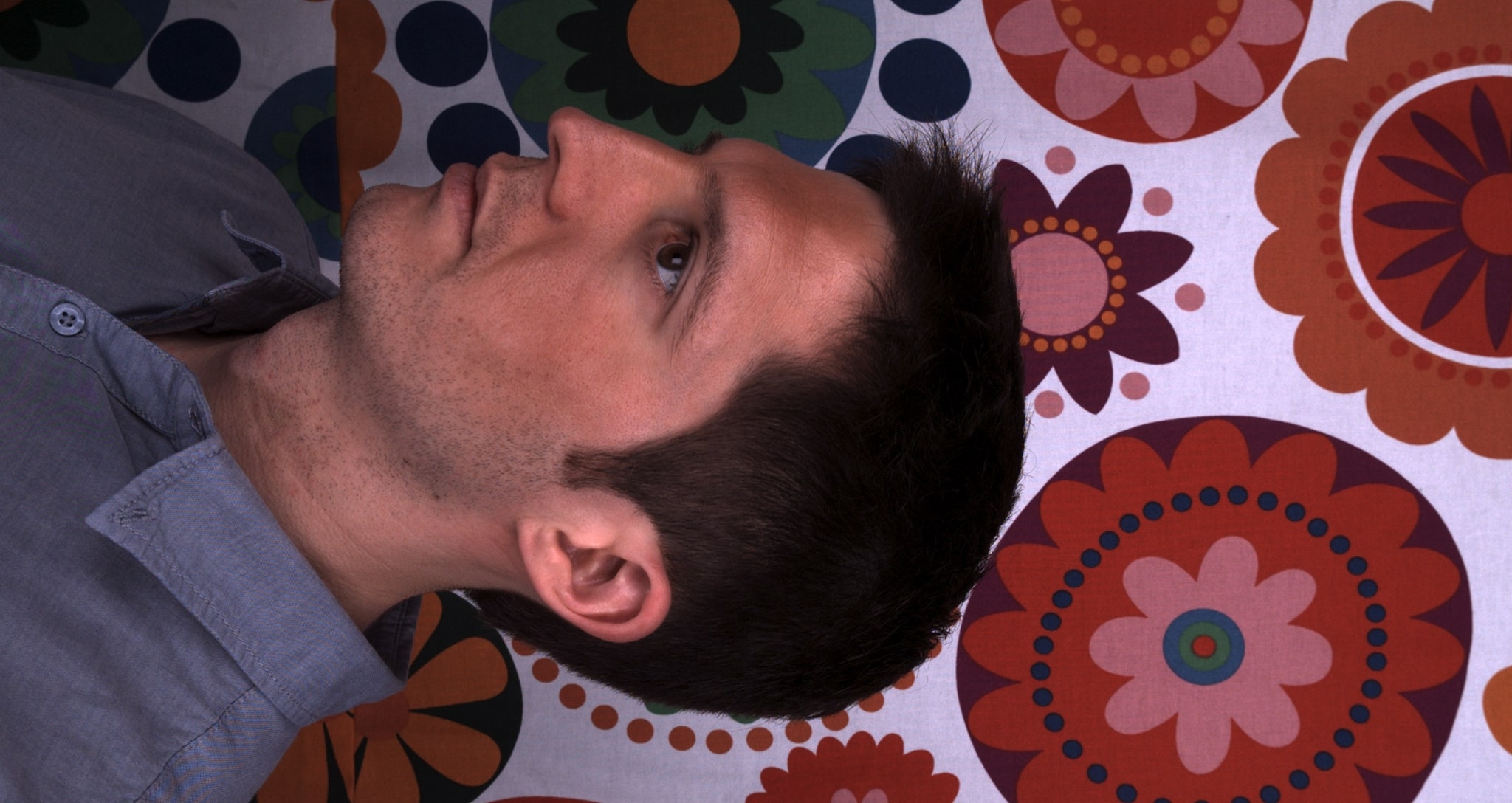}};
        \begin{scope}[x={($1*(image.south east)$)},y={($1*(image.north west)$)}]
          \draw[red,thick] (0.32,0.13) rectangle (0.64,0.48);
        \end{scope}
      \end{tikzpicture}%
    }\hspace{-1mm}\hfill%
    \mpage{0.49}{%
      \adjincludegraphics[height=\linewidth,trim={{0.22\width} {0.36\height} {0.6\width} {0.32\height}},clip,angle=90]{figures/fig_dynamic/img/technicolor1/0002_gt.jpg}\\[0.2mm]%
    }
  }\hfill%
  \mpage{0.32}{%
    \mpage{0.49}{%
      \begin{tikzpicture}
        \node[anchor=south west,inner sep=0] (image) at (0,0) {\adjincludegraphics[height=\linewidth,trim={{0.14\width} 0 {0.3\width} 0},clip,angle=90]{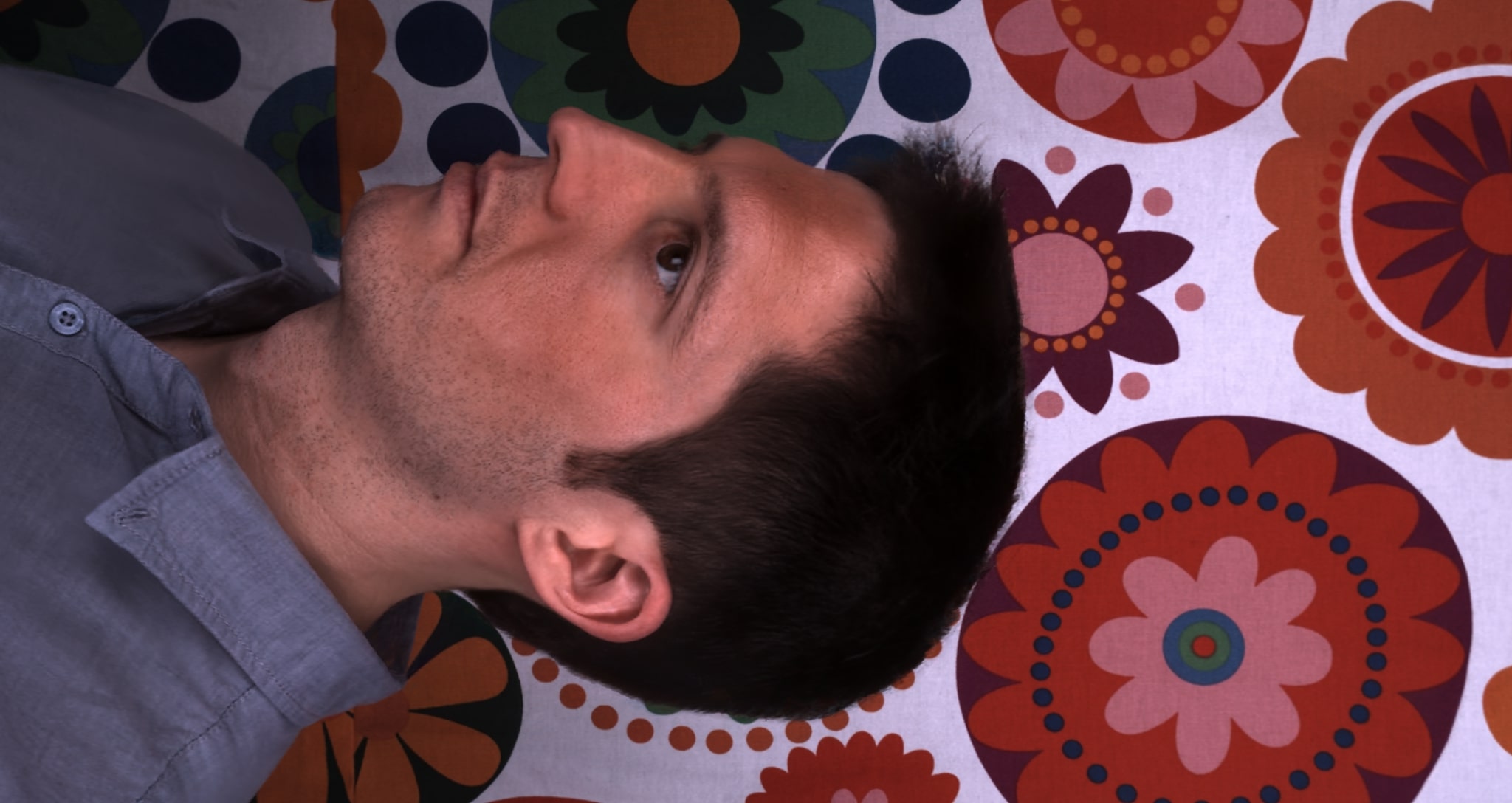}};
        \begin{scope}[x={($1*(image.south east)$)},y={($1*(image.north west)$)}]
          \draw[red,thick] (0.32,0.13) rectangle (0.64,0.48);
        \end{scope}
      \end{tikzpicture}%
    }\hspace{-1mm}\hfill%
    \mpage{0.49}{%
      \adjincludegraphics[height=\linewidth,trim={{0.22\width} {0.36\height} {0.6\width} {0.32\height}},clip,angle=90]{figures/fig_dynamic/img/technicolor1/0002_ours.jpg}\\[0.2mm]%
    }
  }\hfill%
  \mpage{0.32}{%
    \mpage{0.49}{%
      \begin{tikzpicture}
        \node[anchor=south west,inner sep=0] (image) at (0,0) {\adjincludegraphics[height=\linewidth,trim={{0.14\width} 0 {0.3\width} 0},clip,angle=90]{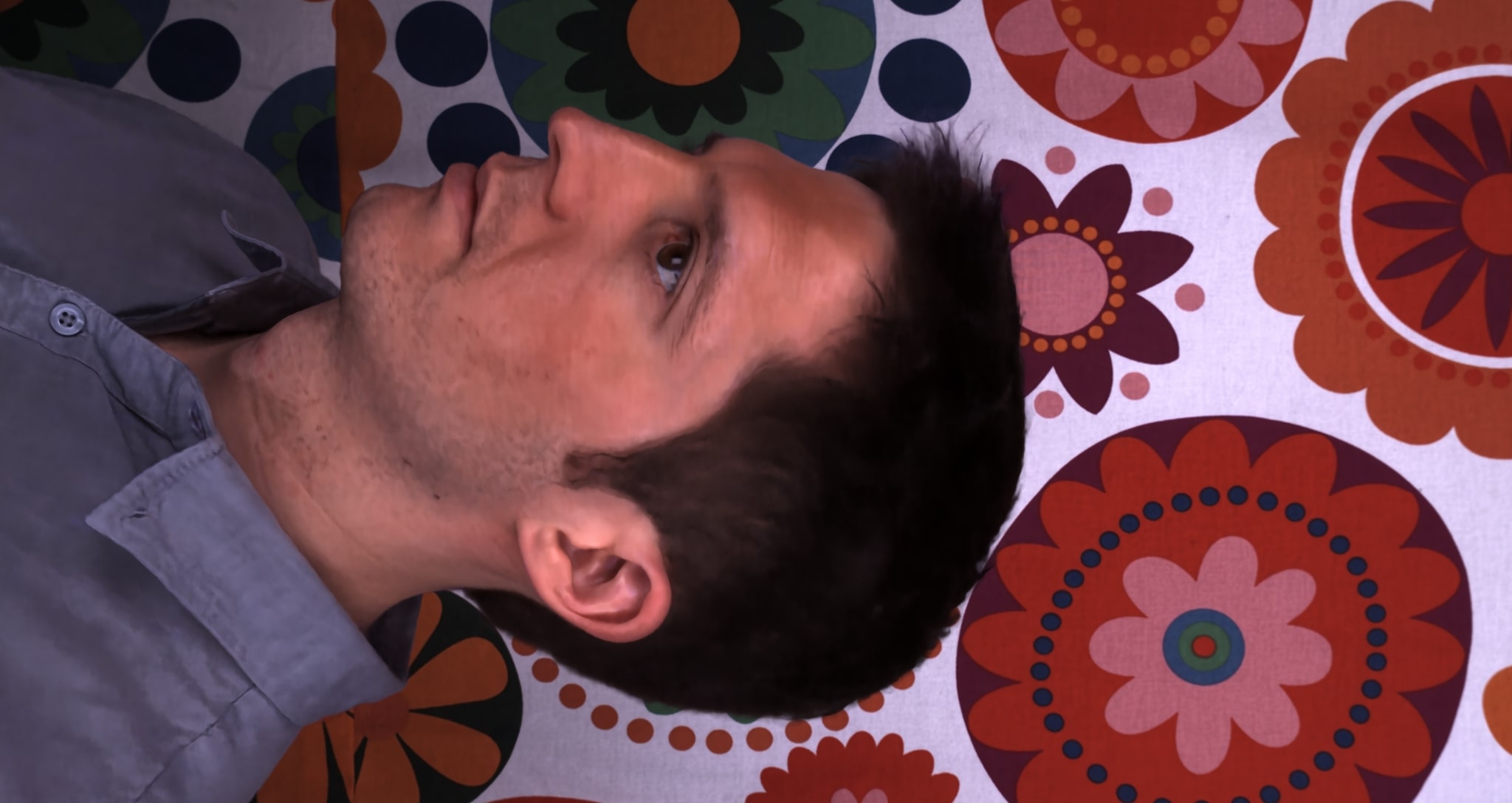}};
        \begin{scope}[x={($1*(image.south east)$)},y={($1*(image.north west)$)}]
          \draw[red,thick] (0.32,0.13) rectangle (0.64,0.48);
        \end{scope}
      \end{tikzpicture}%
    }\hspace{-1mm}\hfill%
    \mpage{0.49}{%
      \adjincludegraphics[height=\linewidth,trim={{0.22\width} {0.36\height} {0.6\width} {0.32\height}},clip,angle=90]{figures/fig_dynamic/img/technicolor1/0002_n3d.jpg}\\[0.2mm]%
    }
  }\\[0.1mm]%

  \mpage{0.32}{%
    \mpage{0.49}{%
      \begin{tikzpicture}
        \node[anchor=south west,inner sep=0] (image) at (0,0) {\includegraphics[width=\linewidth]{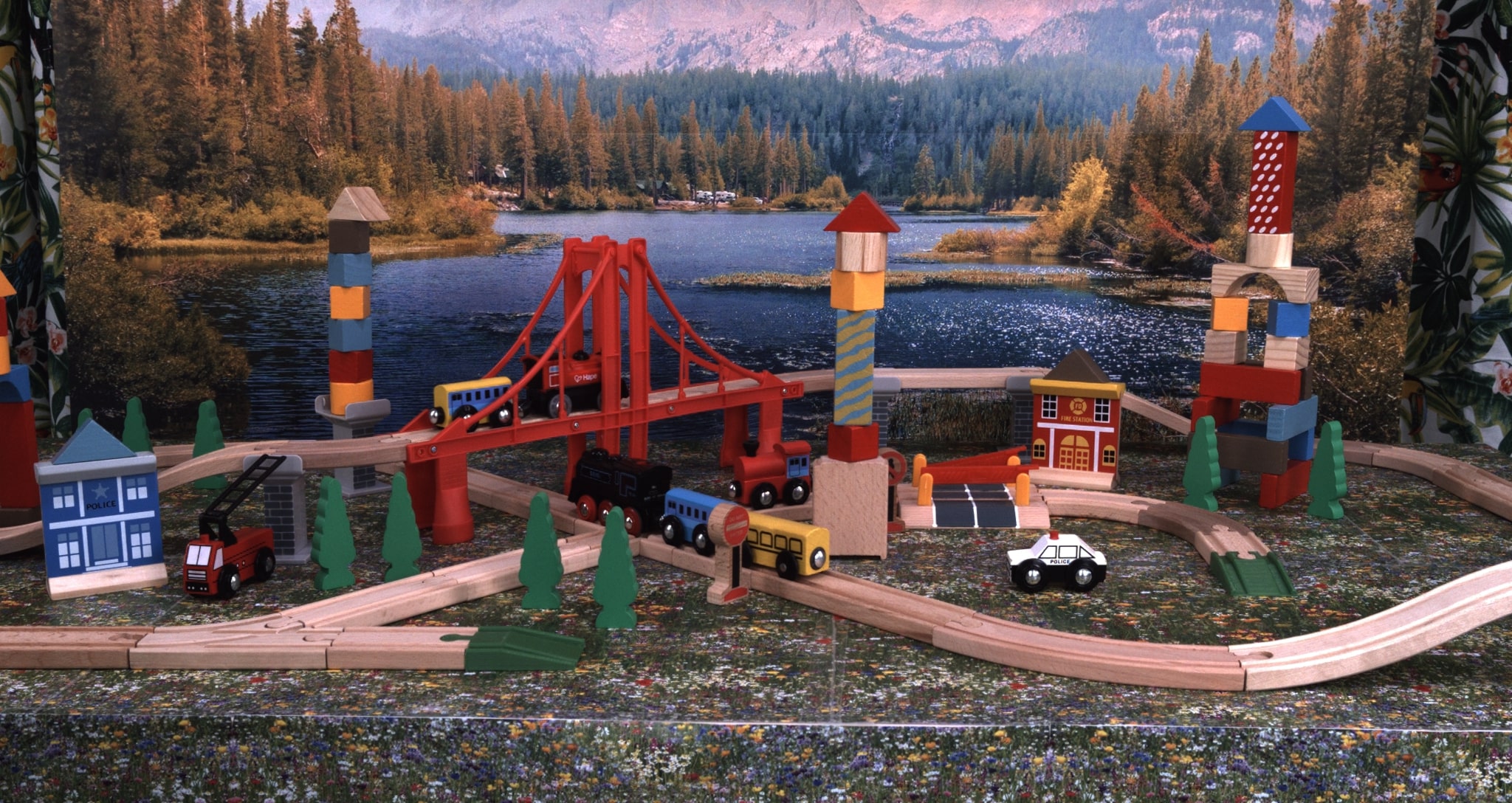}};
        \begin{scope}[x={($1*(image.south east)$)},y={($1*(image.north west)$)}]
          \draw[red,thick] (0.43,0.27) rectangle (0.56,0.4);
        \end{scope}
      \end{tikzpicture}%
    }\hspace{-1mm}\hfill%
    \mpage{0.49}{%
      \adjincludegraphics[width=\linewidth,trim={{0.43\width} {0.27\height} {0.44\width} {0.6\height}},clip]{figures/fig_dynamic/img/technicolor3/0002_gt.jpg}\\[0.2mm]%
    }
  }\hfill%
  \mpage{0.32}{%
    \mpage{0.49}{%
      \begin{tikzpicture}
        \node[anchor=south west,inner sep=0] (image) at (0,0) {\includegraphics[width=\linewidth]{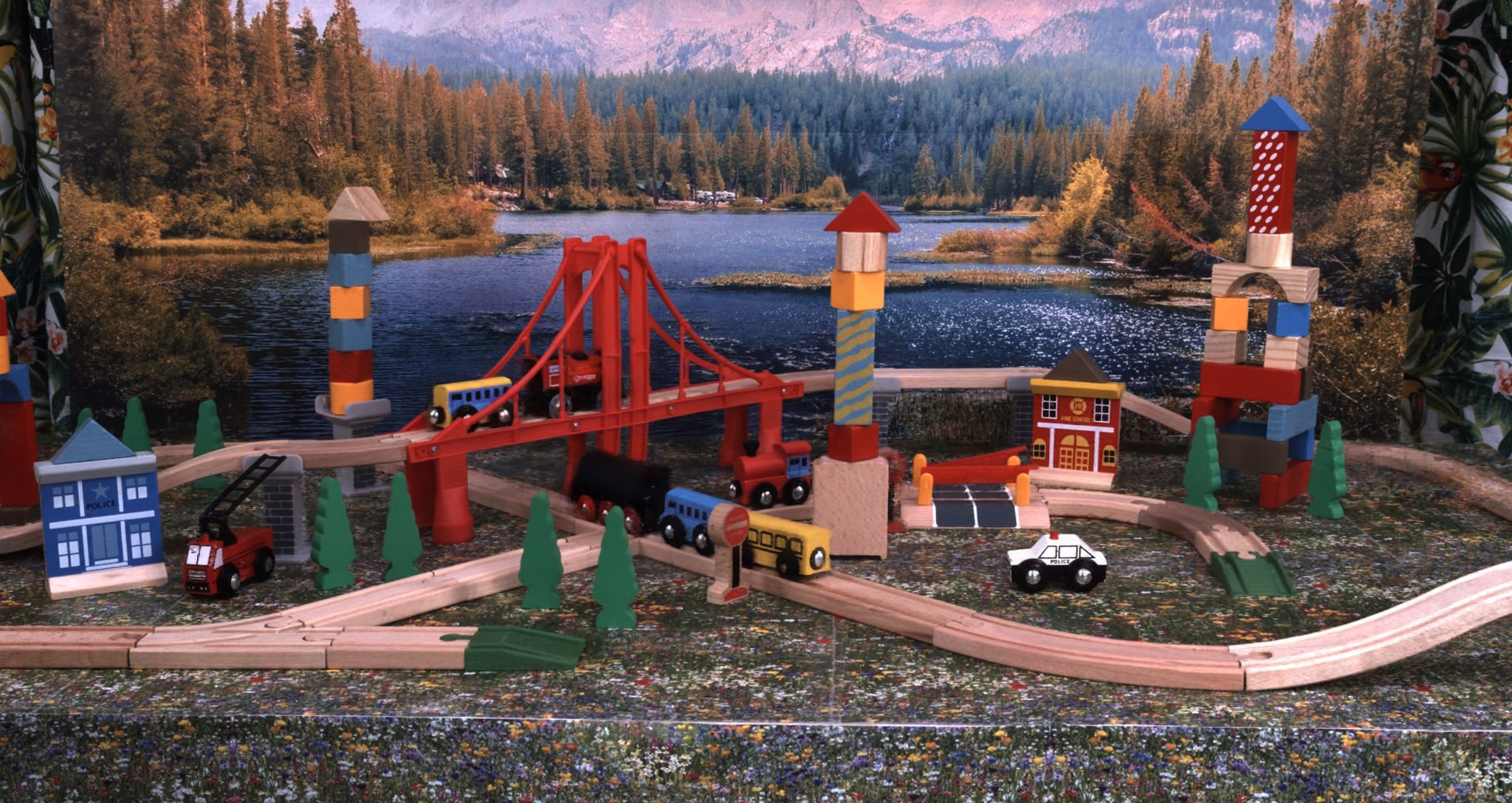}};
        \begin{scope}[x={($1*(image.south east)$)},y={($1*(image.north west)$)}]
          \draw[red,thick] (0.43,0.27) rectangle (0.56,0.4);
        \end{scope}
      \end{tikzpicture}%
    }\hspace{-1mm}\hfill%
    \mpage{0.49}{%
      \adjincludegraphics[width=\linewidth,trim={{0.43\width} {0.27\height} {0.44\width} {0.6\height}},clip]{figures/fig_dynamic/img/technicolor3/0002_ours.jpg}\\[0.2mm]%
    }
  }\hfill%
  \mpage{0.32}{%
    \mpage{0.49}{%
      \begin{tikzpicture}
        \node[anchor=south west,inner sep=0] (image) at (0,0) {\includegraphics[width=\linewidth]{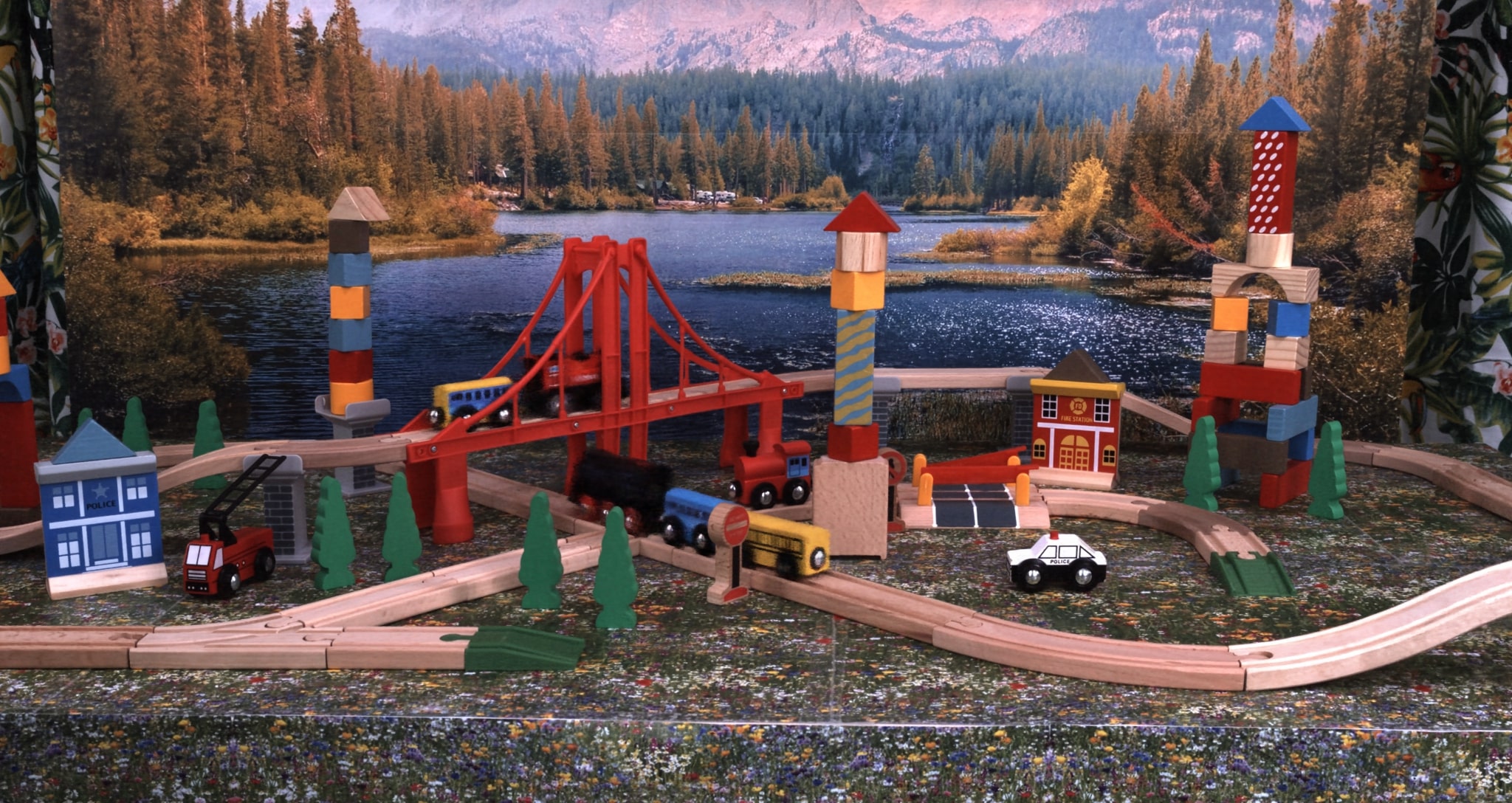}};
        \begin{scope}[x={($1*(image.south east)$)},y={($1*(image.north west)$)}]
          \draw[red,thick] (0.43,0.27) rectangle (0.56,0.4);
        \end{scope}
      \end{tikzpicture}%
    }\hspace{-1mm}\hfill%
    \mpage{0.49}{%
      \adjincludegraphics[width=\linewidth,trim={{0.43\width} {0.27\height} {0.44\width} {0.6\height}},clip]{figures/fig_dynamic/img/technicolor3/0002_n3d.jpg}\\[0.2mm]%
    }
  }\\[0.1mm]%
  \mpage{0.32}{%
    Ground truth (Technicolor~\cite{sabater2017dataset})%
  }\hfill%
  \mpage{0.32}{%
    Ours%
  }\hfill%
  \mpage{0.32}{%
    Neural 3D Video~\cite{LiSZGLKSLGL2022}
  }\\[1mm]%
  \mpage{0.32}{%
    \mpage{0.49}{%
      \begin{tikzpicture}
        \node[anchor=south west,inner sep=0] (image) at (0,0) {\includegraphics[width=\linewidth]{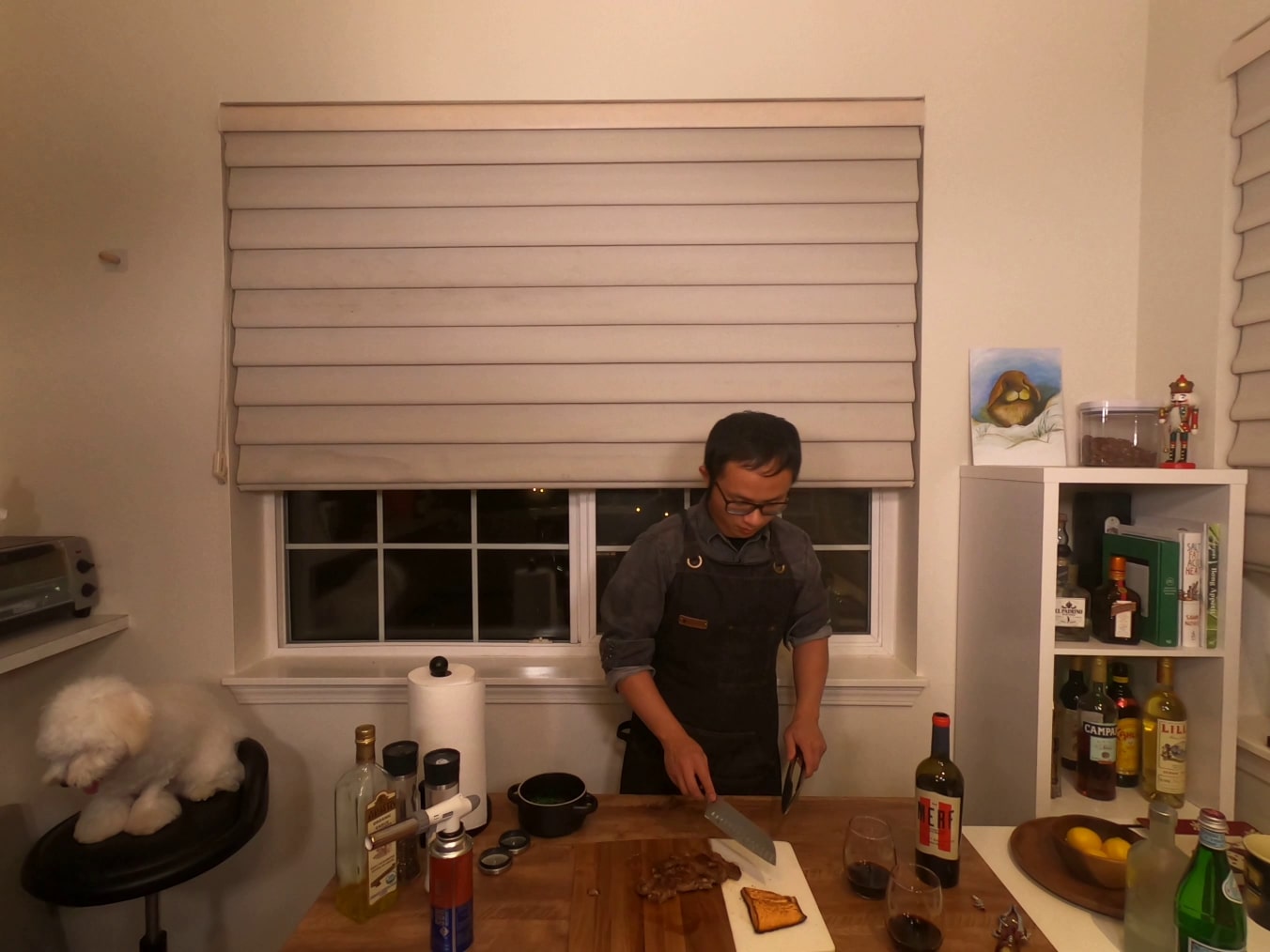}};
        \begin{scope}[x={($1*(image.south east)$)},y={($1*(image.north west)$)}]
          \draw[red,thick] (0.4,0.26) rectangle (0.72,0.58);
        \end{scope}
      \end{tikzpicture}%
    }\hspace{-1mm}\hfill%
    \mpage{0.49}{%
      \adjincludegraphics[width=\linewidth,trim={{0.4\width} {0.26\height} {0.28\width} {0.42\height}},clip]{figures/fig_dynamic/img/n3dv/0010_gt.jpg}\\[0.2mm]%
    }
  }\hfill%
  \mpage{0.32}{%
    \mpage{0.49}{%
      \begin{tikzpicture}
        \node[anchor=south west,inner sep=0] (image) at (0,0) {\includegraphics[width=\linewidth]{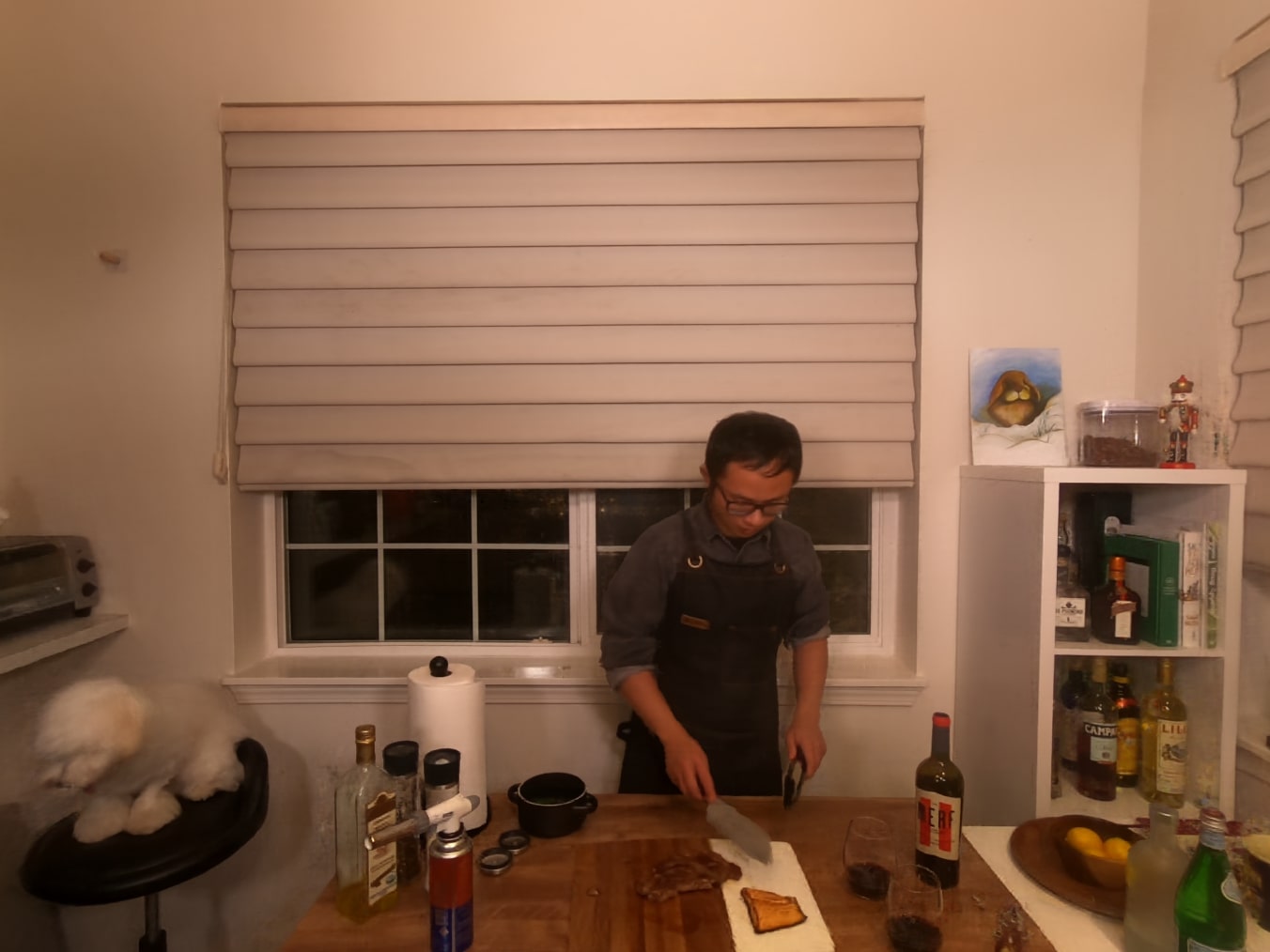}};
        \begin{scope}[x={($1*(image.south east)$)},y={($1*(image.north west)$)}]
          \draw[red,thick] (0.4,0.26) rectangle (0.72,0.58);
        \end{scope}
      \end{tikzpicture}%
    }\hspace{-1mm}\hfill%
    \mpage{0.49}{%
      \adjincludegraphics[width=\linewidth,trim={{0.4\width} {0.26\height} {0.28\width} {0.42\height}},clip]{figures/fig_dynamic/img/n3dv/0010_ours.jpg}\\[0.2mm]%
    }
  }\hfill%
  \mpage{0.32}{%
    \mpage{0.49}{%
      \begin{tikzpicture}
        \node[anchor=south west,inner sep=0] (image) at (0,0) {\includegraphics[width=\linewidth]{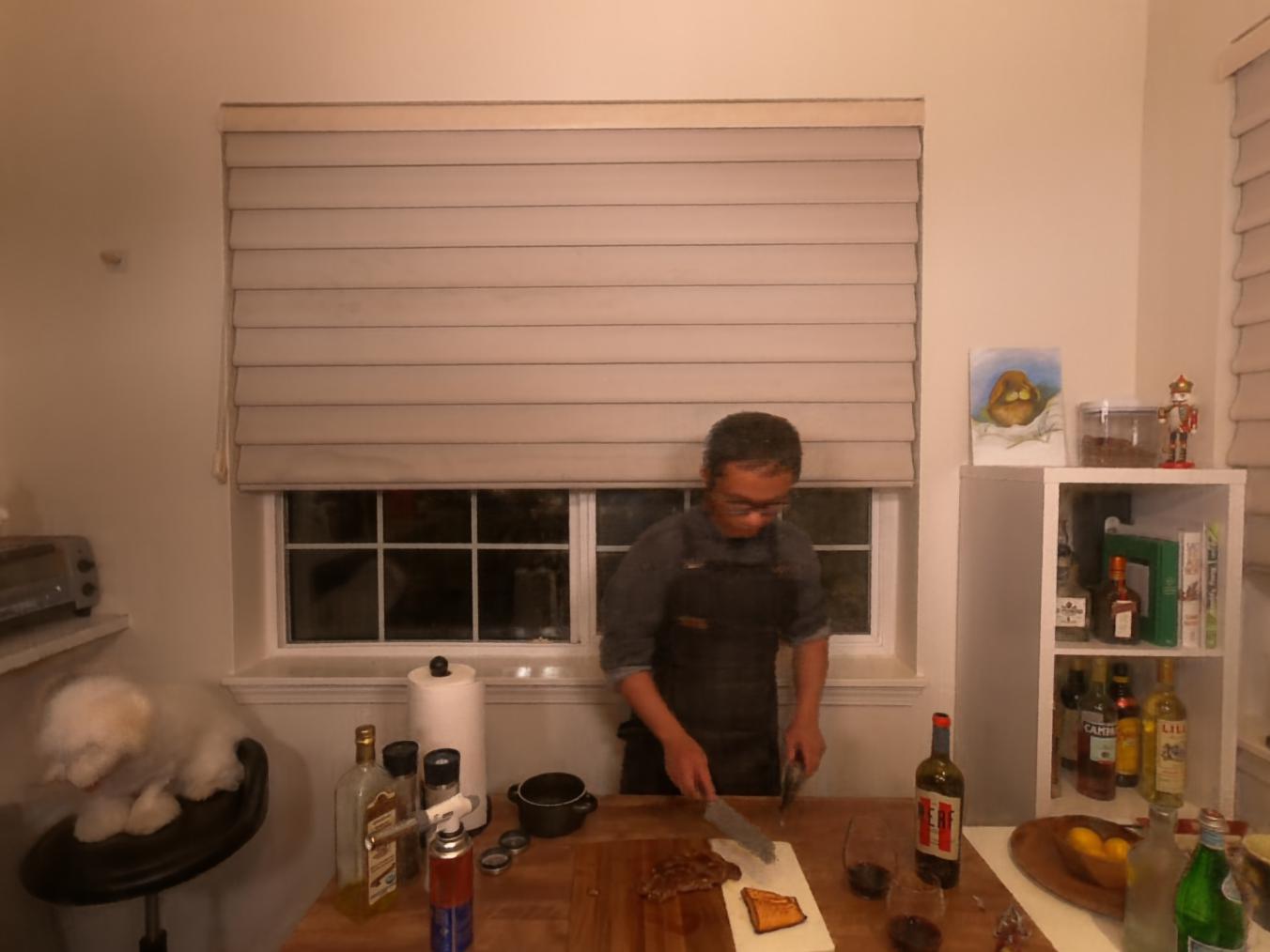}};
        \begin{scope}[x={($1*(image.south east)$)},y={($1*(image.north west)$)}]
          \draw[red,thick] (0.4,0.26) rectangle (0.72,0.58);
        \end{scope}
      \end{tikzpicture}%
    }\hspace{-1mm}\hfill%
    \mpage{0.49}{%
      \adjincludegraphics[width=\linewidth,trim={{0.4\width} {0.26\height} {0.28\width} {0.42\height}},clip]{figures/fig_dynamic/img/n3dv/0010_nerfplayer.jpg}\\[0.2mm]%
    }
  }\\[0.1mm]%
  \mpage{0.32}{%
    Ground truth (Neural 3D Video~\cite{LiSZGLKSLGL2022})%
  }\hfill%
  \mpage{0.32}{%
    Ours%
  }\hfill%
  \mpage{0.32}{%
    NeRFPlayer~\cite{SongCLCCYXG2023}
  }\\[1mm]%
  \mpage{0.32}{%
    \mpage{0.49}{%
      \begin{tikzpicture}
        \node[anchor=south west,inner sep=0] (image) at (0,0) {\includegraphics[width=\linewidth]{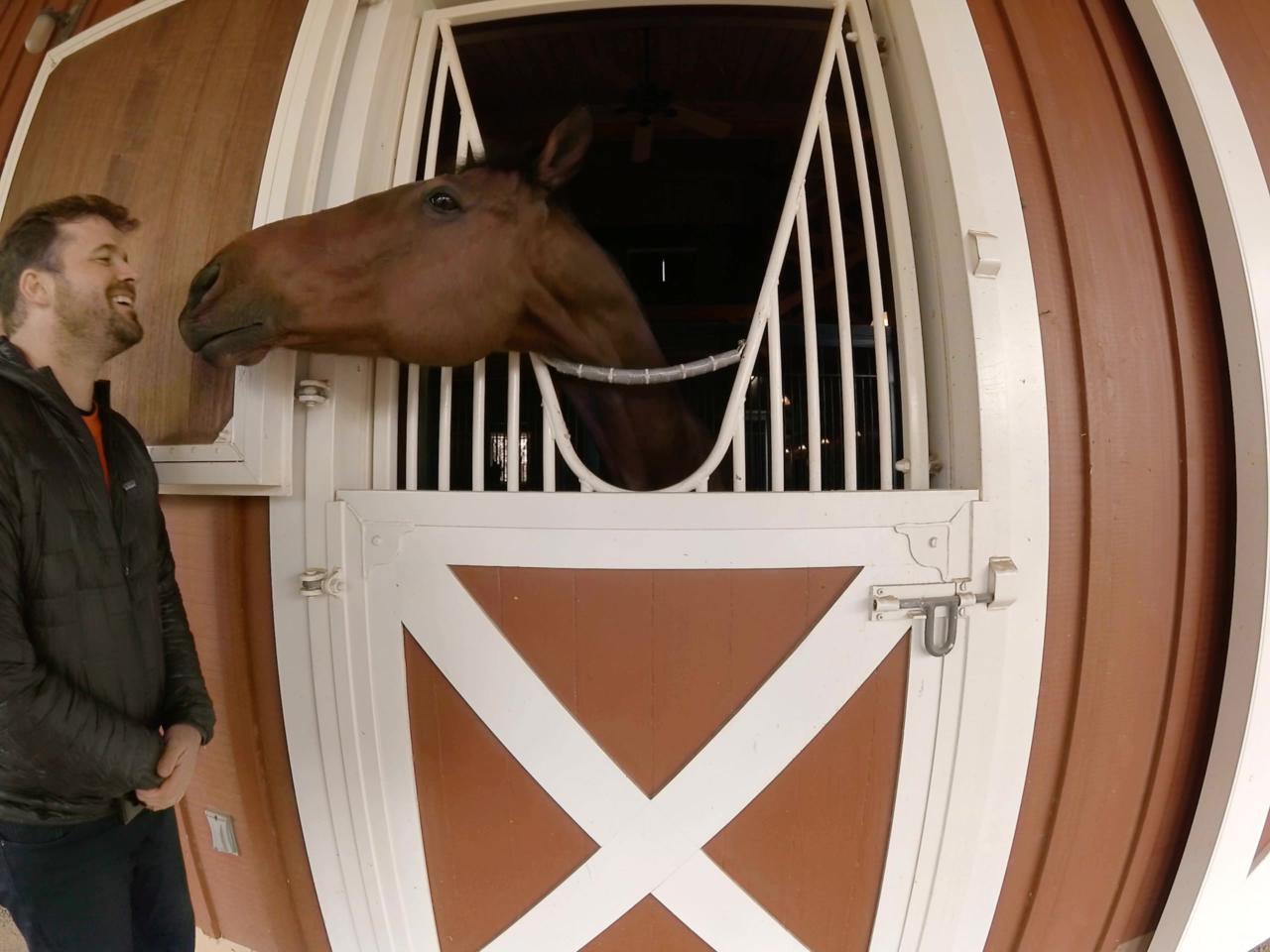}};
        \begin{scope}[x={($1*(image.south east)$)},y={($1*(image.north west)$)}]
          \draw[red,thick] (0,0.47) rectangle (0.5,0.97);
        \end{scope}
      \end{tikzpicture}%
    }\hspace{-1mm}\hfill%
    \mpage{0.49}{%
      \adjincludegraphics[width=\linewidth,trim={{0\width} {0.47\height} {0.5\width} {0.03\height}},clip]{figures/fig_dynamic/img/immersive/gt.jpg}\\[0.2mm]%
    }
  }\hfill%
  \mpage{0.32}{%
    \mpage{0.49}{%
      \begin{tikzpicture}
        \node[anchor=south west,inner sep=0] (image) at (0,0) {\includegraphics[width=\linewidth]{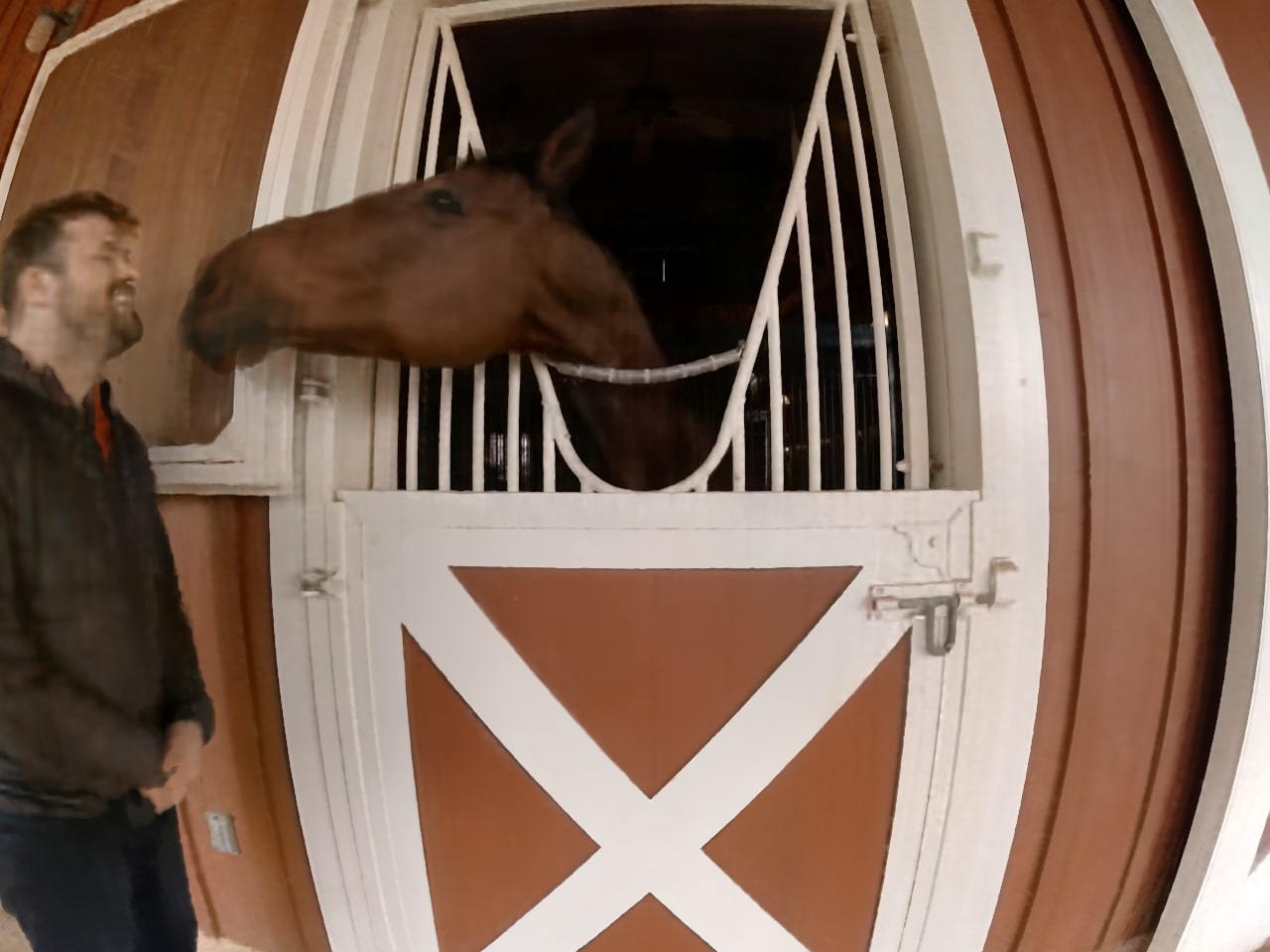}};
        \begin{scope}[x={($1*(image.south east)$)},y={($1*(image.north west)$)}]
          \draw[red,thick] (0,0.47) rectangle (0.5,0.97);
        \end{scope}
      \end{tikzpicture}%
    }\hspace{-1mm}\hfill%
    \mpage{0.49}{%
      \adjincludegraphics[width=\linewidth,trim={{0\width} {0.47\height} {0.5\width} {0.03\height}},clip]{figures/fig_dynamic/img/immersive/ours.jpg}\\[0.2mm]%
    }
  }\hfill%
  \mpage{0.32}{%
    \mpage{0.49}{%
      \begin{tikzpicture}
        \node[anchor=south west,inner sep=0] (image) at (0,0) {\includegraphics[width=\linewidth]{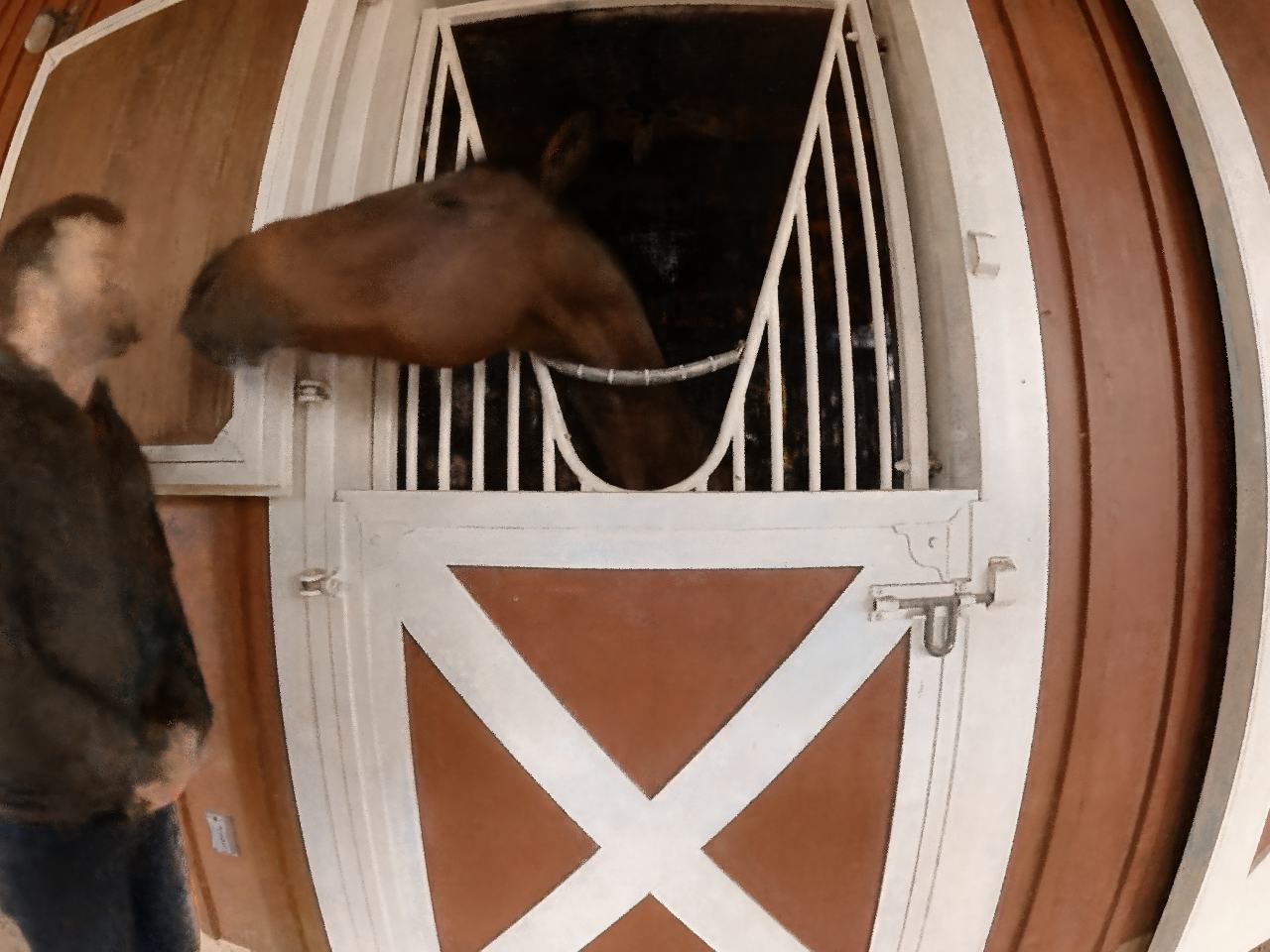}};
        \begin{scope}[x={($1*(image.south east)$)},y={($1*(image.north west)$)}]
          \draw[red,thick] (0,0.47) rectangle (0.5,0.97);
        \end{scope}
      \end{tikzpicture}%
    }\hspace{-1mm}\hfill%
    \mpage{0.49}{%
      \adjincludegraphics[width=\linewidth,trim={{0\width} {0.47\height} {0.5\width} {0.03\height}},clip]{figures/fig_dynamic/img/immersive/nerfplayer.jpg}\\[0.2mm]%
    }
  }\\[0.1mm]%
  \mpage{0.32}{%
    Ground truth (Google Immersive LF Video~\cite{BroxtFOEHDDBWD2020})%
  }\hfill%
  \mpage{0.32}{%
    Ours%
  }\hfill%
  \mpage{0.32}{%
    NeRFPlayer~\cite{SongCLCCYXG2023}
  }\\[-2mm]%
\caption{%
\textbf{Qualitative comparisons of dynamic reconstruction.} 
We show visual comparisons of our method on three datasets against two baselines on heldout views. 
We pick non-keyframe time-steps for evaluation, except for the Google Immersive light field video (last row), for which we pick the matching image to the NeRFPlayer \cite{SongCLCCYXG2023} result.
See our project webpage for more results and comparisons.
}%
\label{fig:qual}
\vspace{-5mm}
\end{figure*}

\paragraph{DoNeRF Dataset}
The DoNeRF dataset \cite{NeffSPKCKS2021} contains six synthetic sequences with images of 800$\times$800 pixel resolution.
Here, we validate the efficacy of our sample prediction network approach by comparing it to existing methods for static view synthesis, including NeRF, InstantNGP, and three sampling-network--based approaches \cite{NeffSPKCKS2021, KurzNLZS2022, PialaC2021}.

As demonstrated in \cref{tab:quant_static_donerf}, our approach outperforms all baselines in terms of quality and improves the performance of other sampling network schemes by a large margin.
Additionally, our model is implemented in vanilla PyTorch and renders 800$\times$800 pixel images at 6.5 FPS on a single RTX 3090 GPU (or 29 FPS with our \textit{Tiny} model).

We also compare our sampling network-based approach to the single-sample R2L light field representation \cite{WangRHOCFT2022} on the downsampled 400$\times$400 resolution DoNeRF dataset (with their provided metrics). 
We outperform their approach quantitatively without using pretrained teacher networks.
Further, inference with our six-layer, 256-hidden-unit network, and TensoRF volume backbone is faster than R2L's deep 88-layer, 256-hidden-unit MLP.

\paragraph{LLFF Dataset}
See supplementary material for additional quantitative comparisons on the LLFF dataset \cite{MildeSTBRN2020}, showing our network achieving high quality on real-world scenes.

\subsection{Comparisons on Dynamic Scenes}

\paragraph{Technicolor Dataset}
The Technicolor light field dataset \cite{sabater2017dataset} contains videos of varied indoor environments captured by a time-synchronized 4$\times$4 camera rig.
Each image in each video stream is 2048$\times$1088 pixels, and we hold out the view in the second row and second column for evaluation.
We compare HyperReel to Neural 3D Video \cite{LiSZGLKSLGL2022} at full image resolution on five sequences (\textit{Birthday, Fabien, Painter, Theater, Trains}) from this dataset, each 50 frames long.
We train Neural 3D Video on each sequence for approximately one week on a machine with 8 NVIDIA V100 GPUs.

We show in \cref{tab:quant_dynamic} that the quality of HyperReel exceeds that of Neural 3D Video \cite{LiSZGLKSLGL2022} while also training in just 1.5 GPU hours per sequence (rather than 1000+ GPU hours for Neural 3D), and rendering far more quickly.

\paragraph{Neural 3D Video Dataset}
The Neural 3D Video dataset \cite{LiSZGLKSLGL2022} contains six indoor multi-view video sequences captured by 20 cameras at 2704$\times$2028 pixel resolution.
We downsample all sequences by a factor of 2 for training and evaluation and hold out the central view for evaluation.
Metrics are averaged over all scenes.
Additionally, due to the challenging nature of this dataset (time synchronization errors, inconsistent white balance, imperfect poses), we output 64 $z$-planes per ray with our sample network rather than 32.

We show in \cref{tab:quant_dynamic} that we quantitatively outperform NeRFPlayer \cite{SongCLCCYXG2023} while rendering approximately 40 times faster.
While StreamRF~\cite{LiSWST2022} makes use of a custom CUDA implementation that renders faster than our model, our approach consumes less memory on average per frame than both StreamRF and NeRFPlayer.

\paragraph{Google Immersive Dataset}
The Google Immersive dataset \cite{BroxtFOEHDDBWD2020} contains light field videos of various indoor and outdoor environments captured by a time-synchronized 46-fisheye camera rig.
Here, we compare our approach to NeRFPlayer and select the same seven scenes as NeRFPlayer for evaluation on this dataset (\textit{Welder, Flames, Truck, Exhibit, Face Paint 1, Face Paint 2, Cave}), holding out the central view for validation.
Our results in \cref{tab:quant_dynamic} outperform NeRFPlayer's by a 3\,dB margin and renders more quickly.%

\paragraph{DeepView Dataset}
As Google's Immersive Light Field Video \cite{BroxtFOEHDDBWD2020} does not provide quantitative benchmarks for the performance of their approach in terms of image quality, we provide an additional comparison of our approach to DeepView \cite{FlynnBDDFOST2019} in the supplementary material.

\subsection{Ablation Studies}

\paragraph{Number of Keyframes.}
In \cref{tab:quant_ablation}, we ablate our method on the Technicolor light field dataset with different numbers of keyframes.
Increasing the number of keyframes allows our model to capture more complex motions, but also distributes the volume's capacity over a larger number of time steps.
Our choice of one keyframe for every four frames strikes a good balance between temporal resolution and spatial rank, and achieves the best overall performance (\cref{tab:quant_ablation}).

\begin{figure}[t]
\footnotesize
\centering%
  \mpage{0.19}{%
    \adjincludegraphics[height=\linewidth,trim={{0.22\width} {0.36\height} {0.6\width} {0.32\height}},clip,angle=90]{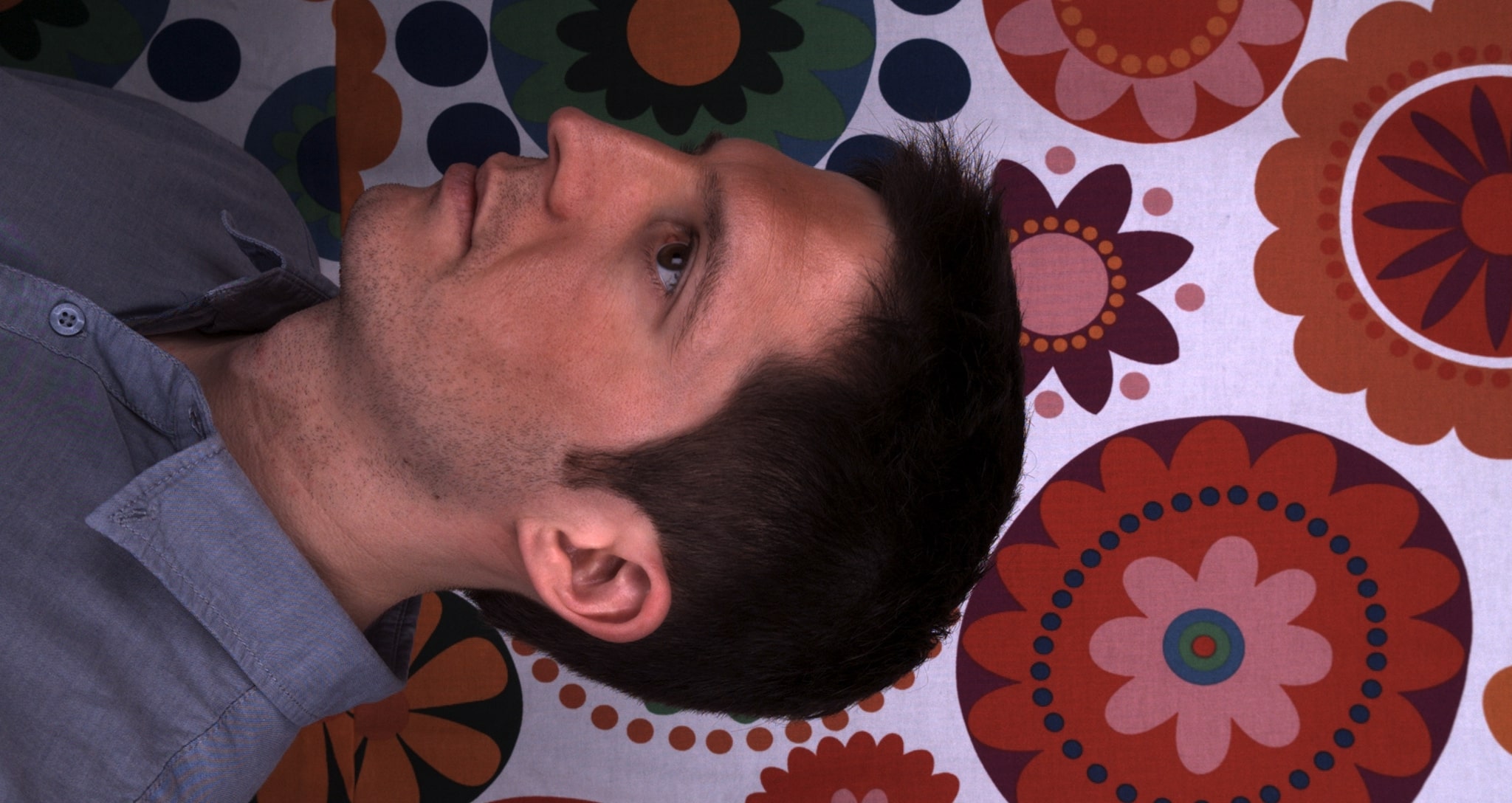}\\[0.2mm]%
  }\hspace{-1mm}\hfill%
  \mpage{0.19}{%
    \adjincludegraphics[height=\linewidth,trim={{0.22\width} {0.36\height} {0.6\width} {0.32\height}},clip,angle=90]{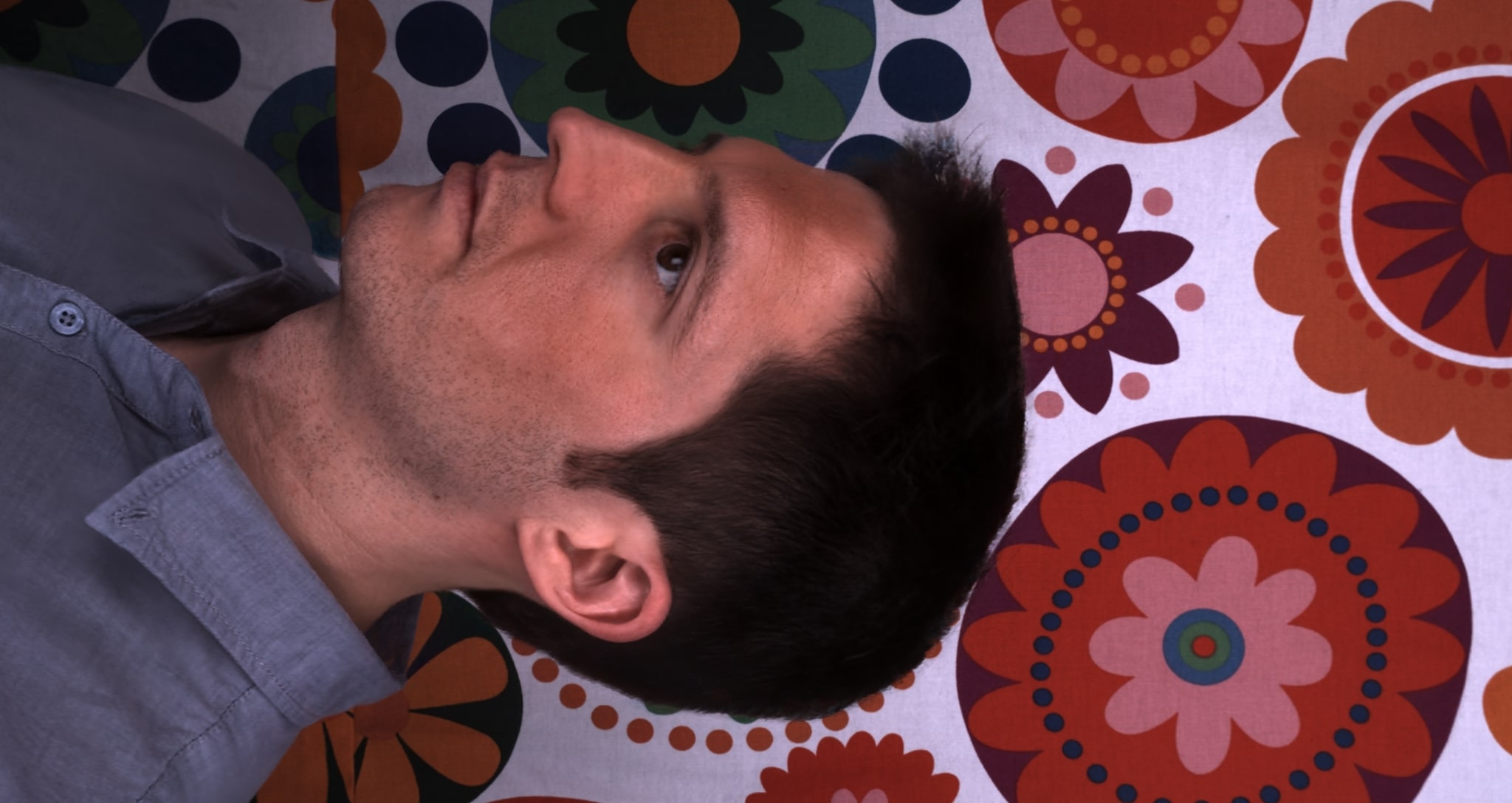}\\[0.2mm]%
  }\hspace{-1mm}\hfill%
  \mpage{0.19}{%
    \adjincludegraphics[height=\linewidth,trim={{0.22\width} {0.36\height} {0.6\width} {0.32\height}},clip,angle=90]{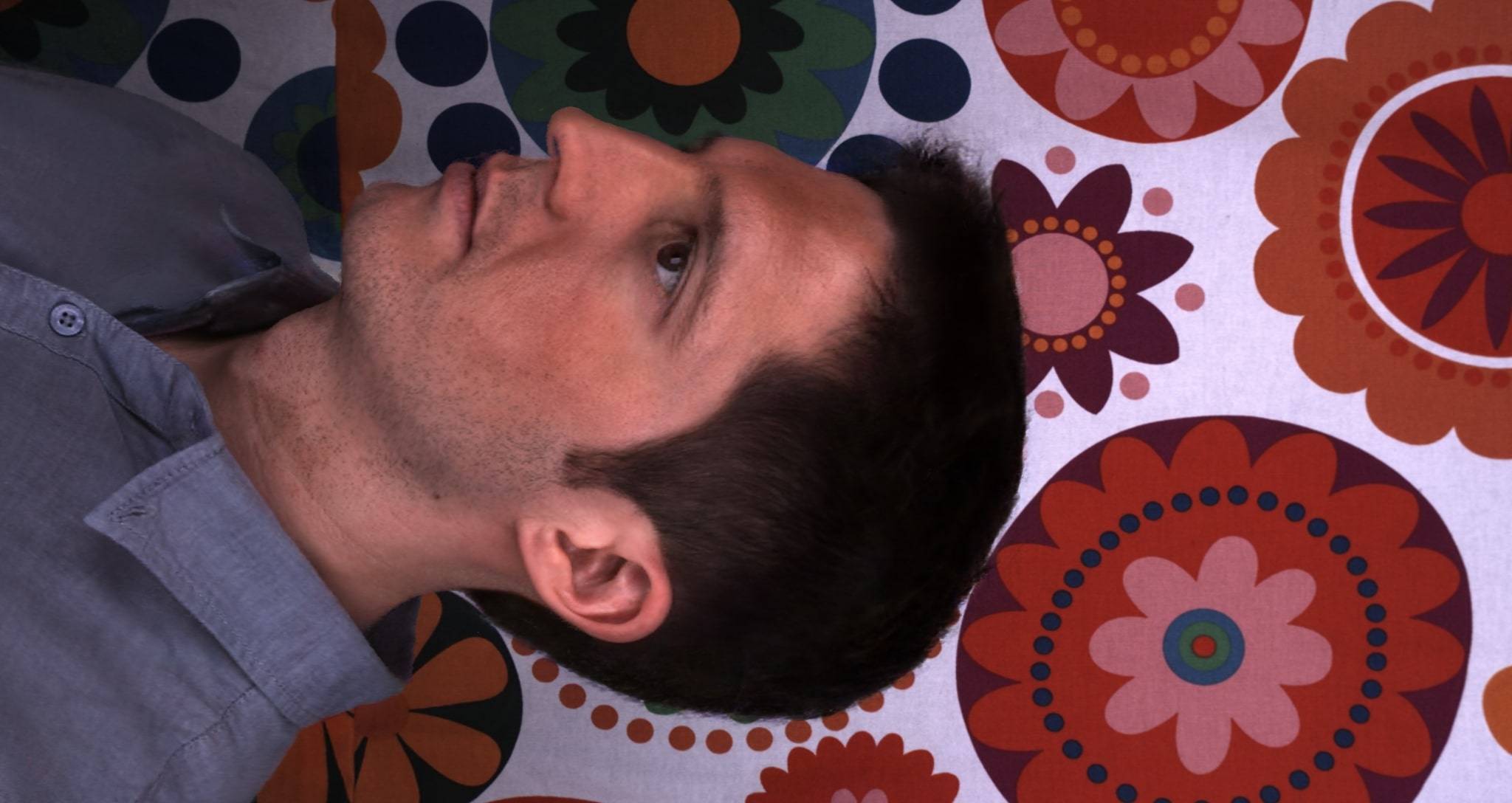}\\[0.2mm]%
  }\hspace{-1mm}\hfill%
  \mpage{0.19}{%
    \adjincludegraphics[height=\linewidth,trim={{0.22\width} {0.36\height} {0.6\width} {0.32\height}},clip,angle=90]{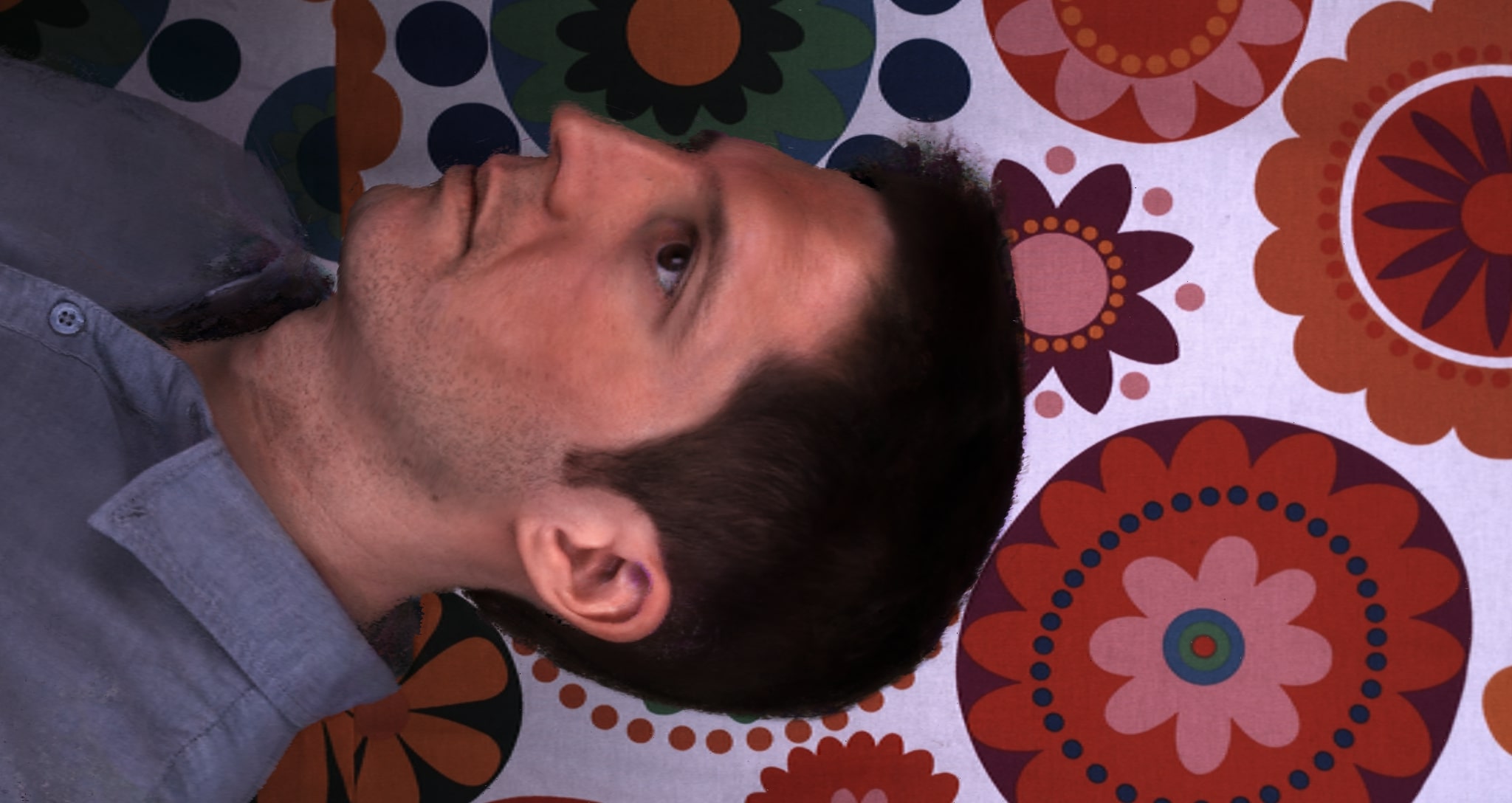}\\[0.2mm]%
  }\hspace{-1mm}\hfill%
  \mpage{0.19}{%
    \adjincludegraphics[height=\linewidth,trim={{0.22\width} {0.36\height} {0.6\width} {0.32\height}},clip,angle=90]{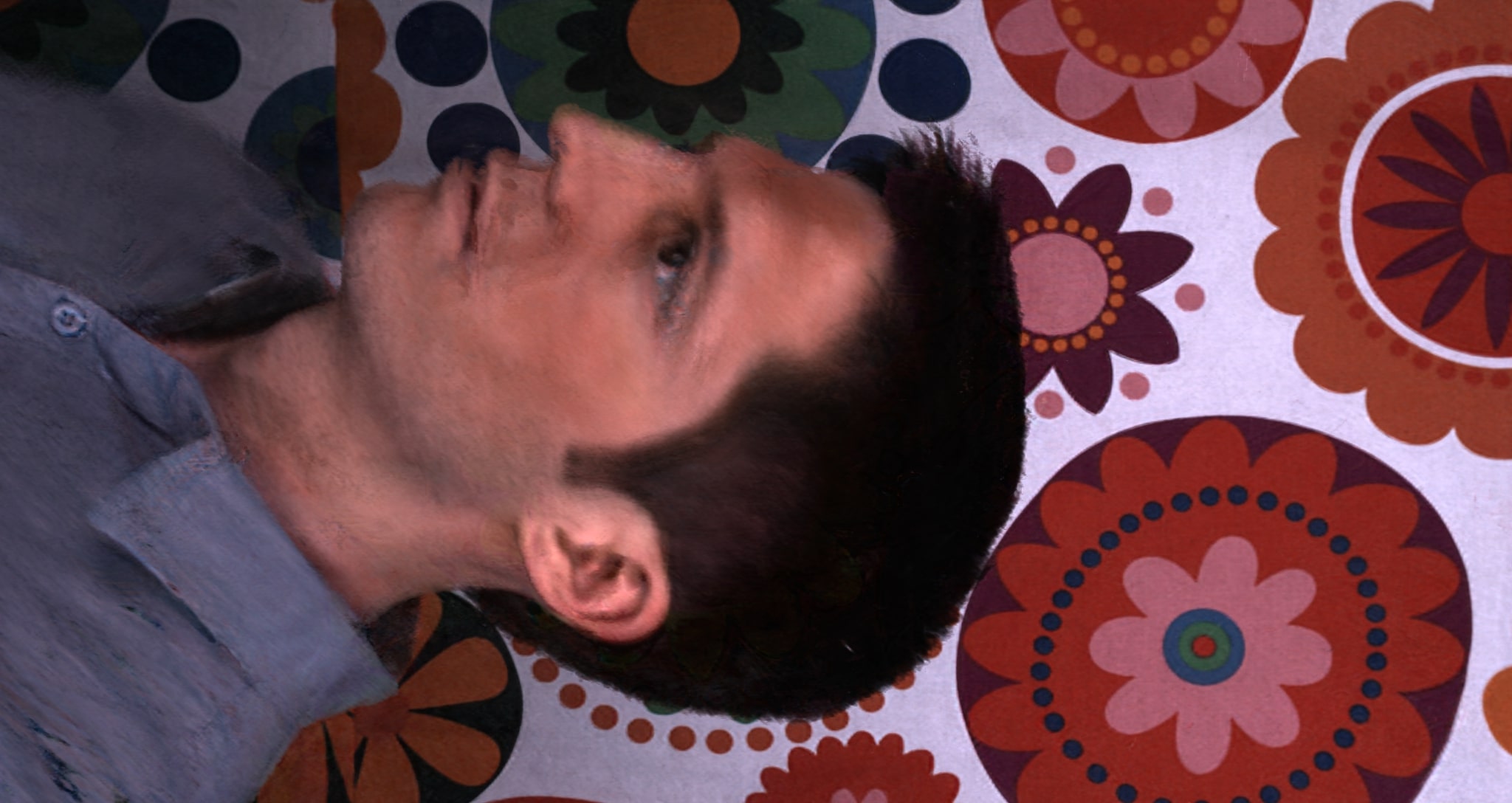}\\[0.2mm]%
  }\\[1mm]%
  \mpage{0.19}{%
    \adjincludegraphics[width=\linewidth,trim={{0.43\width} {0.27\height} {0.44\width} {0.6\height}},clip]{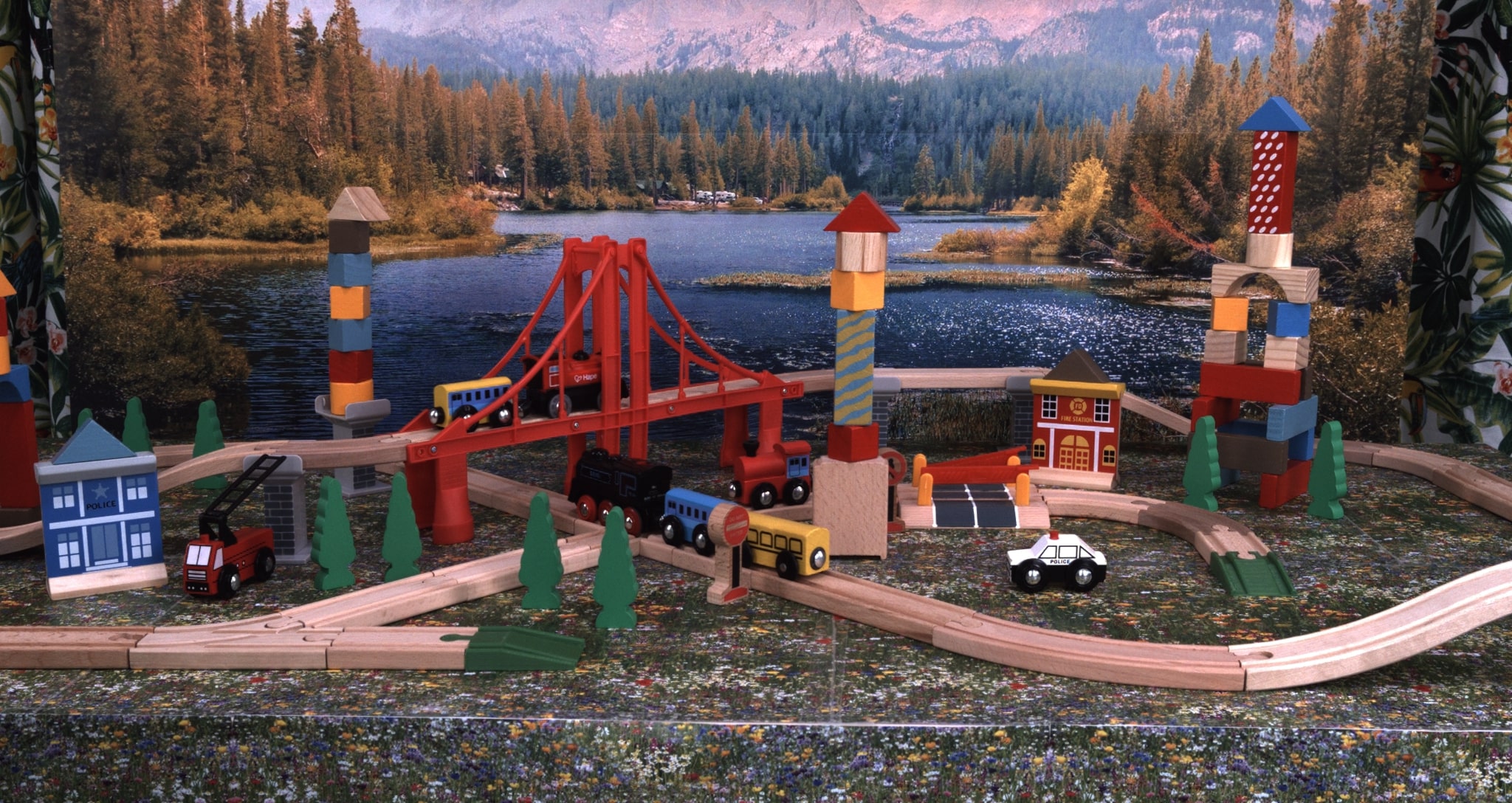}\\[0.2mm]%
  }\hspace{-1mm}\hfill%
  \mpage{0.19}{%
    \adjincludegraphics[width=\linewidth,trim={{0.43\width} {0.27\height} {0.44\width} {0.6\height}},clip]{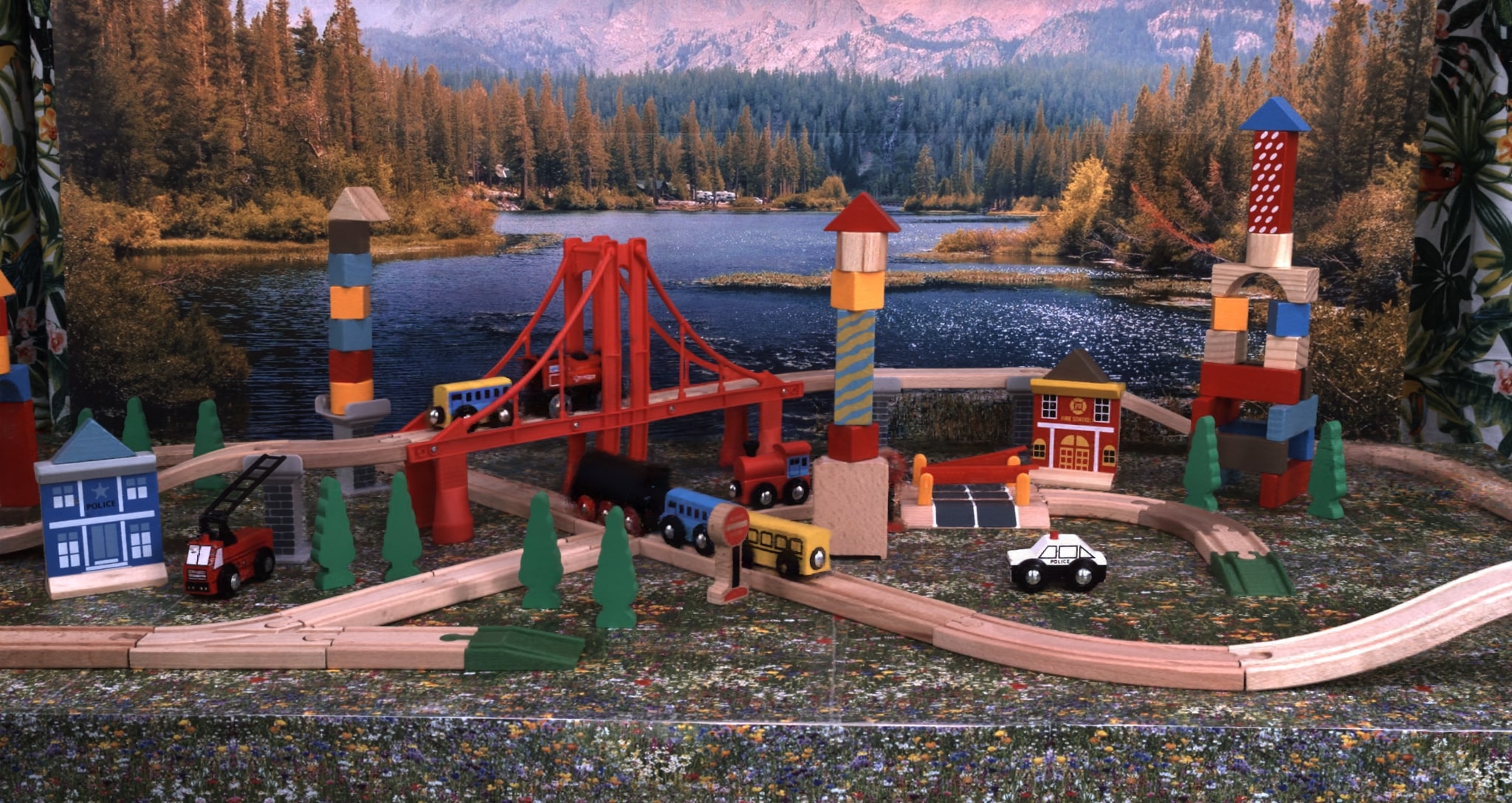}\\[0.2mm]%
  }\hspace{-1mm}\hfill%
  \mpage{0.19}{%
    \adjincludegraphics[width=\linewidth,trim={{0.43\width} {0.27\height} {0.44\width} {0.6\height}},clip]{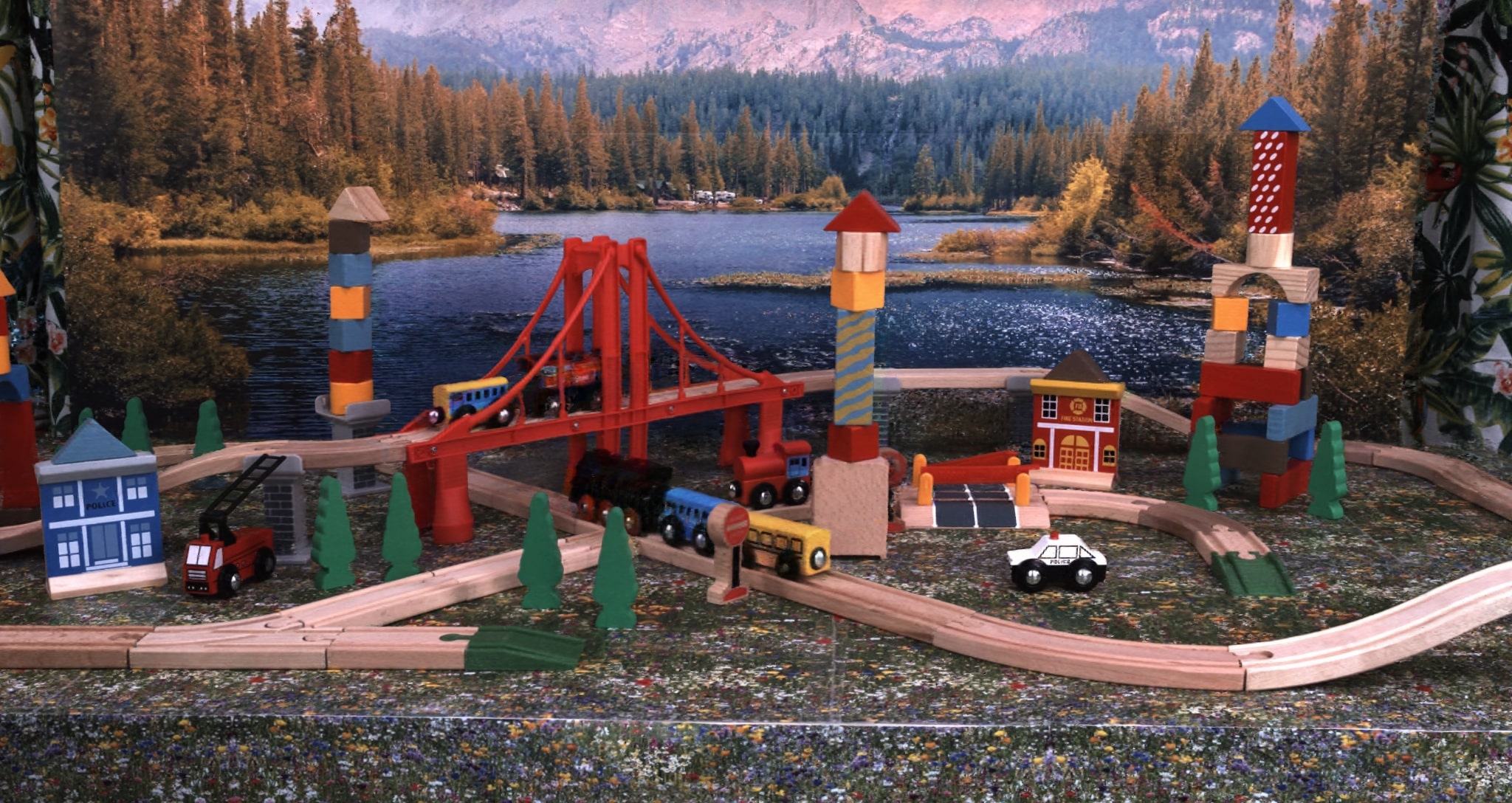}\\[0.2mm]%
  }\hspace{-1mm}\hfill%
  \mpage{0.19}{%
    \adjincludegraphics[width=\linewidth,trim={{0.43\width} {0.27\height} {0.44\width} {0.6\height}},clip]{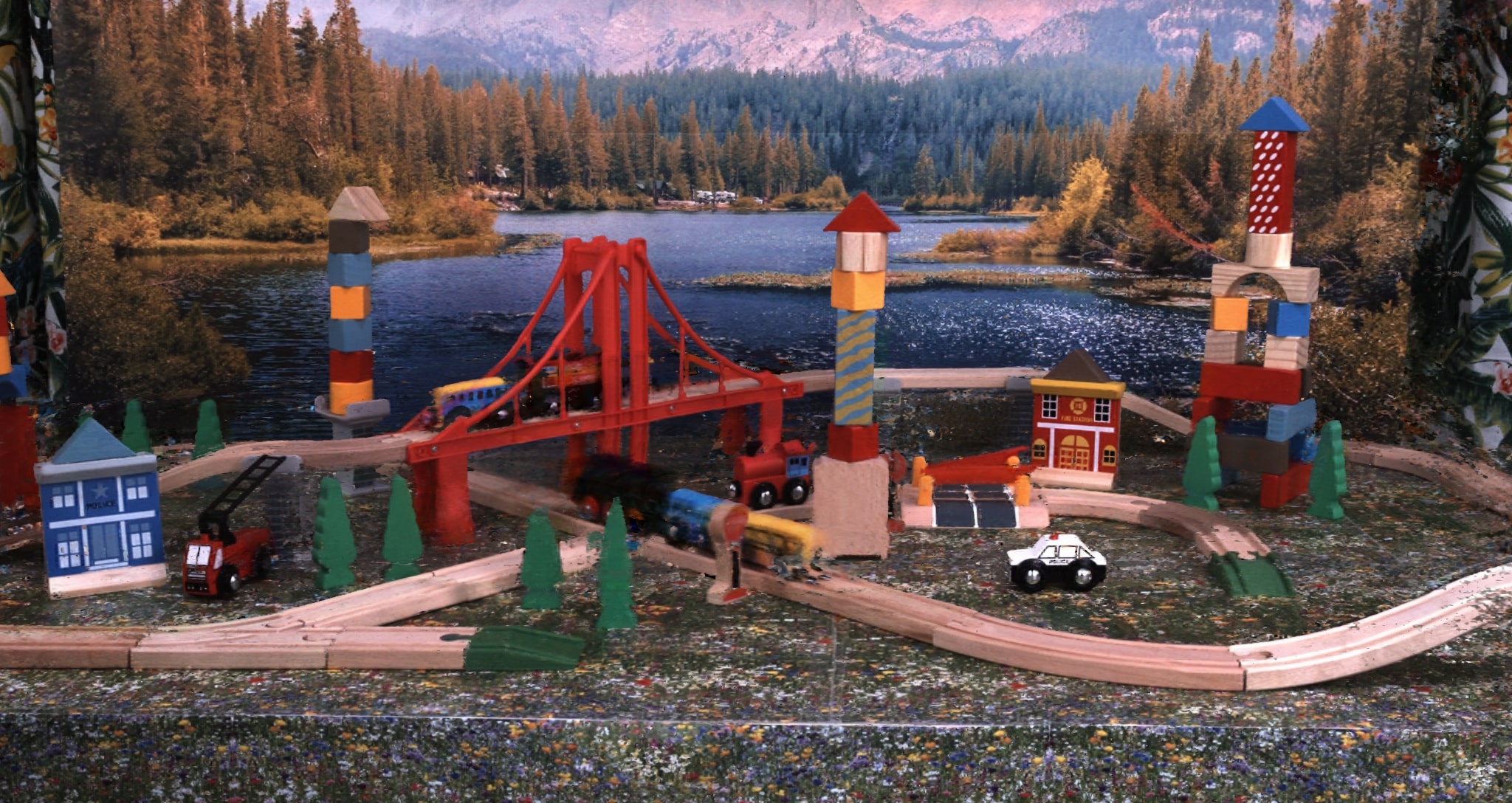}\\[0.2mm]%
  }\hspace{-1mm}\hfill%
  \mpage{0.19}{%
    \adjincludegraphics[width=\linewidth,trim={{0.43\width} {0.27\height} {0.44\width} {0.6\height}},clip]{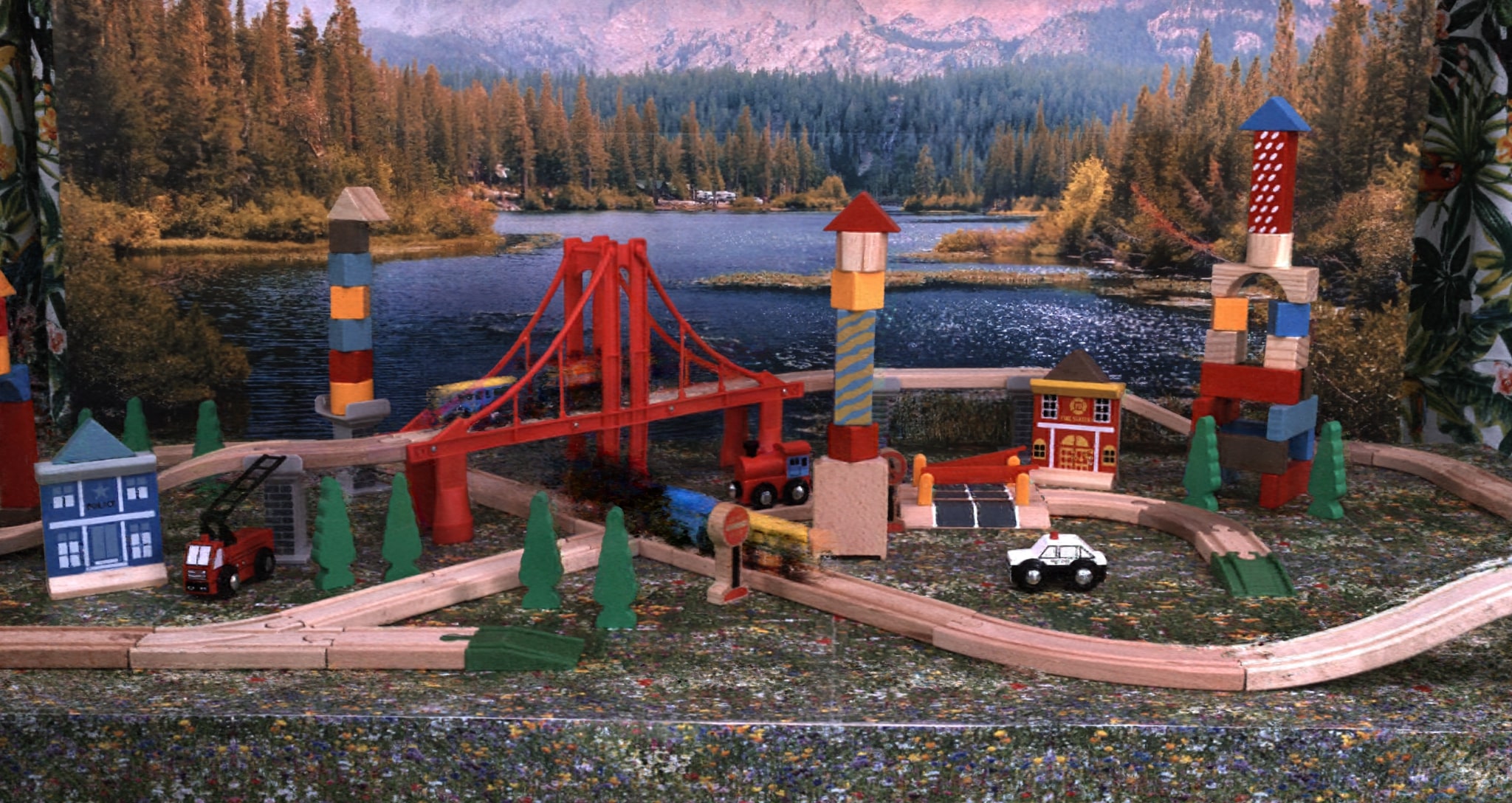}\\[0.2mm]%
  }\\[1mm]%
  \mpage{0.19}{%
    GT%
  }\hspace{-1mm}\hfill%
  \mpage{0.19}{%
    Full model%
  }\hspace{-1mm}\hfill%
  \mpage{0.19}{%
    Small%
  }\hspace{-1mm}\hfill%
  \mpage{0.19}{%
    Tiny%
  }\hspace{-1mm}\hfill%
  \mpage{0.19}{%
    No sampling%
  }\\[2mm]%
\vspace{-4mm}
\caption{\textbf{Ablations on our sampling network.} We show close-up results for various sampling networks architectures on two of the Technicolor sequences also shown in \cref{fig:qual}.}
\label{fig:abl_samples}
\vspace{-2mm}
\end{figure}

\paragraph{Network Size and Number of Primitives.}
We also show the performance of our method with different network designs in \cref{tab:quant_ablation},
including the performance for a \textit{Tiny} model (4-layers, 128-hidden-unit MLP with 8 predicted primitives), and \textit{Small} model (4-layers, 256-hidden-unit MLP with 16 predicted primitives).
Our \textit{Tiny} model runs at 18\,FPS, and our \textit{Small} model runs at 9\,FPS at megapixel resolution, again without any custom CUDA code.
Our \textit{Tiny} model performs reasonably well but achieves worse quality than Neural 3D Video on the Technicolor dataset. 
In contrast, our \textit{Small} model achieves comparable overall performance to Neural3D---showing that we can still achieve good quality renderings at even higher frame rates.
We show accompanying qualitative results for these models in \cref{fig:abl_samples}.

\paragraph{With and Without Sample Prediction Network.}
We show results on the Technicolor dataset \textit{without} our sample prediction network, using every frame as a keyframe, and with 4$\times$ the number of samples (128 vs. 32).
Our full method outperforms this approach by a sizeable margin. 

\paragraph{With and Without Point Offset.}
In \cref{tab:quant_point}, we show results on two static scenes \textit{with} and \textit{without} point offsets (\cref{eqn:point-offsets}): one diffuse and one highly refractive scene.
Point offsets improve quality in both cases, suggesting that they may help with better model capacity allocation in addition to view-dependence---similar to ``canonical frame'' deformations used in Nerfies \cite{ParkSHBBGMS2021} and Neural Volumes \cite{LombaSSSLS2019}.

\section{Conclusion}
\label{sec:conclusion}

\begin{figure}[t]
\footnotesize
\centering%
  \mpage{0.42}{%
    \begin{tikzpicture}
      \node[anchor=south west,inner sep=0] (image) at (0,0) {\adjincludegraphics[height=16mm]{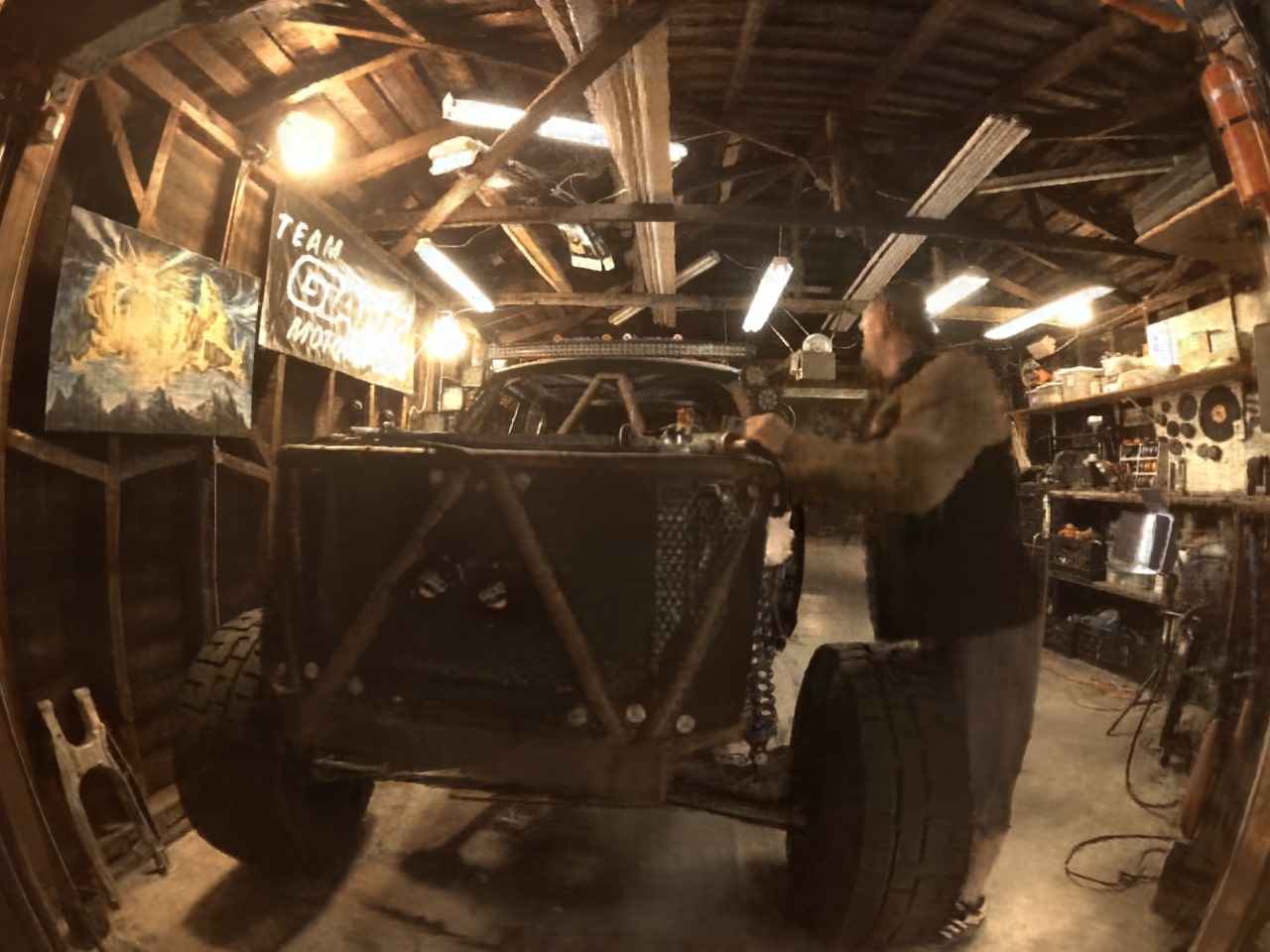}};
      \begin{scope}[x={($1*(image.south east)$)},y={($1*(image.north west)$)}]
        \draw[red,thick] (0.64,0.26) rectangle (0.77,0.75);
      \end{scope}
    \end{tikzpicture}\hfill%
    \adjincludegraphics[height=16mm,trim={{0.64\width} {0.26\height} {0.23\width} {0.25\height}},clip]{figures/fig_failure/img/truck_0002_ours.jpg}\hfill%
    \adjincludegraphics[height=16mm,trim={{0.64\width} {0.26\height} {0.23\width} {0.25\height}},clip]{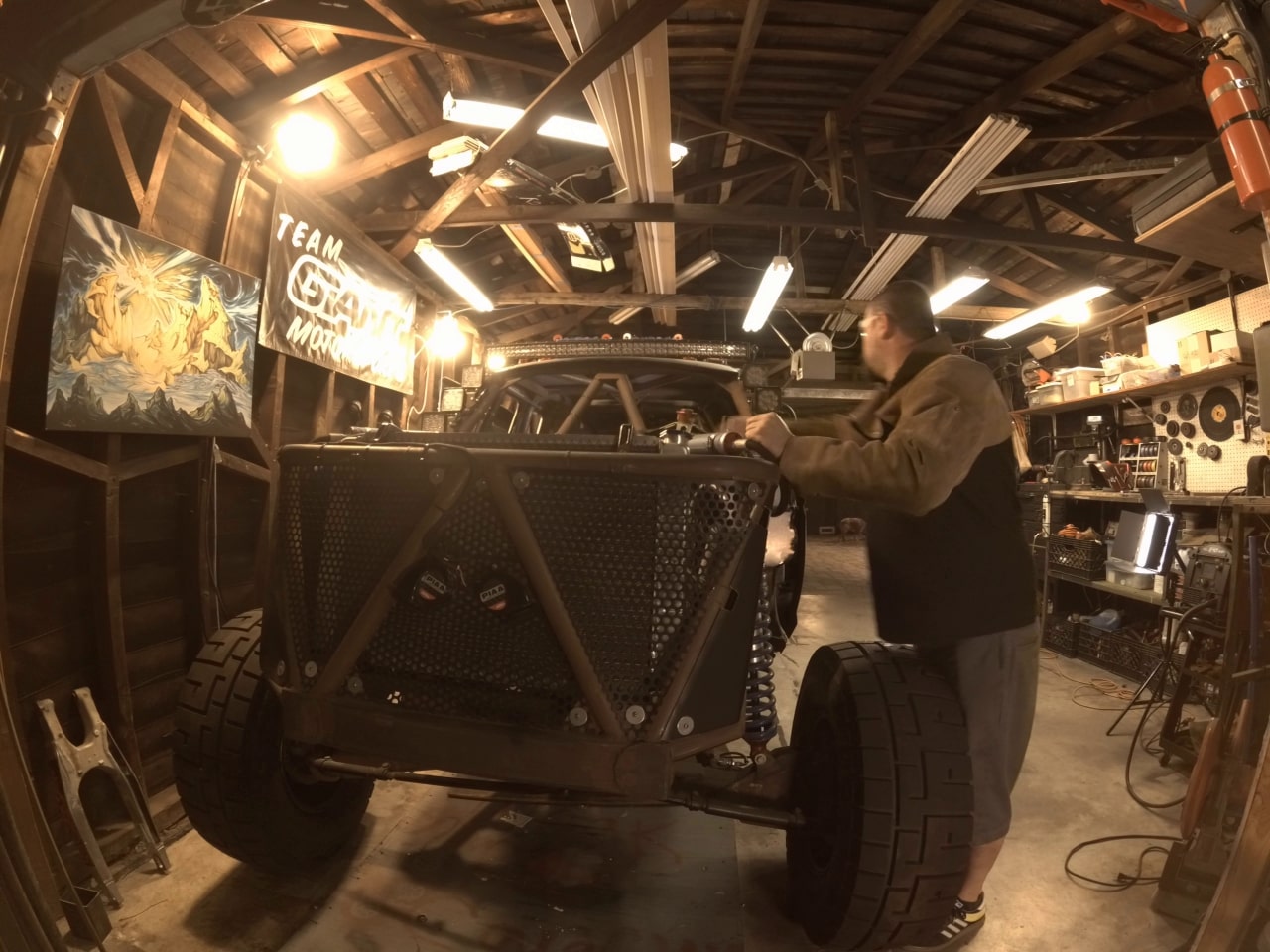}\hfill\\[0.2mm]%
  }\hspace{-1mm}\hfill%
  \mpage{0.56}{%
    \hfill%
    \begin{tikzpicture}
      \node[anchor=south west,inner sep=0] (image) at (0,0) {\adjincludegraphics[height=16mm]{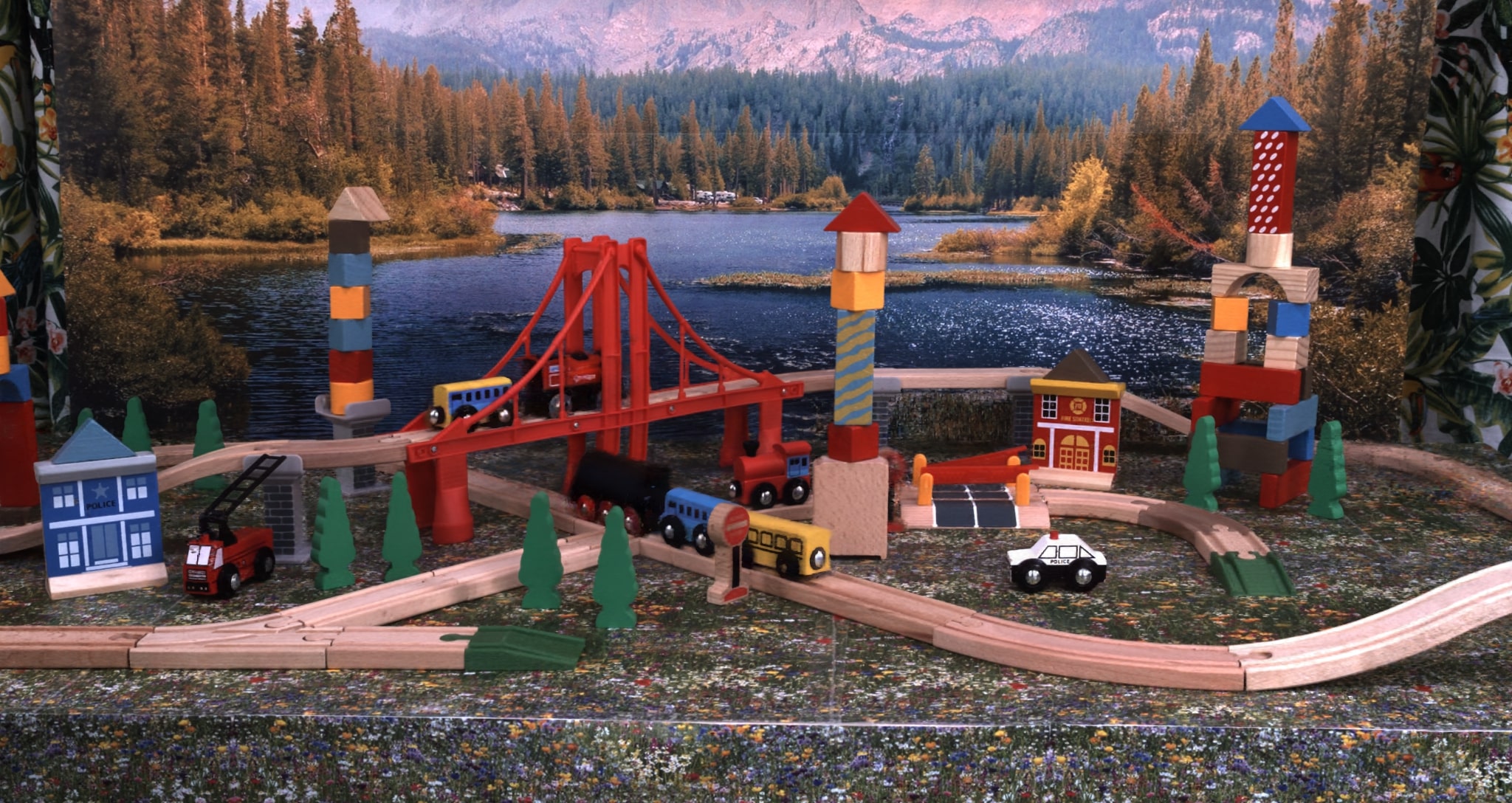}};
      \begin{scope}[x={($1*(image.south east)$)},y={($1*(image.north west)$)}]
        \draw[red,thick] (0.89,0.11) rectangle (1,0.61);
      \end{scope}
    \end{tikzpicture}\hfill%
    \adjincludegraphics[height=16mm,trim={{0.89\width} {0.11\height} 0 {0.39\height}},clip]{figures/fig_failure/img/train_0002_ours.jpg}\hfill%
    \adjincludegraphics[height=16mm,trim={{0.89\width} {0.11\height} 0 {0.39\height}},clip]{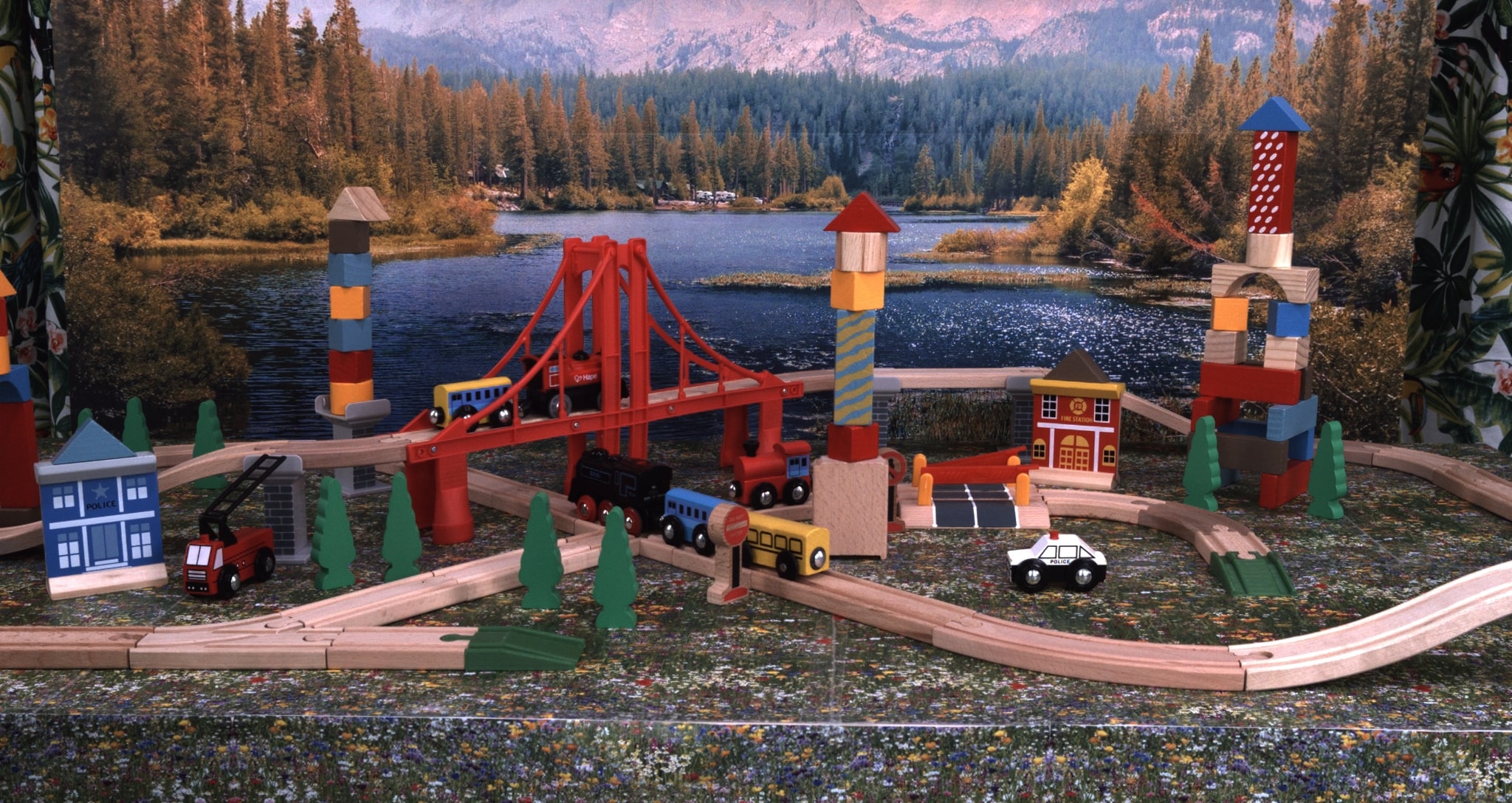}\\[0.2mm]%
  }\\[0.5mm]%
  \mpage{0.34}{%
    $\underbracket[1pt][1.5mm]{\hspace{0.95\linewidth}}_{\substack{\vspace{-3.0mm}\\\colorbox{white}{{Our result}}}}$%
  }\hspace{-2mm}\hfill%
  \mpage{0.08}{%
    GT%
  }\hspace{-1mm}\hfill%
  \mpage{0.48}{%
    $\underbracket[1pt][1.5mm]{\hspace{0.95\linewidth}}_{\substack{\vspace{-3.0mm}\\\colorbox{white}{{Our result}}}}$%
  }\hspace{-2mm}\hfill%
  \mpage{0.08}{%
    GT%
  }\\[-2mm]%
\caption{\textbf{Limitations.} Our approach can sometimes produce blurry reconstructions due to the training ray subsampling scheme (\cref{sec:optimization}) (\emph{left}) or noisy reconstructions in sparsely observed regions due to an under-constrained sampling network (\emph{right}).}
\label{fig:failures}
\vspace{-5mm}
\end{figure}

HyperReel is a novel representation for 6-DoF video, which combines a ray-conditioned sampling network with a keyframe-based dynamic volume representation.
It achieves a balance between high rendering quality, speed, and memory efficiency that sets it apart from existing 6-DoF video representations.
We qualitatively and quantitatively compare our approach to prior and contemporary 6-DoF video representations, showing that HyperReel outperforms each of these works along multiple axes.

\paragraph{Limitations and Future Work}
Our sample network is only supervised by a rendering loss on the training images, and predicts ray-dependent sample points that need not be consistent between different views.
This can lead to a reduction in quality for views outside of the convex hull of the training cameras or for scene content that is only observed in a small number of views---manifesting in some cases as temporal jittering, view-dependent object motion, or noisy reconstructions (see \cref{fig:failures}).
Exploring regularization methods that enable reasonable geometry predictions even for extrapolated views is an important future direction.

Although our keyframe-based representation is more memory efficient than most existing 3D video formats, it cannot be \textit{streamed} like NeRFPlayer \cite{SongCLCCYXG2023} or StreamRF \cite{LiSWST2022}.
However, our sample network approach is in principle compatible with any streaming-based dynamic volume.

Currently, our approach falls short of the rendering speed required for settings like VR (ideally 72\,FPS, in stereo).
As our method is implemented in vanilla PyTorch, we expect to gain significant speedups with more engineering effort.

\paragraph{Acknowledgments}

We thank Thomas Neff, Yu-Lun Liu, and Xiaoming Zhao for valuable feedback and discussions, Zhaoyang Lv for help
with comparisons
\cite{LiSZGLKSLGL2022}, and Liangchen Song for providing information about the Google Immersive Video dataset~\cite{BroxtFOEHDDBWD2020} used in NeRFPlayer~\cite{SongCLCCYXG2023}.
Matthew O'Toole acknowledges support from NSF IIS-2008464.

{\small
\bibliographystyle{ieeenat_fullname}
\bibliography{HyperReel-CR,main}
}

\appendix
\counterwithin{figure}{section}
\counterwithin{table}{section}

\section{Appendix Overview}
\noindent
Within the appendix, we provide:

\begin{enumerate}
    \item Additional details regarding training and evaluation for static and dynamic datasets in ~\cref{sec:evaluation_details};
    \item Additional details regarding sample network design, implementation, and training in~\cref{sec:network_details};
    \item Additional details regarding keyframe-based volume design in~\cref{sec:keyframe_details};
    \item Additional quantitative comparisons against static view synthesis approaches on the LLFF~\cite{MildeSTBRN2020} and DeepView~\cite{FlynnBDDFOST2019} datasets in~\cref{sec:additional_comparisons};
    \item Additional qualitative comparisons to Neural 3D Video Synthesis~\cite{LiSZGLKSLGL2022} on the Technicolor dataset~\cite{sabater2017dataset} in~\cref{sec:additional_n3d};
    \item Additional qualitative results for, \textbf{(a)} full 360 degree FoV captures and \textbf{(b)} highly refractive scenes in~\cref{sec:additional_results};
\end{enumerate}

\noindent Further, we provide a full per-scene breakdown of image metrics for the Technicolor dataset in \cref{tab:technicolor}, the Neural 3D Video dataset in \cref{tab:n3d}, and the Google Immersive Light Field Video dataset in \cref{tab:immersive}.

\section{Website Overview}
Finally, in addition to our appendix, our supplemental website \textbf{\href{https://hyperreel.github.io}{\texttt{https://hyperreel.github.io}}} contains:

\begin{enumerate}
    \item A link to our codebase;
    \item Videos of a demo running in real-time at high-resolution without any custom CUDA code;
    \item Dynamic dataset results from our method on each of Technicolor~(\cite{sabater2017dataset}), Neural 3D Video~(\cite{LiSZGLKSLGL2022}), and Google Immersive Video~(\cite{BroxtFOEHDDBWD2020});
    \item Qualitative results and comparisons on view-dependent static scenes from the Shiny Dataset~(\cite{wizadwongsa2021nex}) and the Stanford Light Field Dataset~(\cite{wilburn2005high});
    \item Qualitative comparison to~\cite{BroxtFOEHDDBWD2020}.
\end{enumerate}

\section{Additional Training \& Evaluation Details}
\label{sec:evaluation_details}

\begin{algorithm*}[t]
\SetAlgoNoLine
\KwIn{Number of videos $\{N\}$, Number of frames $\{M\}$}
\KwOut{Training Rays $\textit{rays}_{GT}$, Ground Truth Colors $C_{GT}$}
\;
\tcp{Initialize rays and colors}
$rays_{GT} =$ \textit{\{\}}\;
$C_{GT} =$ \textit{\{\}}\;
\tcp{Iterate over all $N$ videos}
\For{$n \in \{1, \cdots, \text{N}\}$}{
    \tcp{Iterate over all $M$ frames in video $n$}
    \For{$m \in \{1, \cdots, M\}$}{
            \tcp{Get frame $m$ from video $n$}
            $C_{n, m} = \textit{GetFrame}(n, m)$\;
            \tcp{Get corresponding rays for this frame}
            $\textit{rays}_{n, m} = \textit{GetRays}(n, m)$\;
            \If{$m$ is \textbf{not} divisible by 8}{
                \tcp{Downsample rays and colors by a factor of 4}
                $C_{n, m} \leftarrow \textit{NearestNeighborDownsample}(C_{n, m}, 4)$\;
                $\textit{rays}_{n, m} \leftarrow \textit{NearestNeighborDownsample}(\textit{rays}_{n, m}, 4)$\;
                \If{$m$ is \textbf{not} divisible by 4}{
                    \tcp{Downsample rays and colors by an additional factor of 2}
                    $C_{n, m} \leftarrow \textit{NearestNeighborDownsample}(C_{n, m}, 2)$\;
                    $\textit{rays}_{n, m} \leftarrow \textit{NearestNeighborDownsample}(\textit{rays}_{n, m}, 2)$\;
                }
            }
            \tcp{Add current rays and colors to output}
            $C_{GT} \leftarrow C_{GT} + C_{n, m}$\;
            $\textit{rays}_{GT} \leftarrow \textit{rays}_{GT} + \textit{rays}_{n, m}$\;
    }
}
\caption{Training Ray-Subsampling Scheme}
\label{alg:subsampling}
\end{algorithm*}

\subsection{Training Ray-Subsampling}
We provide pseudo-code for our ray-subsampling scheme in \cref{alg:subsampling}, which is used to enable more memory efficient training.

\subsection{LPIPS Evaluation Details}
For LPIPS computation, we use the AlexNet LPIPS variant for all of our comparisons in the main paper (as do all of the baseline methods).

\subsection{SSIM Evaluation Details}
\label{sec:ssim}

For SSIM computation, we use the \href{https://scikit-image.org/docs/stable/auto_examples/transform/plot_ssim.html}{\textit{structural\_similarity}} \textit{scikit-image} library function, with our images normalized to the range of $[0, 1]$, and the \textit{data\_range} parameter set to $1$. We note, however, that several methods either:

\begin{enumerate}
    \item Use their own implementation of SSIM, which are not consistent with this standard implementation (e.g. R2L~\cite{WangZLZZZWXY2022});
    \item Fail to set the \textit{data\_range} parameter appropriately, so that it defaults to the value of 2.0 (e.g. Neural 3D Video ~\cite{LiSZGLKSLGL2022}).
\end{enumerate}

In both of these cases, the SSIM function returns higher-than-intended values. While we believe that this inconsistency makes SSIM scores somewhat less reliable, we still report our aggregated SSIM metrics in the quantitative result tables in the main paper.

\section{Sample Prediction Network Details}
\label{sec:network_details}

\subsection{Additional Training Details}
For both static and dynamic datasets, we use a batch size of 16,384 rays for training, an initial learning rate of 0.02 for the parameters of the keyframe-based volume, and an initial learning rate of 0.0075 for our sample prediction network.
For Technicolor, Google Immersive, and all static scenes, we set the $w_\text{TV}$ weight Equation 14 to 0.05 for both appearance and density, which is decayed by a factor of 0.1 every 30,000 iterations.
On the other hand, $w_\text{L1}$ starts at $8 \!\cdot\! 10^{-5}$ and decays to $4 \!\cdot\! 10^{-5}$ over 30,000 iterations and is only applied to the density components.

\subsection{Additional Network Details}
In order to make it so that the sample network outputs (primitives $G_1, \dots, G_n$, point offsets $\mathbf{d}_1, \dots, \mathbf{d}_n$, velocities $\mathbf{v_1}, \dots, \mathbf{v}_n$) vary smoothly, we use 1 positional encoding frequency for the ray $\mathbf{r}$ (in both static and dynamic settings) and 2 positional encoding frequencies for the time step $\tau$ (in dynamic settings).

\subsection{Forward Facing Scenes}

For forward facing scenes, we first convert all rays to normalized device coordinates (NDC)~\cite{MildeSTBRN2020}, so that the view frustum of a ``reference'' camera lives within $[-1, 1]^{3}$. After mapping a ray with origin $\mathbf{o}$ and direction $\dirout$ to its two-plane parameterization~\cite{LevoyH1996} (with planes at $z = -1$ and $z=0$), we predict the parameters of a set of planes normal to the z-axis with our sample network. In particular, we predict $(z_1, \dots, z_n)$, and intersect the ray with the axis-aligned planes at these distances to produce our sample points $(\xpos_1, \dots, \xpos_n)$.
Additionally, we initialize the values $(z_1, \dots, z_n)$ in a stratified manner, so that they uniformly span the range of $[-1, 1]$. 

\subsection{Outward Facing Scenes}

For all other (outward facing) scenes, we map a ray to its Pl\"{u}cker parameterization via
\begin{align}
    \mathbf{r} = \textit{Pl\"{u}cker}(\mathbf{o}, \dirout) = \left( \dirout, \dirout \times \mathbf{o} \right) \text{.}
\end{align}
\noindent and predict the radii of a set of spheres centered at the origin $(r_1, \dots, r_n)$. We then intersect the ray with each sphere to produce our sample points. We initialize $(r_1, \dots, r_n)$ so that they range from the minimum distance to the maximum distance in the scene. 

\subsection{Differentiable Intersection}

In both of the above cases, we make use of the implicit form of each primitive (for planes normal to the z-axis, $z = z_k$, and for the spheres centered at the origin $x^2 + y^2 + z^2 = r_k^2$) and the parameteric equation for a ray $\mathbf{o} + t_k\dirout$, to solve for the intersection distances $t_k$ (as is done in typical ray-tracers). The intersection distance is differentiable with respect to the primitive parameters, so that gradients can propagate from the color loss to the sample network.

\subsection{Implicit Color Correction}

In order to better handle multi-view datasets with inconsistent color correction / white balancing, we also output a color scale $\mathbf{c}^{\textit{scale}}_k$ and shift $\mathbf{c}^{\textit{shift}}_k$ from the sample prediction network for each sample point $\xpos_k$. These are used to modulate the color $L_e(\xpos_k, \dirout, \tau_i)$ extracted from the dynamic volume via:

\begin{align}
    L_e(\xpos_k, \dirout, \tau_i) \leftarrow L_e(\xpos_k, \dirout, \tau_i) \cdot 
 \mathbf{c}^{\textit{scale}}_k + \mathbf{c}^{\textit{shift}}_k \text{.}
    \label{eq:scaleshift}
\end{align}

\noindent Note that these outputs vary with low-frequency with respect to the input ray (since we use few positional encoding frequencies for the sample prediction network). Additionally, the density from the volume remains unchanged.

\section{Keyframe-Based Volume Details}
\label{sec:keyframe_details}

We initialize our keyframe-based dynamic volume within a $128^{3}$ grid, so that each of the spatial tensor components have resolution 128$\times$128. Our final grid size is $640^{3}$. We upsample the volume at iterations \textit{4,000, 6,000, 8,000, 10,000}, and \textit{12,000}, interpolating the resolution linearly in log space.

\section{Quantitative Comparisons}
\label{sec:additional_comparisons}

\subsection{LLFF Dataset}
The LLFF dataset \cite{MildeSTBRN2020} contains eight real-world sequences with 1008$\times$756 pixel images.
In \cref{tab:quant_static_llff}, we compare our method to the same approaches as above on this dataset.
Our approach outperforms DoNeRF, AdaNeRF, TermiNeRF, and InstantNGP but achieves slightly worse quality than NeRF.
This dataset is challenging for explicit volume representations (which have more parameters and thus can more easily overfit to the training images) due to a combination of erroneous camera calibration and input-view sparsity.
For completeness, we also include a comparison to R2L on the downsampled 504$\times$378 LLFF dataset, where we perform slightly worse in terms of quality.

\subsection{DeepView Dataset}

Unfortunately, Google's Immersive Light Field Video \cite{BroxtFOEHDDBWD2020} does not provide quantitative benchmarks for the performance of their approach in terms of image quality.
As a proxy, we compare our approach to DeepView \cite{FlynnBDDFOST2019}, the method upon which their representation is built, on the static \textit{Spaces} dataset in \cref{tab:quant_google}.

Our method achieves superior quality, outperforming DeepView by a large margin.
Further, HyperReel consumes less memory per frame than the Immersive Light Field Video's baked layered mesh representation: 1.2\,MB per frame vs. 8.87\,MB per frame (calculated from the reported bitrate numbers \cite{BroxtFOEHDDBWD2020}).
Their layered mesh can render at more than 100\,FPS on commodity hardware, while our approach renders at a little over 4 FPS.
However, our approach is entirely implemented in vanilla PyTorch and can be further optimized using custom CUDA kernels or baked into a real-time renderable representation for better performance.

\begin{table}
\caption{\label{tab:quant_static_llff}%
    \textbf{Quantitative comparisons on LLFF.}
    We compare our approach to others on the
    real-world LLFF dataset \cite{MildeSTBRN2020}.
    FPS is normalized per megapixel; memory in MB.
}
\vspace{-2mm}
\renewcommand{\arraystretch}{1}
\centering
\resizebox{\linewidth}{!}{%
\begin{tabular}{@{}llcc@{\hspace{0.75\tabcolsep}}r@{}}
  \toprule
  Dataset & Method & PSNR$\uparrow$ & FPS$\uparrow$ & Memory $\downarrow$ \\

  \midrule
  
  \multirow{3}{*}{LLFF 504$\times$378}
  & \textit{Single sample} & & &\\
  & \ \ R2L~\cite{WangRHOCFT2022} & \textbf{27.7} & --- & \bf 23.7 \\
  \cmidrule{2-5}
  & Ours (per-frame) & 27.5 & \bf 4.0 & 58.8 \\

  \midrule
  
  \multirow{8}{*}{LLFF 1008$\times$756}
  & \textit{Uniform sampling} & & & \\
  & \ \ NeRF~\cite{MildeSTBRN2020} & \textbf{26.5} & 0.3 & \bf 3.8 \\
  & \ \ Instant NGP~\cite{MuelleESK2022} & 25.6 & 5.3 & 64.0  \\
  \cmidrule{2-5}
  & \textit{Adaptive sampling} &&\\
  & \ \ DoNeRF~\cite{NeffSPKCKS2021} & 22.9 & 2.1 & 4.1 \\
  & \ \ AdaNeRF~\cite{KurzNLZS2022} & 25.7 & \textbf{5.6} & 4.1 \\
  & \ \ TermiNeRF~\cite{PialaC2021} & 23.6 & 2.1 & 4.1 \\
  \cmidrule{2-5}
  & Ours (per-frame) & 26.2 & 4.0 & 58.8 \\
  \bottomrule
\end{tabular}%
}
\end{table}

\begin{table}
\caption{\label{tab:quant_google}%
  \textbf{Quantitative comparisons to DeepView.}
  In addition to the comparison to NeRFPlayer, we report a comparison with DeepView \cite{FlynnBDDFOST2019}, a variant of which
  is used per-frame in immersive LF video \cite{BroxtFOEHDDBWD2020}.
  We thus compare to DeepView as a proxy for quantitative comparison.
  FPS normalized per megapixel.
}
\vspace{-2mm}
\renewcommand{\arraystretch}{1}
\centering
\resizebox{\linewidth}{!}{%
\begin{tabular}{@{}llccc@{\hspace{0.75\tabcolsep}}r@{}}
  \toprule
  Dataset & Method & PSNR$\uparrow$ & SSIM$\uparrow$ & LPIPS$\downarrow$ & FPS$\uparrow$ \\
  \midrule
  \multirow{2}{*}{Spaces~\cite{FlynnBDDFOST2019}}
  & DeepView~\cite{FlynnBDDFOST2019} & 31.60 & 0.965 & 0.085 & \bf $>$100 \\
  & Ours & \bf 35.47 & \bf 0.968 & \bf 0.080 & 4.0  \\
  \bottomrule
\end{tabular}%
}
\end{table}

\section{Qualitative Comparisons to Neural 3D~\cite{LiSZGLKSLGL2022}}
\label{sec:additional_n3d}

\begin{figure*}[t]
\footnotesize
\centering%
  \mpage{0.32}{%
    \mpage{0.49}{%
      \begin{tikzpicture}
        \node[anchor=south west,inner sep=0] (image) at (0,0) {\adjincludegraphics[height=\linewidth,trim={{0.14\width} 0 {0.3\width} 0},clip,angle=0]{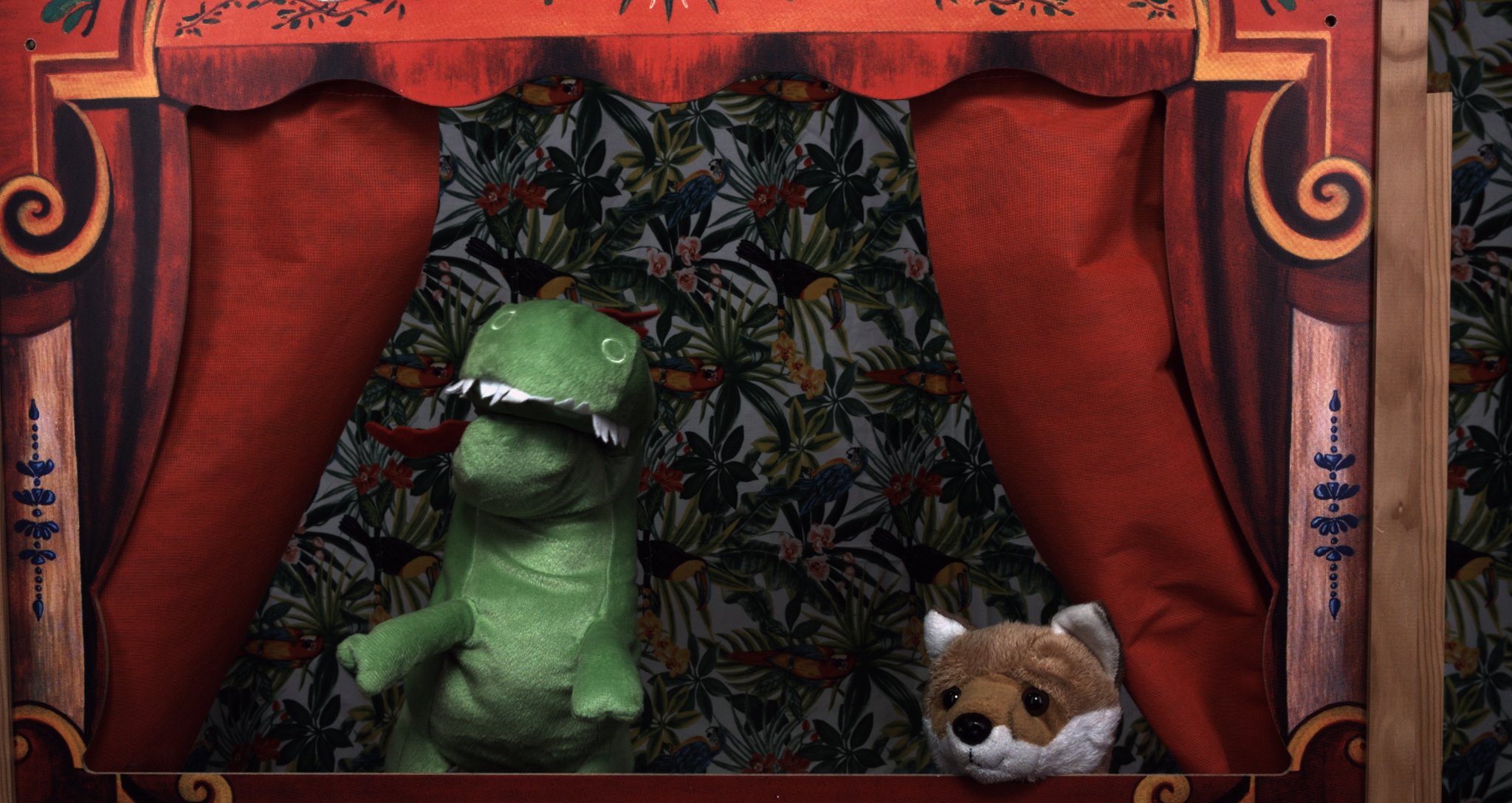}};
        \begin{scope}[x={($1*(image.south east)$)},y={($1*(image.north west)$)}]
          \draw[red,thick] (0.22,0.13) rectangle (0.64,0.68);
        \end{scope}
      \end{tikzpicture}%
    }\hspace{-1mm}\hfill%
    \mpage{0.49}{%
      \adjincludegraphics[height=\linewidth,trim={{0.27\width} {0.22\height} {0.54\width} {0.37\height}},clip,angle=0]{figures/fig_dynamic/img/technicolor4/0000_gt.png}\\[0.2mm]%
    }
  }\hfill%
  \mpage{0.32}{%
    \mpage{0.49}{%
      \begin{tikzpicture}
        \node[anchor=south west,inner sep=0] (image) at (0,0) {\adjincludegraphics[height=\linewidth,trim={{0.14\width} 0 {0.3\width} 0},clip,angle=0]{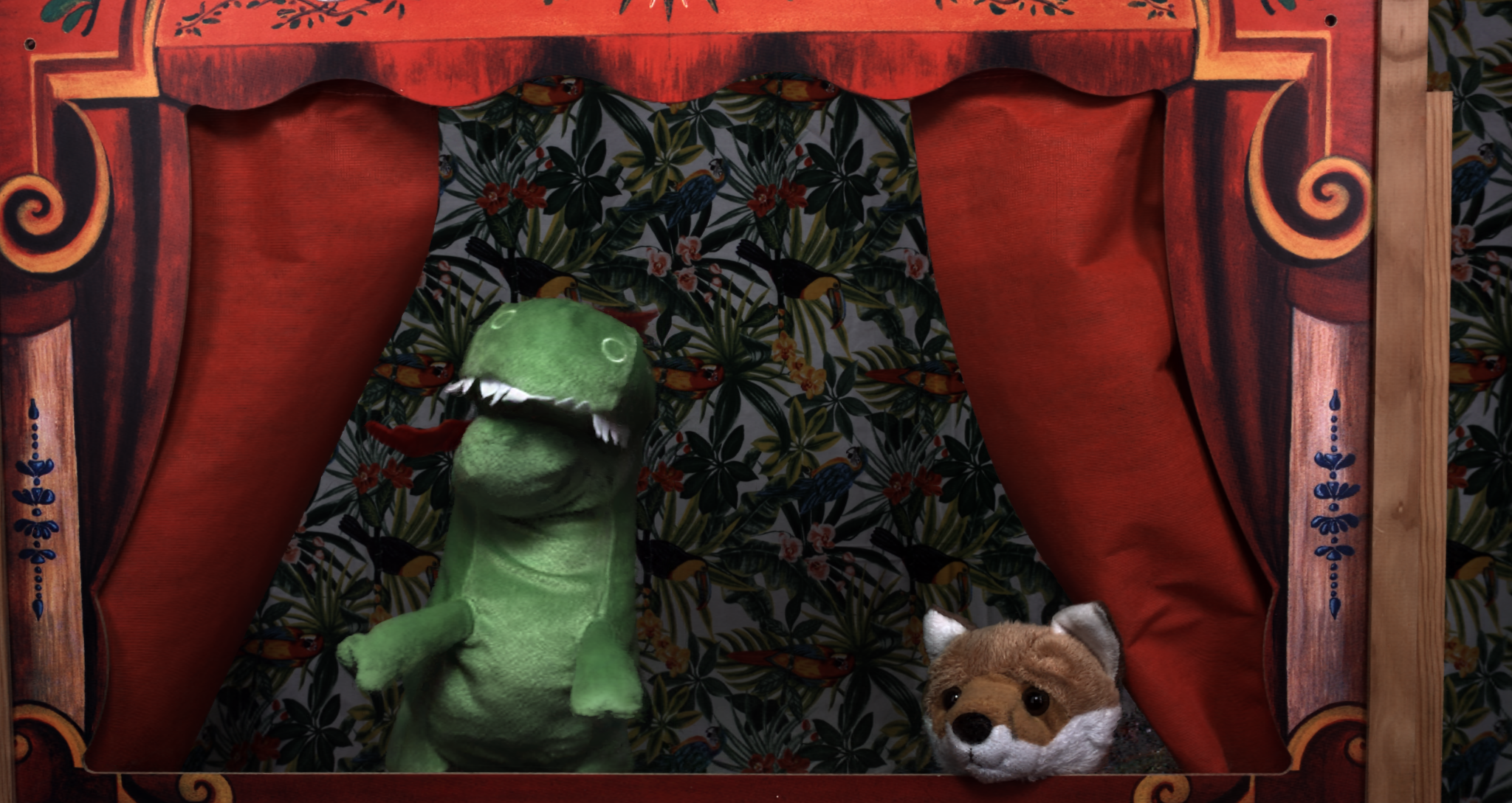}};
        \begin{scope}[x={($1*(image.south east)$)},y={($1*(image.north west)$)}]
          \draw[red,thick] (0.22,0.13) rectangle (0.64,0.68);
        \end{scope}
      \end{tikzpicture}%
    }\hspace{-1mm}\hfill%
    \mpage{0.49}{%
      \adjincludegraphics[height=\linewidth,trim={{0.27\width} {0.22\height} {0.54\width} {0.37\height}},clip,angle=0]{figures/fig_dynamic/img/technicolor4/0000_ours.png}\\[0.2mm]%
    }
  }\hfill%
  \mpage{0.32}{%
    \mpage{0.49}{%
      \begin{tikzpicture}
        \node[anchor=south west,inner sep=0] (image) at (0,0) {\adjincludegraphics[height=\linewidth,trim={{0.14\width} 0 {0.3\width} 0},clip,angle=0]{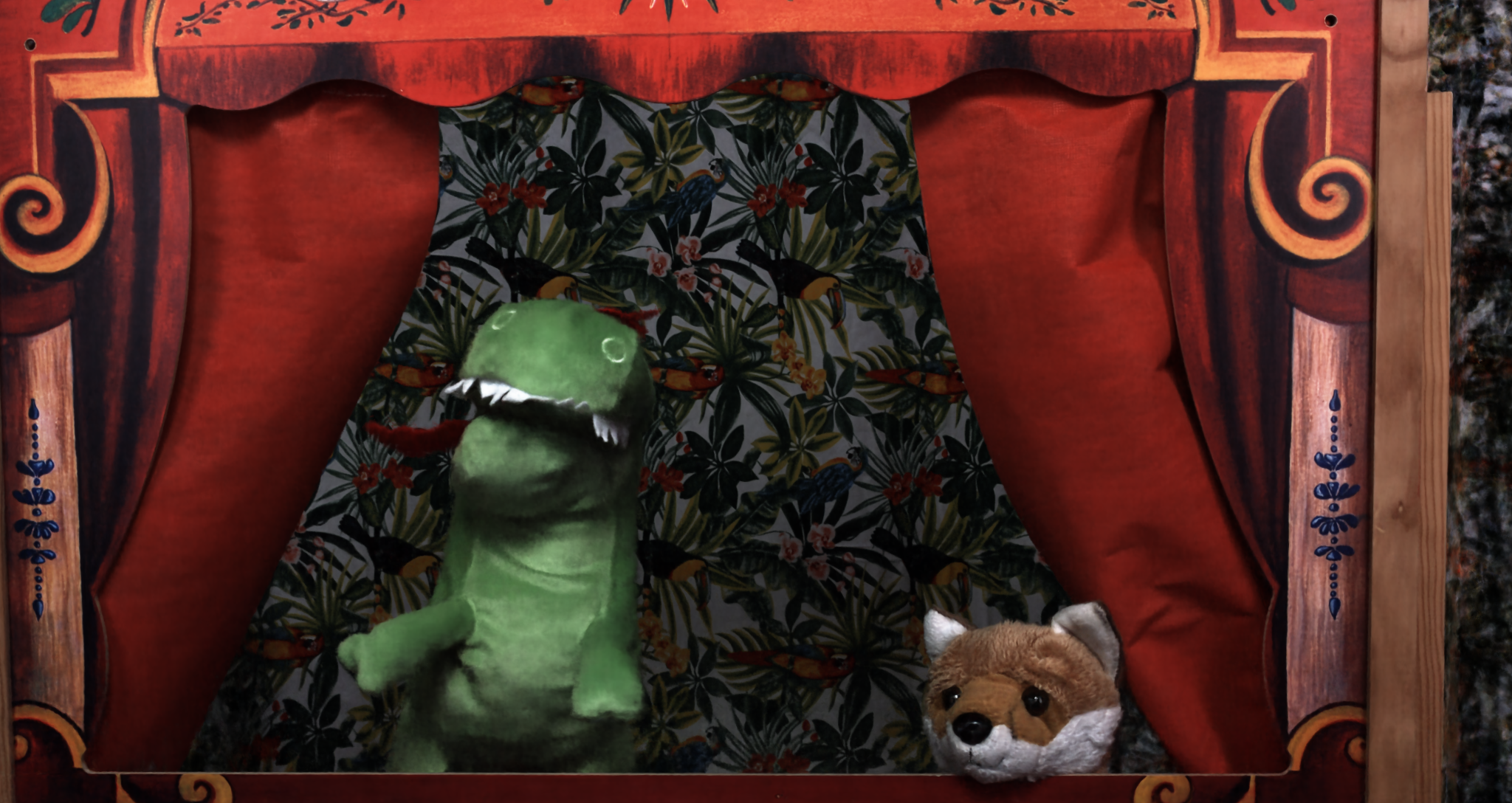}};
        \begin{scope}[x={($1*(image.south east)$)},y={($1*(image.north west)$)}]
          \draw[red,thick] (0.22,0.13) rectangle (0.64,0.68);
        \end{scope}
      \end{tikzpicture}%
    }\hspace{-1mm}\hfill%
    \mpage{0.49}{%
      \adjincludegraphics[height=\linewidth,trim={{0.27\width} {0.22\height} {0.54\width} {0.37\height}},clip,angle=0]{figures/fig_dynamic/img/technicolor4/0000_n3d.png}\\[0.2mm]%
    }
  }\\[0.1mm]%
  \mpage{0.32}{%
    \mpage{0.49}{%
      \begin{tikzpicture}
        \node[anchor=south west,inner sep=0] (image) at (0,0) {\includegraphics[width=\linewidth]{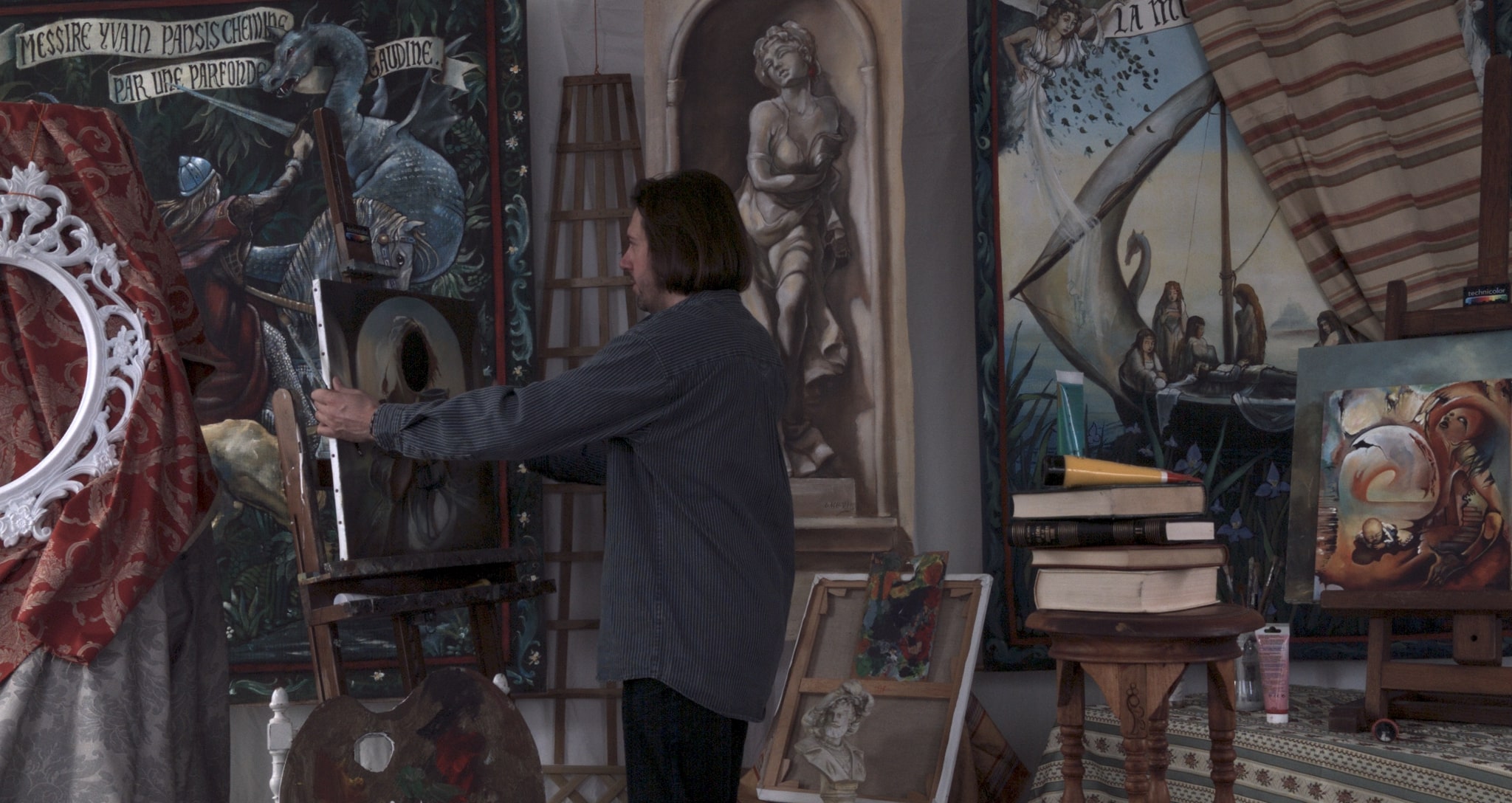}};
        \begin{scope}[x={($1*(image.south east)$)},y={($1*(image.north west)$)}]
          \draw[red,thick] (0.39,0.46) rectangle (0.51,0.58);
        \end{scope}
      \end{tikzpicture}%
    }\hspace{-1mm}\hfill%
    \mpage{0.49}{%
      \adjincludegraphics[width=\linewidth,trim={{0.39\width} {0.46\height} {0.49\width} {0.42\height}},clip]{figures/fig_dynamic/img/technicolor2/0002_gt.jpg}\\[0.2mm]%
    }
  }\hfill%
  \mpage{0.32}{%
    \mpage{0.49}{%
      \begin{tikzpicture}
        \node[anchor=south west,inner sep=0] (image) at (0,0) {\includegraphics[width=\linewidth]{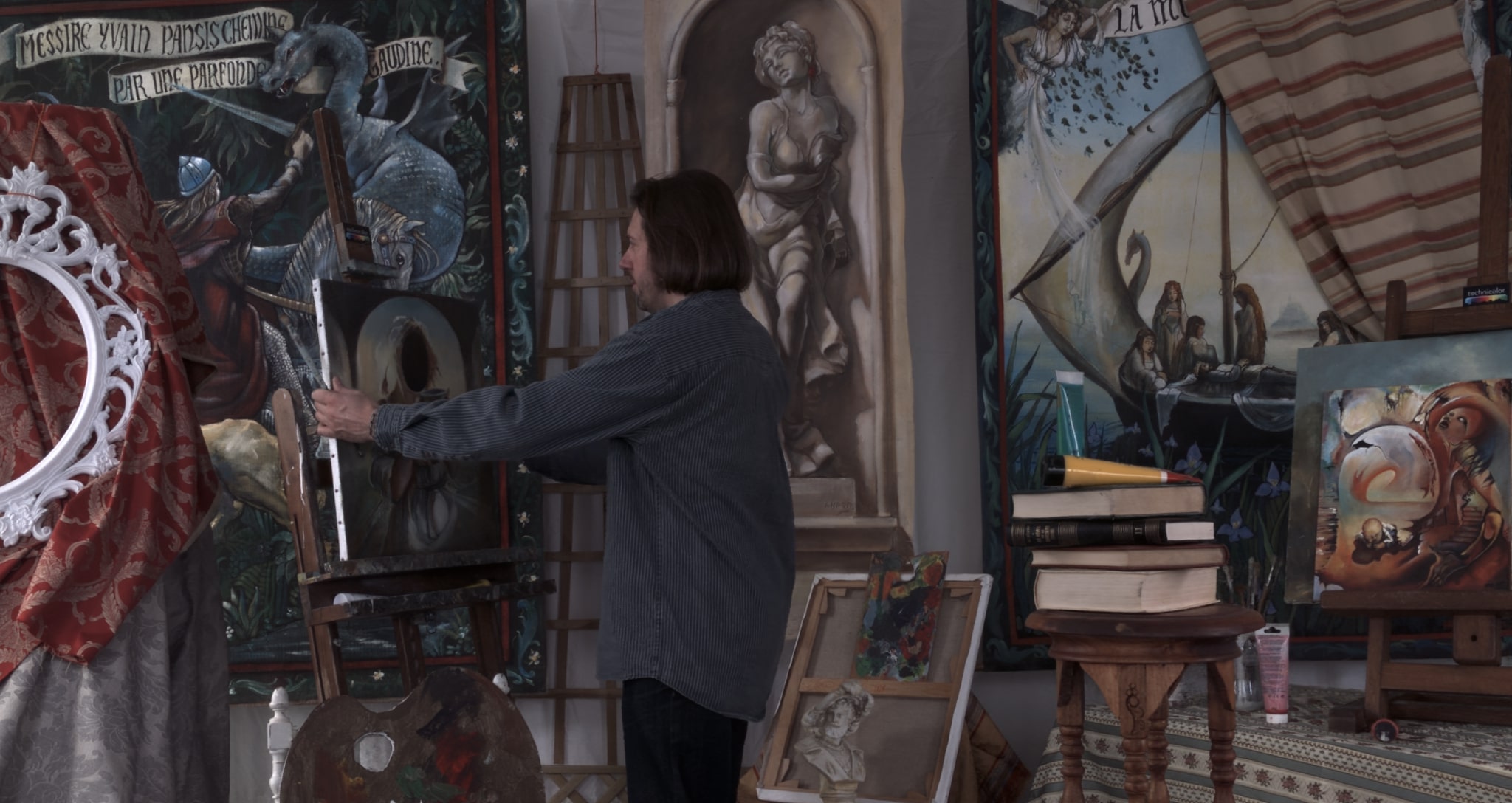}};
        \begin{scope}[x={($1*(image.south east)$)},y={($1*(image.north west)$)}]
          \draw[red,thick] (0.39,0.46) rectangle (0.51,0.58);
        \end{scope}
      \end{tikzpicture}%
    }\hspace{-1mm}\hfill%
    \mpage{0.49}{%
      \adjincludegraphics[width=\linewidth,trim={{0.39\width} {0.46\height} {0.49\width} {0.42\height}},clip]{figures/fig_dynamic/img/technicolor2/0002_ours.jpg}\\[0.2mm]%
    }
  }\hfill%
  \mpage{0.32}{%
    \mpage{0.49}{%
      \begin{tikzpicture}
        \node[anchor=south west,inner sep=0] (image) at (0,0) {\includegraphics[width=\linewidth]{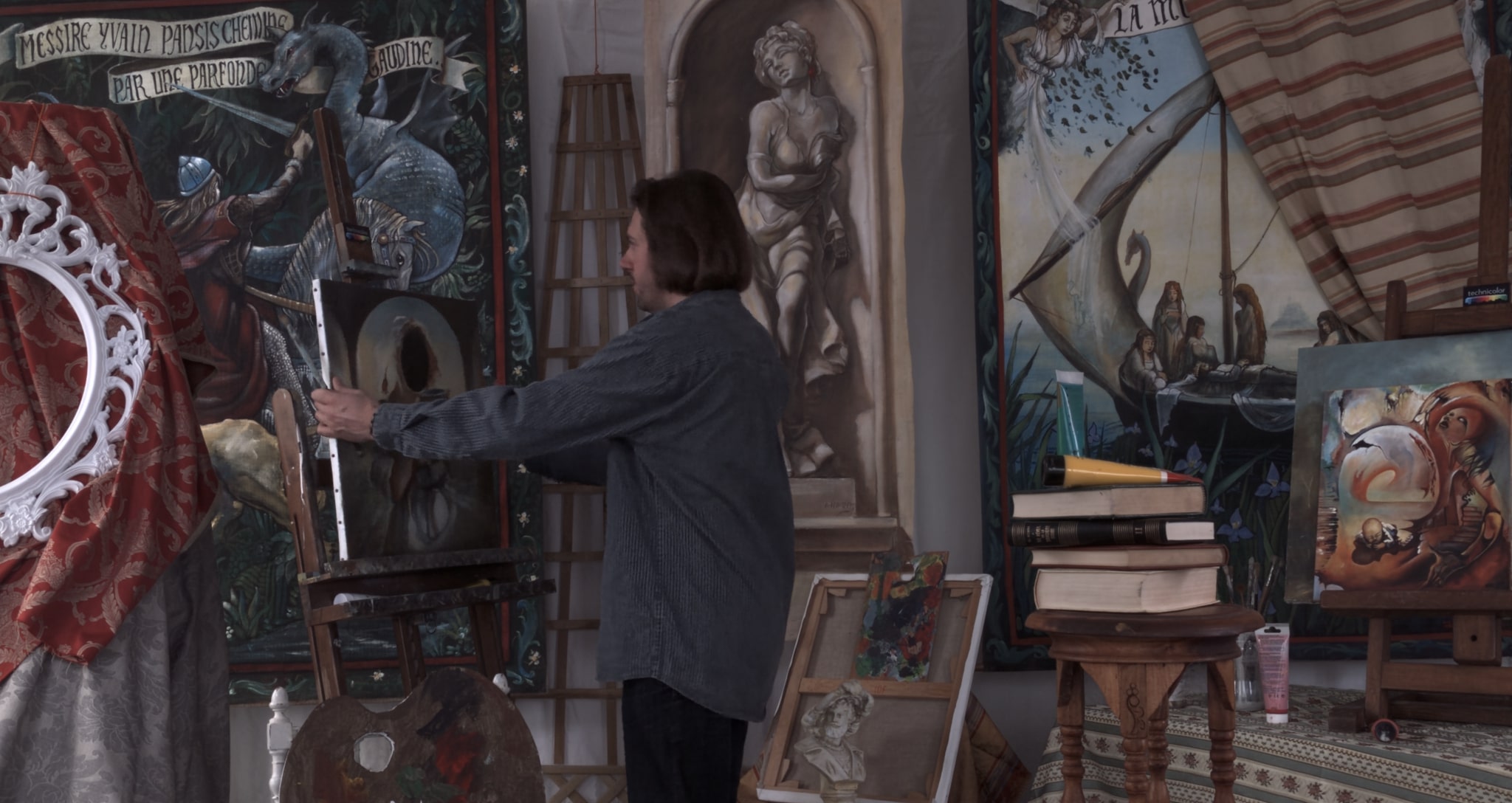}};
        \begin{scope}[x={($1*(image.south east)$)},y={($1*(image.north west)$)}]
          \draw[red,thick] (0.39,0.46) rectangle (0.51,0.58);
        \end{scope}
      \end{tikzpicture}%
    }\hspace{-1mm}\hfill%
    \mpage{0.49}{%
      \adjincludegraphics[width=\linewidth,trim={{0.39\width} {0.46\height} {0.49\width} {0.42\height}},clip]{figures/fig_dynamic/img/technicolor2/0002_n3d.jpg}\\[0.2mm]%
    }
  }\\[0.1mm]%
  \mpage{0.32}{%
    Ground truth (Technicolor~\cite{sabater2017dataset})%
  }\hfill%
  \mpage{0.32}{%
    Ours%
  }\hfill%
  \mpage{0.32}{%
    Neural 3D Video~\cite{LiSZGLKSLGL2022}
  }\\[1mm]%
\caption{%
\textbf{Additional qualitative comparisons to Neural 3D Video Synthesis.} 
We show two additional qualitative comparisons against Neural 3D Video Synthesis~\cite{LiSZGLKSLGL2022} on the Technicolor dataset~\cite{sabater2017dataset}, demonstrating that our approach recovers more accurate/detailed appearance.
}%
\label{fig:dynamic_supp}
\vspace{1mm}
\end{figure*}
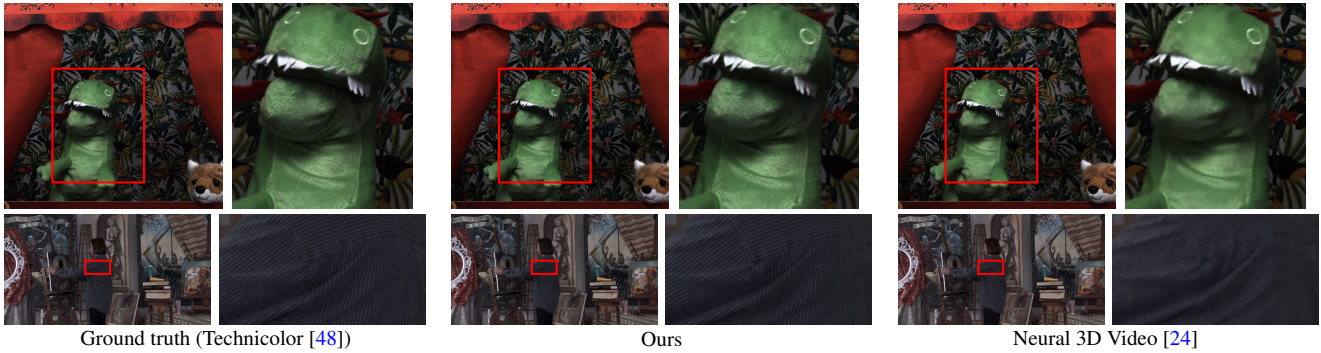

We provide additional qualitative still-frame comparisons to Neural 3D Video Synthesis~\cite{LiSZGLKSLGL2022} in~\cref{fig:dynamic_supp}.

\section{Additional Results}
\label{sec:additional_results}

\subsection{Panoramic 6-DoF Video}

In general, our method can support an unlimited FoV. We show a panoramic rendering of a synthetic 360 degree scene from our model, using spherical primitives in \cref{fig:360}.

\subsection{Point Offsets for Modeling Refractions}

\begin{figure*}[htp]
\footnotesize
\centering%
	\mpage{1.0}{%
		\begin{overpic}[width=\linewidth]{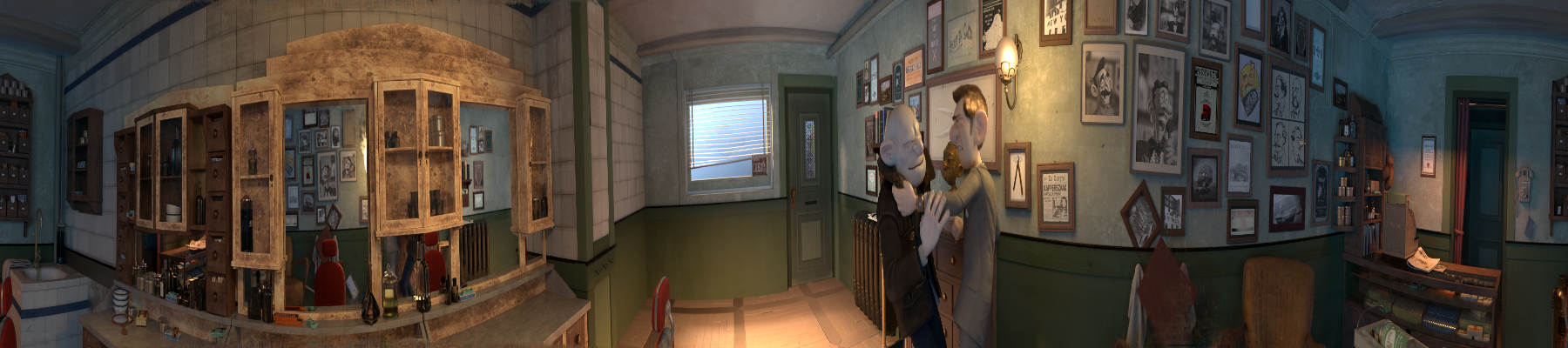}%
		\end{overpic}
	}%
\caption{%
Example panoramic rendering from our approach applied to a synthetic scene with captures spanning a full 360 degree FoV. In this case, our sample network predicts spherical geometric primitives. The scene is one of the shots from the Blender Foundations Agent 327 open movie~\cite{blenderfoundation}.
}%
\label{fig:360}
\vspace{5mm}

\footnotesize
\centering%
	\mpage{0.2}{%
		\begin{overpic}[width=\linewidth]{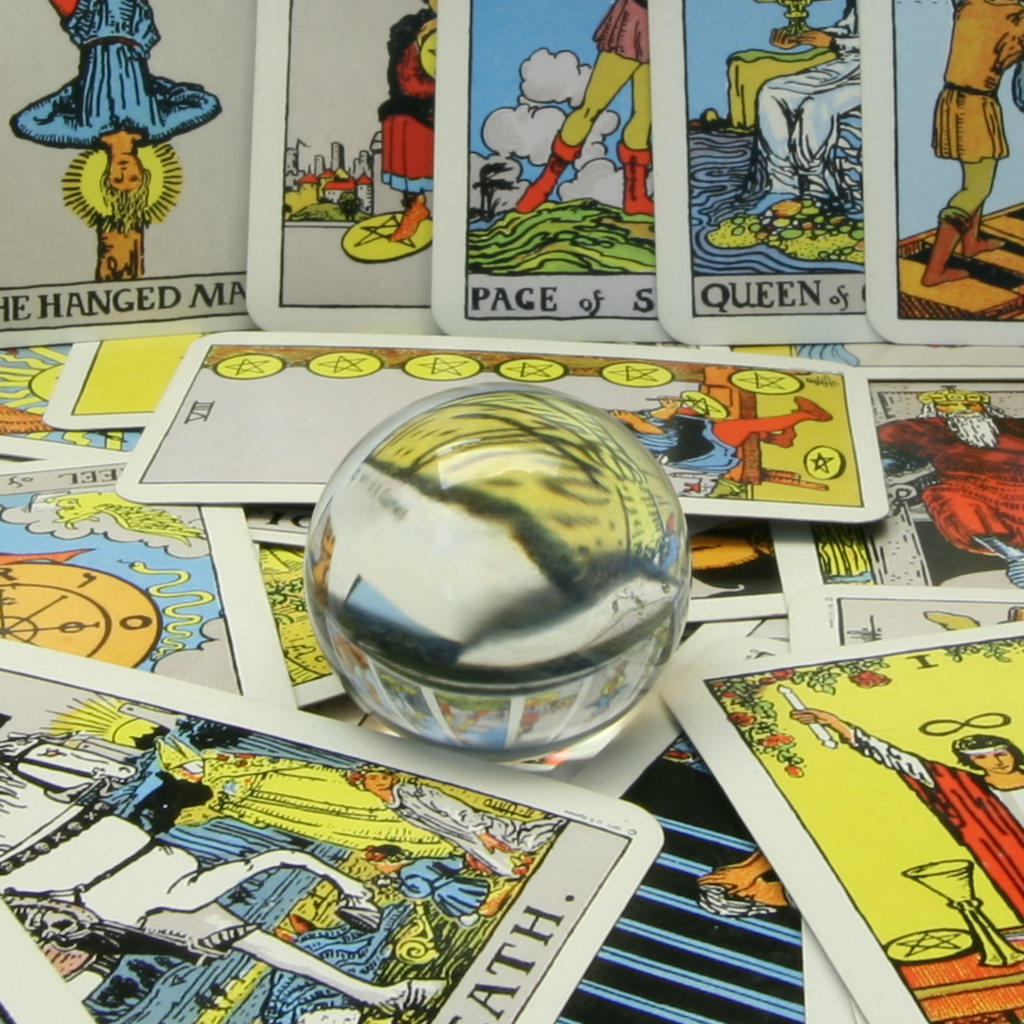}%
		\end{overpic}\\[-0.5mm]%
		\emph{Tarot}
	}%
	\hspace{-1mm}\hfill%
	\mpage{0.79}{%
		\mpage{0.3}{%
			\includegraphics[trim={10cm 8cm 10cm 12cm},clip,width=\linewidth]{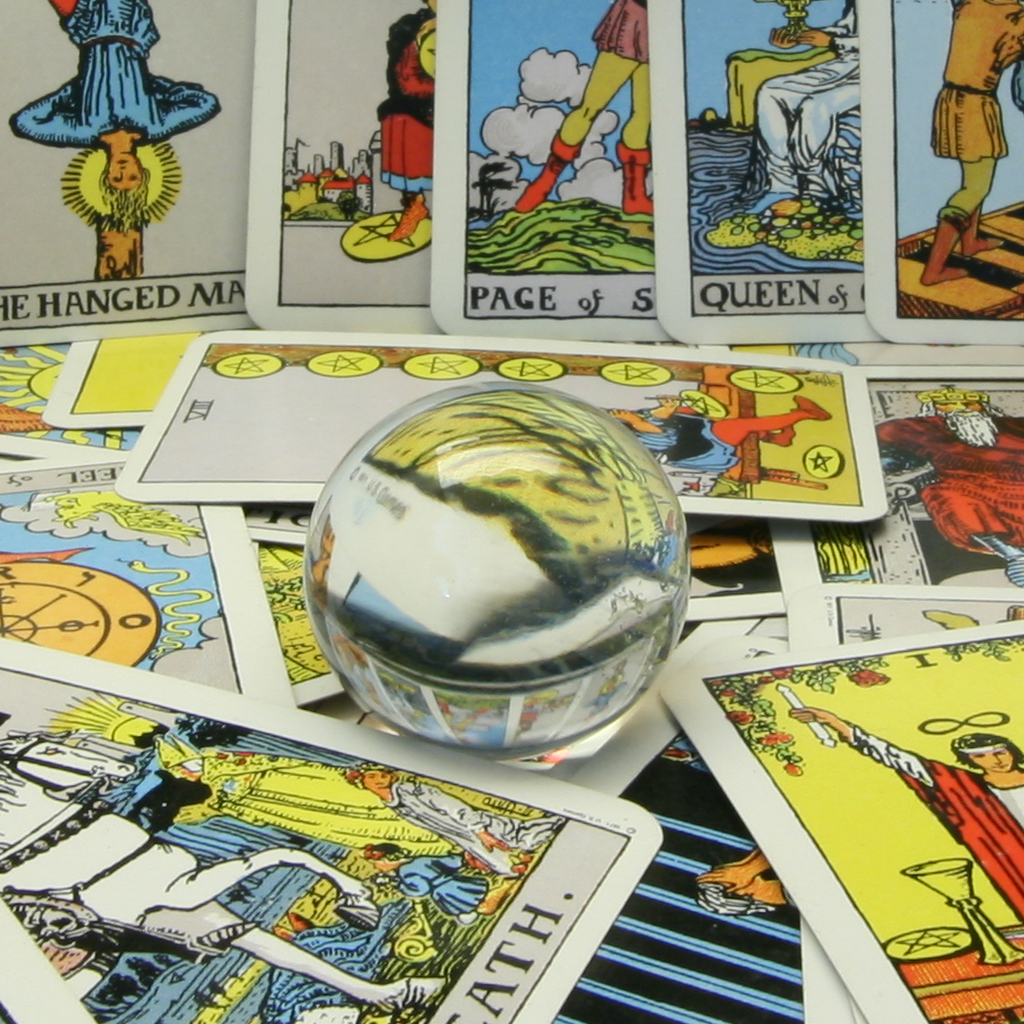}\\[-0.5mm]%
		}\hfill%
		\mpage{0.3}{%
			\includegraphics[trim={10cm 8cm 10cm 12cm},clip,width=\linewidth]{figures/rebuttal/point.png}\\[-0.5mm]%
		}\hfill%
		\mpage{0.3}{%
			\includegraphics[trim={10cm 8cm 10cm 12cm},clip,width=\linewidth]{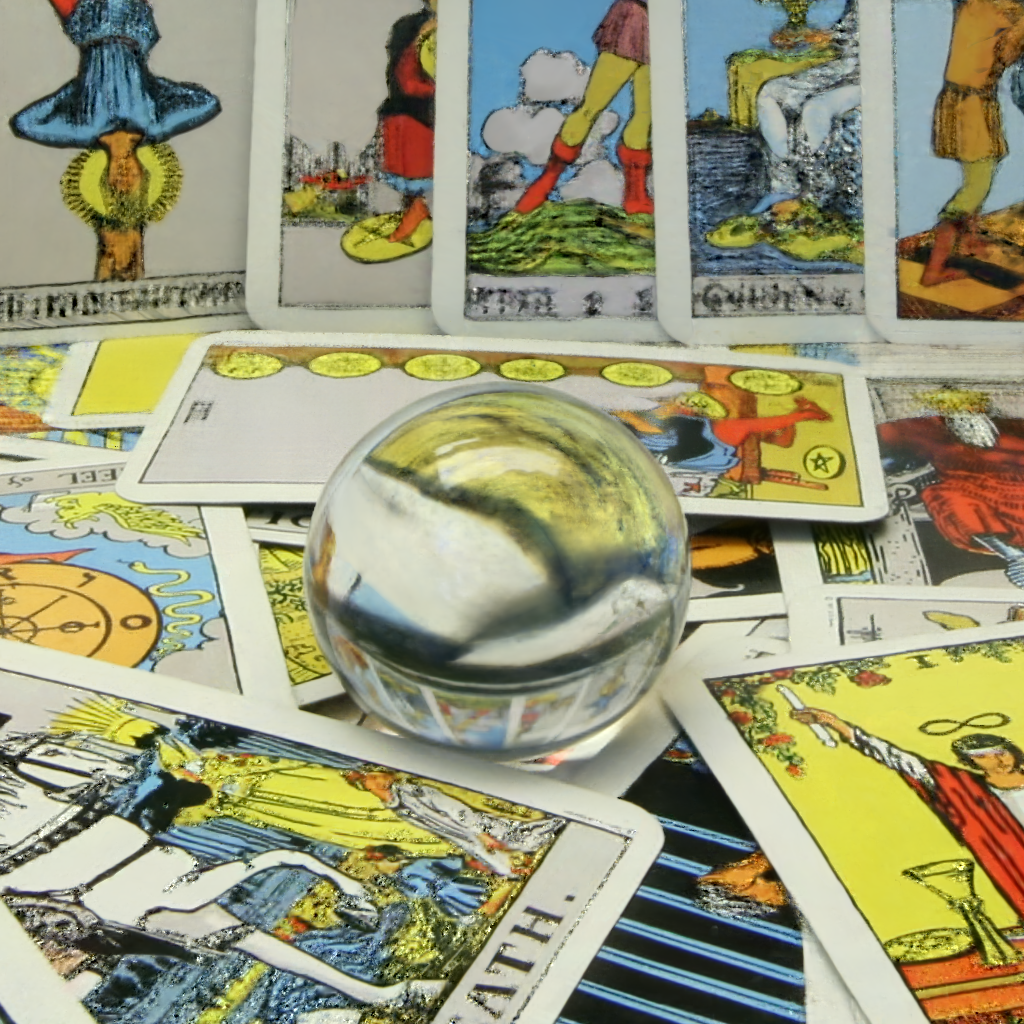}\\[-0.5mm]%
		}
	}\\[1mm]
	\mpage{0.20}{%
	}%
	\hspace{-1mm}\hfill%
	\mpage{0.79}{%
		\mpage{0.24}{%
			GT%
		}\hspace{-1mm}\hfill%
		\mpage{0.24}{%
			w/ offset%
		}\hspace{-1mm}\hfill%
		\mpage{0.24}{%
			w/o offset
		}

	}\\[1mm]%
 	\mpage{0.2}{%
		\begin{overpic}[trim={10cm 0cm 0cm 0cm},clip,width=\linewidth]{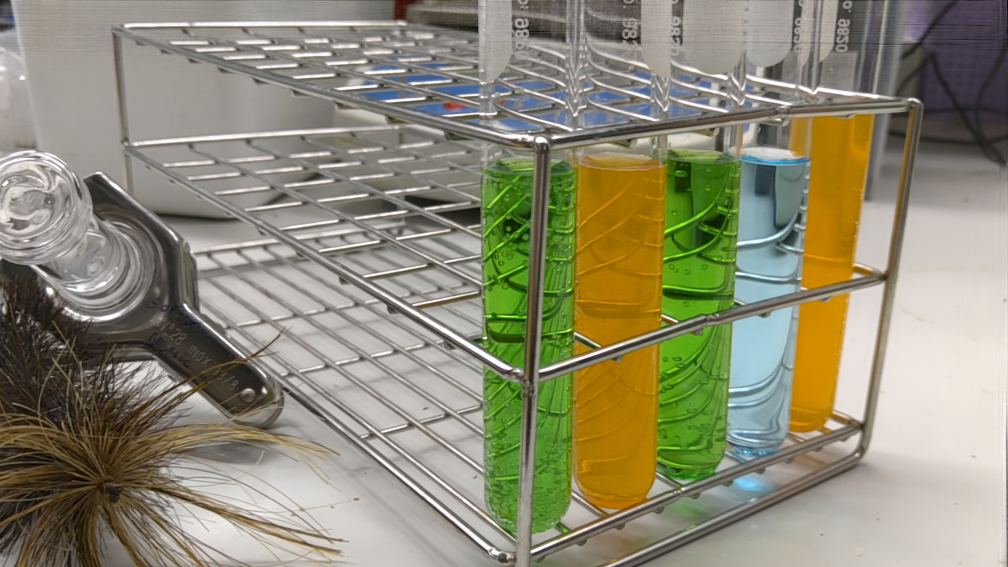}%
		\end{overpic}\\[-0.5mm]%
		\emph{Tarot}
	}%
	\hspace{-1mm}\hfill%
	\mpage{0.79}{%
		\mpage{0.3}{%
			\includegraphics[trim={16cm 2cm 12cm 10cm},clip,width=\linewidth]{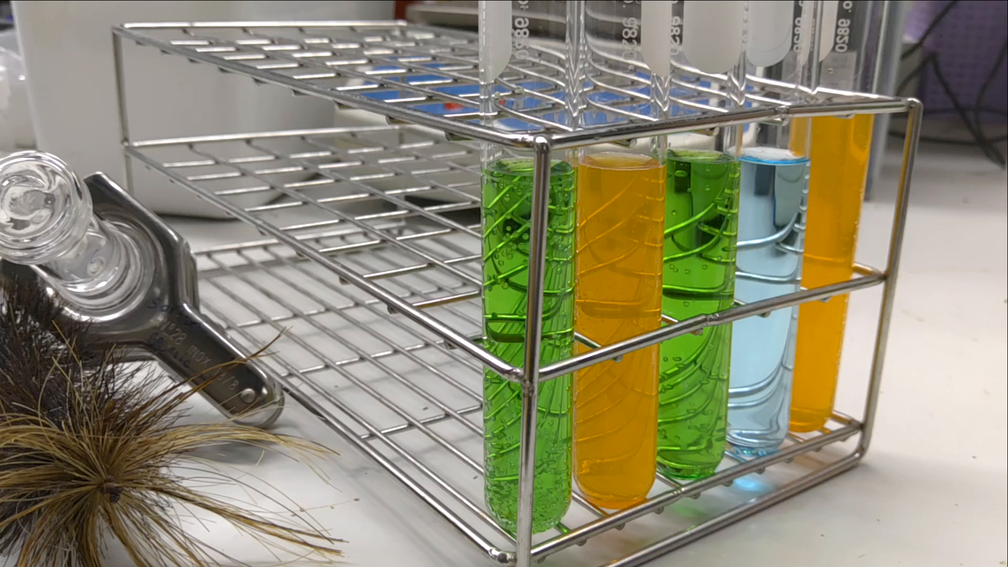}\\[-0.5mm]%
		}\hfill%
		\mpage{0.3}{%
			\includegraphics[trim={16cm 2cm 12cm 10cm},clip,width=\linewidth]{figures/rebuttal/lab_point.png}\\[-0.5mm]%
		}\hfill%
		\mpage{0.3}{%
			\includegraphics[trim={16cm 2cm 12cm 10cm},clip,width=\linewidth]{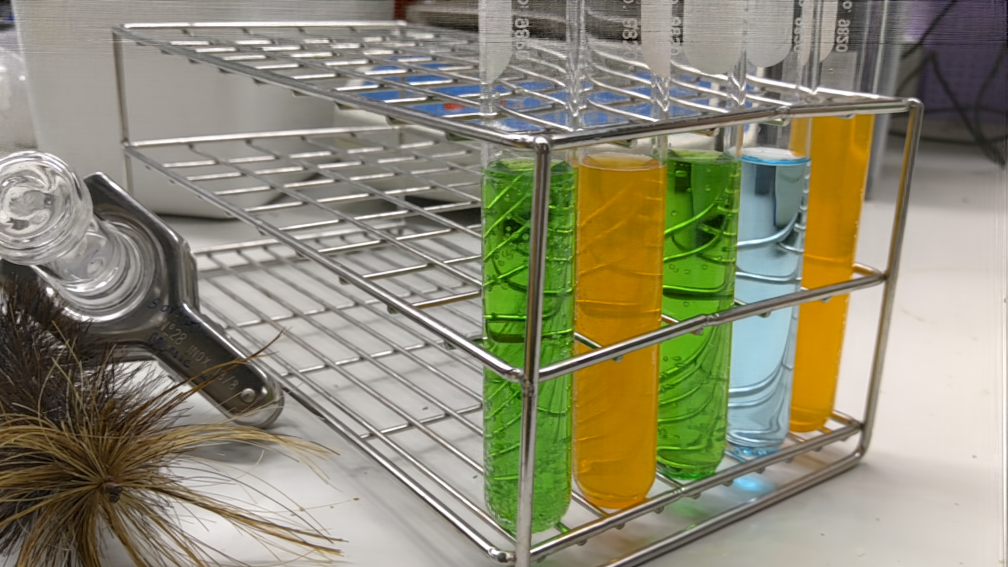}\\[-0.5mm]%
		}
	}\\[1mm]
	\mpage{0.20}{%
	}%
	\hspace{-1mm}\hfill%
	\mpage{0.79}{%
		\mpage{0.24}{%
			GT%
		}\hspace{-1mm}\hfill%
		\mpage{0.24}{%
			w/ offset%
		}\hspace{-1mm}\hfill%
		\mpage{0.24}{%
			w/o offset
		}

	}\\[-2mm]%
\caption{%
Comparison of our method with and without point offset on the \textit{Tarot} sequence from the Stanford Light Field dataset~\cite{WilbuJVTABAHL2005} and \textit{Lab} sequence from the Shiny dataset~\cite{WizadPYS2021}.
}%
\label{fig:refractive}
\vspace{-5mm}
\end{figure*}

Point offsets allow the sample network to capture appearance that violates epipolar constraints, noticeably improving quality for refractive scenes. We show a visual comparison between our approach with and without point offsets in \cref{fig:refractive}. More results are available on the website.

\begin{table*}[!htb]
\centering
\caption{Per-scene results from the Technicolor dataset~\cite{sabater2017dataset}. See Section~\cref{sec:ssim} for a discussion of the reliability of SSIM metrics.}
\vspace{-1mm}
\small
\resizebox{\linewidth}{!}{%
\begin{tabular}{%
  @{}%
  l@{\hspace{3\tabcolsep}}%
  c@{\hspace{0.2\tabcolsep}}c@{\hspace{0.9\tabcolsep}}c@{\hspace{0.8\tabcolsep}}c%
  c@{\hspace{0.2\tabcolsep}}c@{\hspace{0.9\tabcolsep}}c@{\hspace{0.8\tabcolsep}}c%
  c@{\hspace{0.2\tabcolsep}}c@{\hspace{0.9\tabcolsep}}c@{\hspace{0.8\tabcolsep}}c@{}%
}
\toprule
\multirow{2}{*}[-0.5ex]{Scene}
& \multicolumn{4}{c}{PSNR$\uparrow$} & \multicolumn{4}{c}{SSIM$\uparrow$} & \multicolumn{4}{c}{LPIPS$\downarrow$} \\
\cmidrule(r){2-5} \cmidrule(lr){6-9} \cmidrule(l){10-13}
& Neural 3D Video~\cite{LiSZGLKSLGL2022} & Ours & Small & Tiny \hspace{-1mm}
& Neural 3D Video~\cite{LiSZGLKSLGL2022} & Ours & Small & Tiny \hspace{-1mm}
& Neural 3D Video~\cite{LiSZGLKSLGL2022} & Ours & Small & Tiny \hspace{-1mm} \\
\midrule
Birthday     & 29.20 & \bf 29.99 & 29.32 & 27.80 & \bf 0.952 & 0.922 & 0.907 & 0.876 & 0.0668 & \bf 0.0531 & 0.0622 & 0.0898 \\
Fabien     & 32.76 & \bf 34.70 & 33.67 & 32.25 & \bf 0.965 & 0.895 & 0.882 & 0.860 & 0.2417 & \bf 0.1864 & 0.1942 & 0.2233 \\
Painter     & 35.95 & 35.91 & \bf 36.09 & 34.61 & \bf 0.972 & 0.923 & 0.920 & 0.905 & 0.1464 & \bf 0.1173 & 0.1182 & 0.1311 \\
Theater     & 29.53 & \bf 33.32 & 32.19 & 30.74 & \bf 0.939 & 0.895 & 0.880 & 0.845 & 0.1881 & \bf 0.1154 & 0.1306 & 0.1739 \\
Trains     & \bf 31.58 & 29.74 & 27.51 & 25.02 & \bf 0.962 & 0.895 & 0.835 & 0.773 & \bf 0.0670 & 0.0723 & 0.1196 & 0.1660 \\
\bottomrule
\end{tabular}
}
\label{tab:technicolor}

\vspace{10mm}
\caption{Per-scene results from the Neural 3D Video dataset~\cite{LiSZGLKSLGL2022}, available only for our method and NeRFPlayer~\cite{SongCLCCYXG2023}.}
\vspace{-1mm}
\small
\resizebox{0.75\linewidth}{!}{%
\begin{tabular}{%
  @{}%
  l@{\hspace{3\tabcolsep}}%
  c@{\hspace{0.2\tabcolsep}}c@{\hspace{0.9\tabcolsep}}c@{\hspace{0.8\tabcolsep}}c%
  c@{\hspace{0.2\tabcolsep}}c@{\hspace{0.9\tabcolsep}}c@{\hspace{0.8\tabcolsep}}c%
  c@{\hspace{0.2\tabcolsep}}c@{\hspace{0.9\tabcolsep}}c@{\hspace{0.8\tabcolsep}}c@{}%
}
\toprule
\multirow{2}{*}[-0.5ex]{Scene}
& \multicolumn{2}{c}{PSNR$\uparrow$} & \multicolumn{2}{c}{SSIM$\uparrow$} & \multicolumn{2}{c}{LPIPS$\downarrow$} \\
\cmidrule(r){2-3} \cmidrule(lr){4-5} \cmidrule(l){6-7}
& NeRFPlayer~\cite{SongCLCCYXG2023} & Ours \hspace{-1mm}
& NeRFPlayer~\cite{SongCLCCYXG2023} & Ours \hspace{-1mm}
& NeRFPlayer~\cite{SongCLCCYXG2023} & Ours \hspace{-1mm} \\
\midrule
Coffee Martini     & \bf 31.534 & 28.369 & \bf 0.951 & 0.892 & \bf 0.085 & 0.127 \\
Cook Spinach       & 30.577 & \bf 32.295 & 0.929 & \bf 0.941 & 0.113 & \bf 0.089 \\
Cut Roasted Beef   & 29.353 & \bf 32.922 & 0.908 & \bf 0.945 & 0.144 & \bf 0.084 \\
Flame Salmon       & \bf 31.646 & 28.260 & \bf 0.940 & 0.882 & \bf 0.098 & 0.136 \\
Flame Steak        & 31.932 & \bf 32.203 & \bf 0.950 & 0.949 & 0.088 & \bf 0.078 \\
Sear Steak         & 29.129 & \bf 32.572 & 0.908 & \bf 0.952 & 0.138 & \bf 0.077 \\
\bottomrule
\end{tabular}
}
\label{tab:n3d}

\vspace{10mm}
\caption{Per-scene results from the  Google Immersive Light Field Video dataaset~\cite{BroxtFOEHDDBWD2020}, available only for our method and NeRFPlayer~\cite{SongCLCCYXG2023}.}
\vspace{-1mm}
\small
\resizebox{0.75\linewidth}{!}{%
\begin{tabular}{%
  @{}%
  l@{\hspace{3\tabcolsep}}%
  c@{\hspace{0.2\tabcolsep}}c@{\hspace{0.9\tabcolsep}}c@{\hspace{0.8\tabcolsep}}c%
  c@{\hspace{0.2\tabcolsep}}c@{\hspace{0.9\tabcolsep}}c@{\hspace{0.8\tabcolsep}}c%
  c@{\hspace{0.2\tabcolsep}}c@{\hspace{0.9\tabcolsep}}c@{\hspace{0.8\tabcolsep}}c@{}%
}
\toprule
\multirow{2}{*}[-0.5ex]{Scene}
& \multicolumn{2}{c}{PSNR$\uparrow$} & \multicolumn{2}{c}{SSIM$\uparrow$} & \multicolumn{2}{c}{LPIPS$\downarrow$} \\
\cmidrule(r){2-3} \cmidrule(lr){4-5} \cmidrule(l){6-7}
& NeRFPlayer~\cite{SongCLCCYXG2023} & Ours \hspace{-1mm}
& NeRFPlayer~\cite{SongCLCCYXG2023} & Ours \hspace{-1mm}
& NeRFPlayer~\cite{SongCLCCYXG2023} & Ours \hspace{-1mm} \\
\midrule
01\_Welder         & \bf 25.568 & 25.554 & \bf 0.818 & 0.790 & 0.289 & \bf 0.281 \\
02\_Flames         & 26.554 & \bf 30.631 & 0.842 & \bf 0.905 & \bf 0.154 & 0.159 \\
04\_Truck          & 27.021 & \bf 27.175 & \bf 0.877 & 0.848 & \bf 0.164 & 0.223 \\
09\_Exhibit        & 24.549 & \bf 31.259 & 0.869 & \bf 0.903 & 0.151 & \bf 0.140 \\
10\_Face\_Paint\_1 & 27.772 & \bf 29.305 & \bf 0.916 & 0.913 & 0.147 & \bf 0.139 \\
11\_Face\_Paint\_2 & \bf 27.352 & 27.336 & \bf 0.902 & 0.879 & \bf 0.152 & 0.195 \\
12\_Cave           & 21.825 & \bf 30.063 & 0.715 & \bf 0.881 & 0.314 & \bf 0.214 \\
\bottomrule
\end{tabular}
}
\label{tab:immersive}
\end{table*}

\end{document}